\documentclass[acmsmall,screen]{acmart}

\AtBeginDocument{%
  }

\usepackage{mathrsfs}
\usepackage{subfigure}
\usepackage{microtype,flushend}
\usepackage{colortbl}
\usepackage{balance}
\usepackage[most]{tcolorbox}
\usepackage{multirow}
\usepackage{fontawesome}
\usepackage{threeparttable}
\usepackage{color}
\usepackage[ruled,vlined]{algorithm2e}
\usepackage{xspace}
\usepackage{subfigure}
\usepackage{hyperref}
\usepackage{graphicx}
\usepackage{caption}
\usepackage{adjustbox}
\usepackage{mathtools}
\usepackage{amsmath}
\usepackage{paralist}

\usepackage{booktabs}
\usepackage{xcolor}
\usepackage{stfloats}
\usepackage{pifont}
\usepackage{mdframed}
\usepackage{enumitem}
\usepackage{url}
\usepackage{wrapfig}
\usepackage{ulem}

\renewcommand\footnotetextcopyrightpermission[1]{}

\newcommand{\revise}[1]{{{#1}}}
\newcommand{\minor}[1]{{{#1}}}

\newcommand{\finding}[1]{
  \vspace{2.3mm}
 \begin{mdframed}[linecolor=gray,roundcorner=12pt,backgroundcolor=gray!15,linewidth=3pt,innerleftmargin=2pt, leftmargin=0cm,rightmargin=0cm,topline=false,bottomline=false,rightline = false]
 #1
 \end{mdframed}
}

\lstdefinestyle{mystyle}{
    numberstyle=\tiny,
    basicstyle=\ttfamily\footnotesize,
    breakatwhitespace=false,         
    breaklines=true,                 
    captionpos=b,                    
    keepspaces=true,                 
    numbers=left,                    
    numbersep=5pt,                  
    showspaces=false,                
    showstringspaces=false,
    showtabs=false,                  
    tabsize=2,
    frame={bottomline}
}
\setcopyright{acmcopyright}
\copyrightyear{2018}
\acmYear{2018}
\acmDOI{XXXXXXX.XXXXXXX}

\begin{document}

\title{Anatomizing Deep Learning Inference in Web Browsers}


\author{Qipeng Wang}\authornote{Qipeng Wang's work was conducted during an internship at Microsoft Research Asia.}
\affiliation{
  \institution{Peking University}
  \city{Beijing}
  \country{China}}
\email{wangqipeng@stu.pku.edu.cn}

\author{Shiqi Jiang}\authornotemark[2]
\affiliation{
  \institution{Microsoft Research Asia}
  \city{Beijing}
  \country{China}}
\email{shijiang@microsoft.com}

\author{Zhenpeng Chen}\authornotemark[2]
\affiliation{
  \institution{Nanyang Technological University}
  \country{Singapore}}
\email{zhenpeng.chen@ntu.edu.sg}

\author{Xu Cao}
\affiliation{
  \institution{Microsoft Research Asia}
  \city{Beijing}
  \country{China}}
\email{caox@microsoft.com}

\author{Yuanchun Li}
\affiliation{
  \institution{Institute for AI Industry Research (AIR), Tsinghua University}
  \city{Beijing}
  \country{China}}
\email{liyuanchun@air.tsinghua.edu.cn}

\author{Aoyu Li}
\affiliation{
  \institution{Nanyang Technological University}
  \country{Singapore}}
\email{aoyuli2000@outlook.com}

\author{Yun Ma}\authornote{Corresponding authors}
\affiliation{
  \institution{Peking University}
  \city{Beijing}
  \country{China}}
\email{mayun@pku.edu.cn}

\author{Ting Cao}
\affiliation{
  \institution{Microsoft Research Asia}
  \city{Beijing}
  \country{China}}
\email{ticao@microsoft.com}

\author{Xuanzhe Liu}
\affiliation{
  \institution{Peking University}
  \city{Beijing}
  \country{China}}
\email{liuxuanzhe@pku.edu.cn}

\renewcommand{\shortauthors}{Wang et al.}
\renewcommand{\sout}[1]{}
\newcommand{\para}[1]{\smallskip\noindent{\bf {#1}. }}


\begin{abstract}
Web applications have increasingly adopted Deep Learning (DL) through \textit{in-browser inference}, wherein DL inference performs directly within Web browsers. The actual performance of in-browser inference and its impacts on the quality of experience (\textit{QoE}) remain unexplored, and urgently require new QoE measurements beyond traditional ones, e.g., mainly focusing on page load time. To bridge this gap, we make the first comprehensive performance measurement of in-browser inference to date. Our approach proposes new metrics to measure in-browser inference: responsiveness, smoothness, and inference accuracy. 
Our extensive analysis involves 9 representative DL models across Web browsers of 50 popular PC devices \minor{and 20 mobile devices}.
The results reveal that in-browser inference exhibits a substantial latency gap, averaging 16.9 times slower on CPU and 4.9 times slower on GPU compared to native inference \minor{on PC devices. The gap on mobile CPU and mobile GPU is 15.8 times and 7.8 times, respectively}.
Furthermore, we identify contributing factors to such latency gap, including underutilized hardware instruction sets, inherent overhead in the runtime environment, resource contention within the browser, and inefficiencies in software libraries and GPU abstractions. 
Additionally, in-browser inference imposes significant memory demands, at times exceeding 334.6 times the size of the DL models themselves, partly attributable to suboptimal memory management. 
We also observe that in-browser inference leads to a significant 67.2\% increase in the time it takes for GUI components to render within Web browsers, significantly affecting the overall user QoE of Web applications reliant on this technology.

\end{abstract}



\maketitle

\section{Introduction}
\label{sec:intro}
Deep Learning (DL) has been increasingly adopted in various Web applications by deploying DL models in Web browsers for inference tasks, i.e., \textit{in-browser inference}~\cite{ma2019moving, web-llm, facetest, aige, office365-grammar-check, teams-bkgnd}. In this context, DL frameworks and models are encapsulated and delivered within Web pages without additional installations on user devices, and DL inference is performed directly in Web browsers~\cite{ma2019moving}, effectively streamlining user access to DL-powered functionalities. \minor{
    Many Web sites have already deployed in-browser inference. For example, online Office 365~\cite{office365-grammar-check} utilizes in-browser inference to assist with the online editing of documents and slides; online Teams blurs the background of people for video meetings~\cite{teams-bkgnd}; Facetest~\cite{facetest}, Aige~\cite{aige}, and Theremix~\cite{theremix} apply in-browser inference for tasks including image classification, face recognition, and music generation.
}

The proliferation of DL-powered Web applications makes in-browser inference a critical topic in various research communities, including Software Engineering (SE)~\cite{chen2022deepperform, gao2020estimating, guo2019empirical, cao2022understanding, liu2022modeling, quan2022towards}, Web~\cite{ma2019moving, zhang2022comprehensive}, and Mobile Computing~\cite{tian2022parallelizing, huang2021integrated}.
Firstly, the performance of in-browser inference is crucial to Web-based software applications. For instance, understanding the detailed performance of in-browser inference can assist software developers in refining the applications, thereby offering users a smoother and better service~\cite{zhang2019empirical, yang2022survey}. Moreover, the Quality of Experience (QoE) is pivotal in Web contexts~\cite{upadhyaya2015quality}, as it reflects the user’s experience while browsing Web pages. Improved QoE can attract a larger user base, making it vital for the development and deployment of software applications.

Existing research efforts primarily focus on DL inference performance in native\footnote{We use ``native'' to refer to the non-browser environment, including servers, PCs, and mobile devices}~\cite{chen2022deepperform, gao2020estimating, guo2019empirical, cao2022understanding, liu2022modeling, zhang2022comprehensive, tang2023lut}, regarding inference latency, memory footprint, etc. For instance, Tang et al.~\cite{tang2023lut} accelerate DL model inference in PCs and mobile devices through a lookup table; Gao et al.~\cite{gao2020estimating} estimate the GPU memory footprint during DL model training in the server.
\revise{
    However, in the scenario of in-browser inference, the
}
browser acts as an intermediary, isolating Web applications from the underlying operating systems and restricting the application's ability to fully harness hardware capabilities. This creates a distinction between the browser environment and the native environment. 
\revise{
    \sout{
        Therefore, there is still a notable knowledge gap about DL in-browser inference performance in the literature.
    }
    The current work demonstrates an insufficient exploration of in-browser inference and there are significant differences between browser and native environments, e.g., the absence of some advanced single instruction multiple data (SIMD) instructions. These factors contribute to a notable knowledge gap in the literature regarding in-browser inference performance. This gap underscores the need for more focused research to better understand and optimize the performance of in-browser inference.
}
Moreover, existing Web QoE measurement efforts primarily focus on the loading phase of Web pages such as page load time~\cite{speedindex, bocchi2016measuring, tian2019understanding, PLT, ling2018linca, yeo2019snapshot}. For example, Speed Index~\cite{speedindex} measures how quickly the Web page is visually displayed during page loading.
To the best of our knowledge, no previous work has considered the impact of in-browser inference on the QoE.

To fill the knowledge gap, we present the first comprehensive measurement study of in-browser inference from two aspects: inference performance and QoE. Our extensive study involves 9 representative DL models across Web browsers of 50 popular PC devices \minor{and 20 mobile devices}. Furthermore, we use the most widely adopted DL frameworks for deploying DL models in browsers, including TensorFlow.js~\cite{tfjs}, ONNX Web Runtime~\cite{ortjs}.

For \textbf{inference performance}, we consider two critical metrics: \textit{latency} and \textit{memory footprint}. These metrics provide insights into the time overhead and memory requirements, respectively. We present a deep analysis of model-level/kernel-level latency and memory footprint, where a kernel is the basic unit of a DL model, such as a convolution operation. Furthermore, the inherent variability in device capacities motivates us to analyze the performance variance of in-browser inference across devices.

We observe a significant disparity in prediction latency between in-browser and native inference, averaging 16.9$\times$ on CPU and 30.6$\times$ on GPU \minor{on PC devices, and the gap on mobile CPU and mobile GPU is 15.8$\times$ and 7.8$\times$, respectively}. In the CPU case, this disparity arises primarily from the underutilization of SIMD instructions \minor{on PC devices}, WebAssembly (Wasm) runtime-inherent inefficiency, as well as competition for resources with inherent browser tasks. On the GPU side, the inefficiency of software libraries on the Web, \revise{i.e.,} WebGL~\cite{webgl-shader}, compared to native CUDA~\cite{cuda}, contributes to the gap. Furthermore, our analysis reveals substantial performance variations across different devices, with disparities reaching up to 28.4$\times$ on CPU and 19.4$\times$ on GPU. These variations stem from the diverse capacities of devices and the hardware affinities of pre-designed DL kernels. Additionally, we identify ongoing concerns regarding memory footprint during in-browser inference. In certain scenarios, the memory footprint can escalate to 334.6$\times$ the model size during super-resolution model inference. These findings underscore the inefficiency of the in-browser inference on memory management when the intermediate data size increases.

For \textbf{QoE}, we are the first to measure it within the context of in-browser inference, proposing new metrics dedicated specifically to this endeavor beyond existing QoE metrics. 
Specifically, we quantitatively measure in-browser inference QoE from three aspects: \textit{responsiveness}, \textit{smoothness}, and \textit{inference accuracy}. Responsiveness measures the speed at which user requests are successfully responded to. Smoothness evaluates the fluidity and seamless rendering of Web page content. Inference accuracy gauges the accuracy of DL models within Web applications.
Our findings reveal an interesting observation: reducing inference latency does not necessarily equate to an improvement in QoE. For instance, while WebGL-based inference generally boasts faster execution time compared to Wasm, it also results in significant degradation of up to 62.7\% in smoothness due to resource conflicts arising from the competition between inference and the web application's reliance on GPU resources.


Based on the results, we provide a series of actionable implications for various stakeholders. 
\textbf{(1)} \textbf{Web browser vendors} can extend support for Wasm SIMD and streamline its integration with in-browser inference frameworks to avert potential performance degradation. 
\textbf{(2)} \textbf{Web application developers} can enhance the QoE by configuring inference to utilize the Wasm backend, albeit with a slight increase in inference latency. 
\textbf{(3)} \textbf{DL framework vendors} can employ efficient memory management techniques, such as model information-guided memory pool, to mitigate the memory overhead during inference. 

\revise{
    \sout{
        As an additional contribution, we will publicly release our code and data$^2$ to facilitate the replication and extension of this work.
    }
    To summarize, this paper makes the following contributions:
    \begin{itemize}
        \item We conducted a detailed and granular analysis of in-browser inference performance at both the model and kernel levels across multiple devices, and examined the performance gap and underlying reasons compared to native inference. 
        \item We proposed specific Quality of Experience (QoE) metrics tailored for in-browser inference scenarios and conducted quantitative analyses to investigate the impact of in-browser inference on QoE. 
        \item We provided recommendations for browser vendors, inference framework developers, and Web application developers to help them enhance the performance of in-browser inference and improve user QoE.
        \item As an additional contribution, we publicly released our code and data\footnote{https://github.com/qipengwang/InBrowserInference} to facilitate the replication and extension of this work.
    \end{itemize}
}

\section{Background}
\label{sec:background}

\subsection{In-browser Inference}
Web browsers have emerged as a pivotal platform for the widespread adoption of DL technologies~\cite{chen2020comprehensive, ma2019moving}. 
\minor{
    \sout{
        DL models find applications across various Web applications, enabling in-browser inference tasks that encompass a range of capabilities such as grammar checking for online document~\cite{office365-grammar-check}, background blurring for virtual meetings~\cite{teams-bkgnd}, extended reality~\cite{bi2023demystifying}, and enhancing online chat interactions~\cite{web-llm}.
    }
    Different from native applications that require installation and upgrade, raising barriers for end-users, and hindering AI service deployments, Web browsers, however, are promising platforms for DL adoption, without the requirement for extra installation. Besides, in-browser inference does not need to send the user data to the remote server, preserving user privacy. 
    In-browser inference has attracted tremendous research attention~\cite{chen2022deepperform, gao2020estimating, guo2019empirical, cao2022understanding, liu2022modeling, quan2022towards, ma2019moving, zhang2022comprehensive, tian2022parallelizing, huang2021integrated, jia2024empowering, dong2023webinf}.
    In-browser inference empowers many new abilities of Web applications. For example, Web Office 365 integrates grammar checking and typing completion~\cite{office365-grammar-check}; online Teams~\cite{teams-online} use in-browser inference to analyze facial contours for online meetings; Facetest~\cite{facetest} applies in-browser inference for face detection; Aige~\cite{aige} applies in-browser inference to guess the age of users according to face image; Theremix~\cite{theremix} utilizes in-browser inference for music generation according to users' control. Many existing Web applications can benefit from in-browser inference. For instance, online subtitle generation will be enabled for living applications; Web augmented reality, virtual reality, and mixed reality may be more vivid when leveraging DL.
}

\begin{figure}[t]
    \centering
    \includegraphics[width=0.75\textwidth]{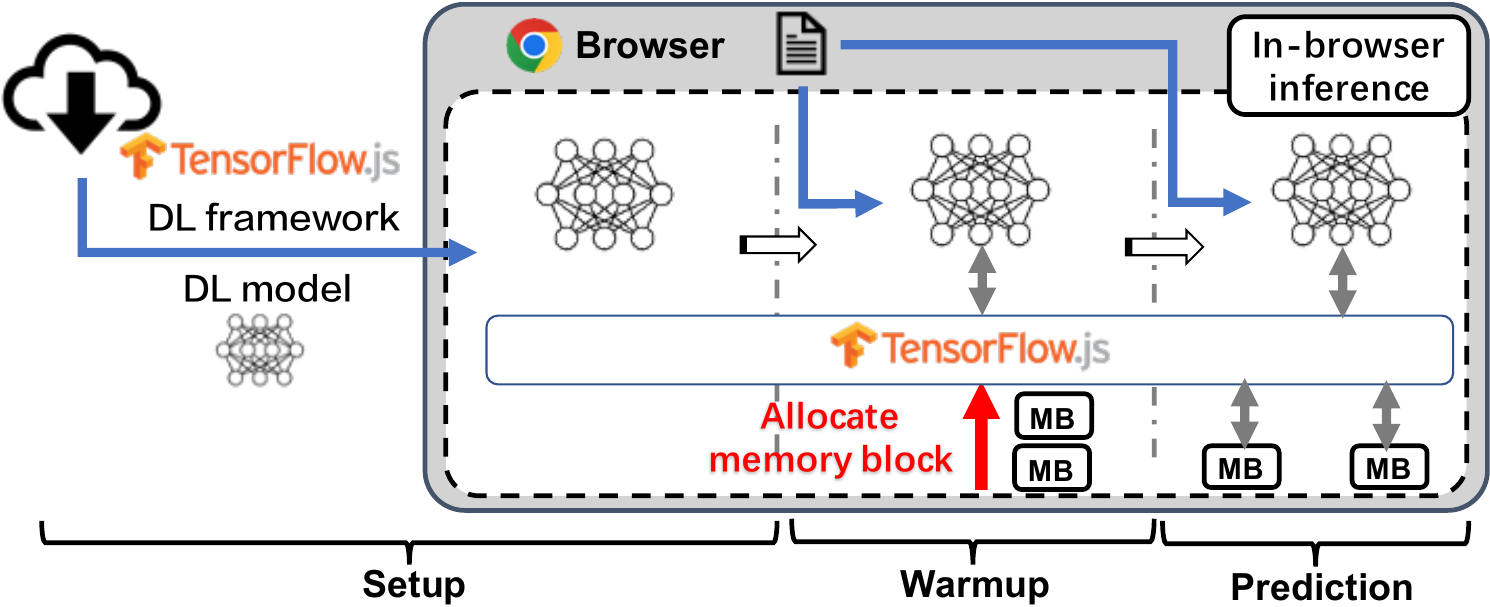}
    \vspace{-3mm}
    \caption{Workflow of in-browser inference. ``MB'' denotes memory block.} 
    \label{fig:scope}
    \vspace{-6mm}
\end{figure}

In-browser inference is the process where Web applications execute DL inference directly within Web browsers. Figure~\ref{fig:scope} illustrates the workflow of this process, which comprises three primary stages: setup, warmup, and prediction. During the setup stage, Web applications download the necessary DL models and DL frameworks and then load them into Web browsers. In the warmup stage, browsers, assisted by DL frameworks, allocate the required memory for inference and initiate the first DL inference operation. Finally, in the prediction stage, DL models carry out subsequent inference tasks using the capabilities provided by DL frameworks.
\revise{
    The warmup and prediction stages in the native environment are identical to those in the Web environment, while the setup stage differs between the two environments. In the native environment, the setup stage merely requires loading the model and framework from local storage, eliminating the need to download them from the server.
}

\subsection{Frameworks and Backends}
Dedicated DL frameworks are developed to facilitate DL deployment in Web applications and in-browser inference. 
Among them, TensorFlow.js (TF.js)~\cite{tfjs} and ONNX Web Runtime (ORT.js)~\cite{ortjs} are actively maintained and have garnered the most GitHub stars. TF.js is an offshoot of  TensorFlow (TF)~\cite{abadi2016tensorflow} and maintains compatibility with it, while ORT.js is derived from ONNX Runtime (ORT)~\cite{ortjs, onnx}. In this study, we consider both of these frameworks due to their widespread adoption for in-browser inference in real-world Web applications~\cite{ma2019moving, guo2019empirical, chen2020comprehensive}.



Both TF.js and ORT.js support using backends for inference acceleration, including WebAssembly (Wasm)~\cite{haas2017bringing, wasm} on CPU and WebGL~\cite{webgl} on GPU.
Wasm serves as a binary instruction format for a stack-based virtual machine \revise{(VM)} within Web browsers. It serves the purpose of a portable compilation target for programming languages, facilitating their deployment on the Web. Wasm stands out for its efficiency, owing to its compact size and swift loading capabilities~\cite{wasm}. 
In Wasm, Single-Instruction-Multiple-Data (SIMD) and multithreading are available for computation efficiency. SIMD is a hardware architecture that processes multiple data elements in one instruction, greatly enhancing computational speed. Multithreading is a technique that allows a program to execute multiple tasks concurrently, making efficient use of system resources for improving performance.
WebGL is a JavaScript library designed for rendering graphics directly within Web browsers, implementing an API that adheres to the OpenGL standards~\cite{opengl}. Developers can harness the power of WebGL by crafting WebGL shaders~\cite{webgl-shader} to expedite parallel computation processes.

Recently, a new backend called WebGPU has been introduced~\cite{webgpu-w3c}. 
Nevertheless, the WebGPU standard has yet to attain official release status. 
WebGPU is still in its early and continually evolving phase, with no support for general models but only limited ones in TF.js. Therefore, we opt not to incorporate WebGPU into this paper. 

\section{Methodology}

\subsection{Experimental Setup}
\label{subsec:experimental_setup}
We start by outlining our choices for DL frameworks, backends, DL models, PC devices, and Web browsers.

\noindent \textbf{Selection of DL frameworks and backends.}  For in-browser inference, we opt for TF.js 4.2.0 and ORT.js 1.14.0 as our DL frameworks. Our selected backends include Wasm and WebGL, while we omit the pure JavaScript and WebGPU backends due to the lack of support for these options in ORT.js.

\noindent \textbf{Selection of DL models.} We select nine representative DL models for study, covering three common types of Web applications, i.e., image classification, object detection, and grammar checking. For each type, we select three widely adopted DL models. Specifically, for image classification, we select MobileNetV2 (denoted as M1; 14MB in size)~\cite{mobilenetv22018}, ResNet50 (denoted as M2; 98MB in size)~\cite{he2016deep}, and VGG16 (denoted as M3; 528MB in size)~\cite{simonyan2014very}; for object detection, we select SSD-MobileNetV2 (denoted as M4; 26MB in size)~\cite{liu2016ssd}, EfficientDet (denoted as M5; 23MB in size)~\cite{tan2020efficientdet}, and YoloV5-Middle (denoted as M6; 83MB in size)~\cite{redmon2016you}; for grammar checking, we select Bert-base (denoted as M7; 104MB in size)~\cite{ahmad2019reqa}, MobileBert (denoted as M8; 95MB in size)~\cite{sun2020mobilebert}, and small Bert (denoted as M9; 17MB in size)~\cite{devlin2018bert}. We use the pre-trained models~\cite{tfhub, tf-model-api} and convert them to the desired target format using official model converters~\cite{tfjs-convertor, onnx-convertor}.

\noindent \textbf{Selection of \revise{\sout{PC}} devices and Web browsers.} Our measurement \revise{mainly} involves 50 different PC devices, featuring a diverse array of CPU models, including both Intel CPUs and Apple Silicon. Within the Intel CPU category, we encompass core-i5 and core-i7 processors spanning from Skylake to Rocket Lake architectures~\cite{skylake, rocketlake}. In the Apple Silicon category, we consider M1 and M2 CPU models. Moreover, these devices are equipped with a range of GPU models, covering both integrated and discrete graphics solutions. The discrete GPU options encompass Nvidia models such as GTX 980, 1060, and 2060, while integrated GPUs are embedded within the CPU and include variants such as Intel HD 530, 630, and 645, among others. For the Web browser, we employ Google Chrome, which stands as the most widely used browser, commanding a global market share of more than 63.5\% as of September 2023~\cite{browser-market-share}.
\revise{
    To explore the difference between different browsers, we consider Firefox to be another Web browser platform. According to browser market share data~\cite{browser-market-share}, Firefox holds the fourth largest share. We excluded Safari because it is only available on Apple devices, and we excluded Edge because it is also a Chromium-based browser, the same as Chrome.
    To explore the performance of in-browser inference on mobile devices, we involve 20 mobile devices. The mobile devices also feature a wide spectrum of SoC models, including both Snapdragon SoC models and Kirin SoC models. The Snapdragon SoC spans from Snapdragon 855 to Snapdragon 8 Gen 3; The Kirin SoC spans from Kirin 980 to Kirin 9000. On these mobile devices, we use Chrome as the experiment Web browser platform.
}





\subsection{Measurement of Inference Performance}
First, we measure the performance of in-browser inference from two aspects: \textit{latency} and \textit{memory footprint}.

\revise{
    \noindent\textbf{Experiment steps.} In the experiment, we implemented the following procedures: \textbf{(1)} Kernel-level\footnote{\revise{A DL model can be represented as a series of interconnected kernels, such as matrix multiply and convolution. The interconnections of these kernels form a computation graph.}} profiling enabling. We modified the inference framework to enable profiling kernel-level latency and memory footprint. \textbf{(2)} Web service deployment. We deployed a Web service that supports inference using various models through TF.js or ORT.js frameworks. \textbf{(3)} Model inference on devices. Devices accessed the service to perform model inference, during which we recorded the latency of each model inference. \textbf{(4)} We repeated step (3) multiple times and calculated the average latency and memory footprint on each device and browser, as well as the latency variance.
}

\noindent \textbf{Latency.} We measure the latency in both \textit{model-level} and \textit{kernel-level}. 
In terms of latency, the total inference latency consists of three types: setup latency, warmup latency, and prediction latency.
\textbf{(1)} \textit{Model-level latency}. 
We analyze the average latency gap and latency distribution among different DL frameworks, backends, and PC devices, and analyze the root cause.
There are different acceleration techniques in the Wasm backend, i.e., SIMD and multithreading. The two DL frameworks that we adopt both support them for acceleration. SIMD processes multiple data within a single instruction and multithreading performs computation in parallel.
To explore to what extent each acceleration technique reduces the latency, we measure the latency under different settings, i.e., whether each technique is enabled or not.
\textbf{(2)} \textit{Kernel-level latency}.
\revise{\sout{A DL model can be represented as a series of interconnected kernels, such as matrix multiply and convolution.}}
Understanding the performance of different kernels can help developers design kernel-specific optimization to improve the inference performance.
For kernel-level latency, we measure the average kernel latency proportion on different backends. We do not consider the setup stage in this part because this stage does not involve kernel execution. In addition, we carry out a breakdown analysis to investigate how much SIMD and multithreading contribute to the acceleration in the Wasm backend.

\noindent \textbf{Memory footprint.} We next measure the memory footprint, i.e., the memory usage associated with in-browser inference. Memory is a critical metric both for the inference process~\cite{wadhwani2022squeezenerf, molchanov2016pruning} and Web applications as a whole~\cite{qazi2020mobile}. To gain deep insights into the memory footprint of in-browser inference, we measure the memory footprint of each kernel and analyze the memory footprint growth. We omit the setup stage from our analysis since the inference execution does not take place during this phase. Similarly, we exclude the prediction stage from consideration because the DL framework retains allocated memory for operational efficiency, resulting in a constant memory footprint throughout this phase. As a result, we measure the memory footprint of only the warmup stage.

\subsection{Definition of QoE Metrics Dedicated to In-browser Inference}
Existing Web QoE metrics primarily focus on evaluating the speed at which a Web page becomes visually accessible during the page loading process~\cite{speedindex, PLT, TTFB}. In contrast, the QoE associated with in-browser inference extends its influence across the entirety of a user's browsing experience.
Thus, we propose three new QoE metrics specialized for in-browser inference (including responsiveness, smoothness, and inference accuracy), and quantify it during model inference. For QoE measurement, we do not differentiate the three stages described in Figure \ref{fig:scope}, because QoE is an average metric that spans the entire Web application usage. 

\begin{align}
RunsPerMinute = \frac{\#Testcase}{Time_{Benchmark}} \label{responsiveness-formular}
\end{align}
\begin{align}
fps = \frac{\#Frames_{Rendered}}{Time_{Rendering}} \label{smoothness-formular}
\end{align}
\begin{align}
&InfAcc =  \begin{cases}
    	\frac{\#Frames_{correctlyClassified}}{\#Frames_{total}}, &classification   \\
    	\frac{1}{N} \sum_{i=1}^{N} \int_{0}^{1} PR_i,  &object\ detection	
        \end{cases}\label{inference-accuracy-formular}
        \\
&where\ PR_i\ is\ Precision\mbox{-}Recall\ curve\ of\ class\ i,\ and\ N\ is\ \#class.\notag
\end{align}

\noindent \textbf{Responsiveness.} Users prefer Web pages that respond to user requests quickly. The responsiveness reflects how quickly a Web page responds to user requests. It is quantified by the response time to user requests. We measure the responsiveness of Web pages based on Speedometer~\cite{speedometer-site}, a benchmark for Web application responsiveness. 
\revise{
    \sout{Speedometer simulates various Web activities, including 480 different tasks. }
    Given the diverse application scenarios in the Web environment, it is difficult to have metrics generalize to all models or use cases. To address this challenge, the WebKit team officially provides the Speedometer dataset that covers various use cases as a benchmark. Specifically, this benchmark encompasses multiple high-level user journeys, including editing rich text, rendering charts, and reading news sites, with each journey further comprising multiple workloads for testing. In this benchmark, ``runs per minute'' is suggested to reflect the responsiveness QoE.
}
As defined in Eq.\ref{responsiveness-formular}, we use the score ``runs per minute'' to quantify the responsiveness. The score represents the number of benchmark tests completed per unit time.
We inject the inference code into the Speedometer Web page \revise{through the ``Selenium'' library, a library for automated Web application development and testing, allowing the developer to execute any JavaScript programs on any Web page. We \sout{and}} perform in-browser inference during the benchmark test \revise{intermittently, to avoid the benchmark test being blocked if the model inference is performed continually. During the test, model inference accounts for 50\% of the total duration}.

\noindent \textbf{Smoothness.} Users prefer Web pages that render smoothly without any lag. The smoothness reflects how timely the GUI content is rendered. 
It is quantified by the rendering/video frames per second (fps). As defined in Eq.\ref{smoothness-formular}, we measure the fps of playing video and rendering Web pages. The fps stands for the average number of rendered frames per unit time during the entire testing phase.
For vision tasks, we select YouTube as the target Web page for measurement; for the text task, we select Google Docs.
The two Websites are among the most popular Websites for online videos and documents. YouTube provides continuous video streams and we can input text continuously into Google Docs.
We measure the video fps in YouTube and render fps in Google Docs.
\revise{
    We used the first recommended playback page on the YouTube homepage. Regarding Google Docs experiments, we use different input texts on the devices at varying speeds ranging from 20 to 60 words per minute.
}
We inject the inference code into each Web page. We run the inference for 10 minutes while using the Web page and use the averaged results as the final obtained fps.

\noindent \textbf{Inference accuracy.} Due to insufficient computation resource, inference accuracy might degrade inevitably. For example, video frames may be skipped and the application reuses previous results. Thus, as defined in Eq.\ref{inference-accuracy-formular}, we use inference accuracy, such as classification accuracy and detection mean average precision (mAP), to quantify the QoE. 
We use vision tasks, i.e., image classification and object detection, as the targets. We do not consider grammar checking in this part because, to our best knowledge, there lacks a benchmark dataset that contains both text and continuous timestamps. 
We use two benchmark datasets, including the YouTube bounding boxes dataset~\cite{youtube-bb} for image classification and the YouTube Face dataset~\cite{youtube-face} for object detection. \revise{
    \sout{
        Both datasets contain continuous video frames and ground truth, as well as timestamps. 
    }
    The two datasets are widely used~\cite{fan2019lasot, liu2023objects, cai2019exploring, wang2018cosface, pfister2015flowing, schroff2015facenet} because they both contain continuous real-world video stream frames and corresponding timestamps.
}

\revise{
    \sout{
        Note that frames can use the previous inference results if they are not processed by DL models~\cite{kang2017noscope, xu2020approximate}.
        We use the same models as before. We simulate in-browser inference with the fps obtained in the previous experiments.
    }
    \noindent \textbf{Experiment steps.} In our experiment, we executed the following procedures: \textbf{(1)} Code injection using Selenium. We utilized the Selenium~\cite{selenium} library to inject model inference code into various Web pages. \textbf{(2)} Model inference and QoE monitoring. After the Web pages were fully loaded, we used Selenium to initiate the model inference functions. The inference process was performed for a period, during which we recorded responsiveness and smoothness QoE. We enable multithreading and set 4 threads. \textbf{(3)} Continuous in-browser inference simulation. We simulated continuous in-browser inference using the two datasets and the fps values obtained from the smoothness experiment. If a video frame could be processed using the model, the inference results of that frame were used. If a frame could not be processed due to low fps or high resource usage, the results from the previous frame were reused. This approach is quite common in the field of video processing~\cite{kang2017noscope, xu2020approximate}. We calculate the inference accuracy after all frames are processed.
}

\section{Results of Inference Performance Measurement}
\label{sec:performance}


\begin{table*}[h]
    \caption{Latency (ms) of different models running in native. The GPU used is the discrete one. \revise{
        The results show that both the TF and ORT frameworks can support the inference of these models. When performing inference on GPU, a 5.0$\times$ average lower prediction latency is achieved, but the setup and warmup latency is higher, being 1.2$\times$ and 14.1$\times$ higher compared to CPU, respectively.
    }}
    \label{tab:e2e-native-all}
    \vspace{-4mm}
    \resizebox{0.95\linewidth}{!}{%
        \begin{tabular}{@{}l|rrrrrr|rrrrrr@{}}
        \toprule[1.5pt]
                                  & \multicolumn{6}{c|}{TensorFlow}                                                                                                                                                               & \multicolumn{6}{c}{ONNX Runtime}                                                                                                                                                            \\ \cmidrule(l){2-13} 
                                  & \multicolumn{3}{c|}{CPU}                                                                      & \multicolumn{3}{c|}{GPU}                                                                      & \multicolumn{3}{c|}{CPU}                                                                       & \multicolumn{3}{c}{GPU}                                                                    \\
                                  & \multicolumn{1}{l}{setup} & \multicolumn{1}{l}{warmup}     & \multicolumn{1}{l|}{prediction}  & \multicolumn{1}{l}{setup} & \multicolumn{1}{l}{warmup}     & \multicolumn{1}{l|}{prediction}  & \multicolumn{1}{l}{setup} & \multicolumn{1}{l}{warmup}     & \multicolumn{1}{l|}{prediction}   & \multicolumn{1}{l}{setup} & \multicolumn{1}{l}{warmup}   & \multicolumn{1}{l}{prediction}  \\ \midrule
        (M1) MobilenetV2          & 1351.5                    & 463.5                          & 17.6                             & 2069.9                    & 2074.8                         & 7.9                              & 140.0                     & 6.8                            & 2.9                               & 1394.2                    & 1271.8                       & 2.8                             \\
        (M2) Resnet50             & 6407.3                    & 997.4                          & 82.2                             & 6223.4                    & 1000.4                         & 11.2                             & 488.0                     & 25.4                           & 23.4                              & 130.7                     & 187.7                        & 5.6                             \\
        (M3) VGG16                & 873.2                     & 2308.4                         & 153.6                            & 3050.8                    & 681.2                          & 3.4                              & 2057.3                    & 90.4                           & 101.8                             & 839.8                     & 126.7                        & 8.2                             \\
        (M4) SSD-MobilenetV2      & 9359.6                    & 2571.7                         & 43.3                             & 10031.3                   & 5128.5                         & 65.0                             & 1245.9                    & 41.7                           & 29.4                              & 1562.8                    & 730.2                        & 65.0                            \\
        (M5) EfficientDet         & 27722.3                   & 4823.6                         & 482.6                            & 27107.2                   & 6522.6                         & 228.1                            & 2056.7                    & 478.7                          & 337.1                             & 2522.4                    & 9124.3                       & 274.4                           \\
        (M6) Yolo5-Middle         & 11719.9                   & 1209.3                         & 84.3                             & 10599.7                   & 1562.8                         & 39.5                             & 517.7                     & 49.8                           & 41.7                              & 249.3                     & 288.4                        & 8.0                             \\
        (M7) Bert-base            & 5153.0                    & 856.4                          & 11.7                             & 4938.3                    & 1196.8                         & 19.1                             & 614.2                     & 39.6                           & 31.0                              & 453.8                     & 18.0                         & 16.0                            \\
        (M8) MobileBert           & 29766.9                   & 3526.9                         & 18.5                             & 30586.4                   & 3269.2                         & 42.5                             & 970.4                     & 77.4                           & 53.2                              & 1215.7                    & 61.8                         & 56.4                            \\
        (M9) Bert-Small           & 2044.5                    & 168.3                          & 3.6                              & 2037.6                    & 109.7                          & 5.0                              & 212.6                     & 4.6                            & 2.2                               & 126.7                     & 6.0                          & 3.8                             \\ \hline
        Average                   & 10488.7                   &1880.6                          &99.7                              &10738.3                    &2394.0                          &46.9                              &922.5                      &90.5                            &69.2                               &943.9                      &1312.8                        &48.9                             \\
        \bottomrule[1.5pt]
        \end{tabular}%
    }
\end{table*}
    
\begin{table*}[h]
    \centering
    \caption{Latency (ms) of supported models in TF.js in Wasm backend. ``-SIMD/thread" indicates enabled technique. \revise{
        We found that the Wasm backend of TF.js has insufficient support for models, supporting only four models. After enabling SIMD and multithreading respectively, the in-browser inference latency decreased by 49.1\% and 10.7\%. When both were enabled, the prediction latency decreased by 64.7\%, demonstrating the potential of both technologies in accelerating inference.
    }}
    \label{tab:e2e-tfjs-wasm}
    \vspace{-4mm}
    \resizebox{0.95\linewidth}{!}{%
    \begin{tabular}{@{}l|rrrrrrrrrrrr@{}}
    \toprule[1.5pt]
                              & \multicolumn{3}{c|}{Wasm}                                                                   & \multicolumn{3}{c|}{Wasm-thread}                                                          & \multicolumn{3}{c|}{Wasm-SIMD}                                                            & \multicolumn{3}{c}{Wasm-thread-SIMD}                                                      \\ \cmidrule(l){2-13} 
                              & \multicolumn{1}{c}{setup} & \multicolumn{1}{c}{warmup} & \multicolumn{1}{c|}{prediction}    & \multicolumn{1}{c}{setup} & \multicolumn{1}{c}{warmup} & \multicolumn{1}{c|}{prediction}  & \multicolumn{1}{c}{setup} & \multicolumn{1}{c}{warmup} & \multicolumn{1}{c|}{prediction}  & \multicolumn{1}{c}{setup} & \multicolumn{1}{c}{warmup} & \multicolumn{1}{c}{prediction}   \\ \midrule
    (M1) MobilenetV2          & 381.5                     & 320.9                      & 235.0                              & 335.2                     & 284.0                      & 222.2                            & 400.3                     & 194.5                      & 116.6                            & 254.3                     & 161.5                      & 89.2                             \\
    (M2) Resnet50             & 2362.6                    & 1984.3                     & 1710.3                             & 2109.7                    & 1724.7                     & 1416.1                           & 2485.1                    & 899.7                      & 539.5                            & 1665.2                    & 647.3                      & 300.8                            \\
    (M3) VGG16                & 10283.0                   & 8303.8                     & 6069.9                             & 10265.0                   & 7597.0                     & 5334.9                           & 10747.3                   & 5482.7                     & 3060.7                           & 9111.0                    & 2402.5                     & 1021.5                           \\
    (M4) SSD-MobilenetV2      & 574.6                     & 1386.6                     & 1160.6                             & 575.2                     & 1174.6                     & 1064.6                           & 608.2                     & 1024.1                     & 833.4                            & 474.8                     & 962.4                      & 795.0                            \\ \hline
    Average                   & 3400.4                    & 2998.9                     & 2293.9                             & 3321.3                    & 2695.1                     & 2009.4                           & 3560.2                    & 1900.2                     & 1137.5                           & 2876.3                    & 1043.4                     & 551.6                            \\
    \bottomrule [1.5pt]
    \end{tabular}%
    }
\end{table*}

\begin{table*}[h]
    \centering
    \caption{Latency (ms) of models in ORT.js Wasm backend. ``-SIMD/thread" indicates enabled technique. \revise{
        We found that the Wasm backend of ORT.js supports all nine models. After SIMD and multithreading technologies were enabled, the prediction latency decreased by 50.7\% and 31.2\%, respectively. When both were enabled, the prediction latency decreased by 63.4\%.
    }}
    \label{tab:e2e-ort-wasm}
    \vspace{-4mm}
    \resizebox{0.95\linewidth}{!}{%
    \begin{tabular}{@{}l|rrrrrrrrrrrr@{}}
    \toprule[1.5pt]
                                & \multicolumn{3}{c|}{Wasm}                                                                 & \multicolumn{3}{c|}{Wasm-thread}                                                          & \multicolumn{3}{c|}{Wasm-SIMD}                                            & \multicolumn{3}{c}{Wasm-thread-SIMD}                                                      \\ \cmidrule(l){2-13} 
                                & \multicolumn{1}{c}{setup} & \multicolumn{1}{c}{warmup} & \multicolumn{1}{c|}{prediction}  & \multicolumn{1}{c}{setup} & \multicolumn{1}{c}{warmup} & \multicolumn{1}{c|}{prediction}  & \multicolumn{1}{c}{setup}  & \multicolumn{1}{c}{warmup} & \multicolumn{1}{c|}{prediction}     & \multicolumn{1}{c}{setup} & \multicolumn{1}{c}{warmup} & \multicolumn{1}{c}{prediction}   \\ \midrule
    (M1) MobilenetV2            & 581.4                     & 148.1                      & 120.6                            & 1306.6                    & 216.9                      & 122.4                            & 1252.1                     & 162.8                      & 106.5                               & 331.0                     & 139.6                      & 105.5                            \\
    (M2) Resnet50               & 539.3                     & 1299.5                     & 1275.3                           & 699.1                     & 395.1                      & 394.7                            & 1412.3                     & 411.9                      & 358.1                               & 1219.3                    & 306.9                      & 172.3                            \\
    (M3) VGG16                  & 7545.3                    & 9024.7                     & 8275.6                           & 4437.5                    & 8147.2                     & 8140.1                           & 2695.2                     & 3164.6                     & 2947.2                              & 2423.2                    & 1353.6                     & 1334.1                           \\
    (M4) SSD-MobilenetV2        & 1120.5                    & 287.6                      & 279.4                            & 1338.1                    & 197.3                      & 123.9                            & 1171.8                     & 174.5                      & 126.1                               & 1467.8                    & 84.5                       & 81.4                             \\
    (M5) EfficientDet           & 1581.2                    & 2665.4                     & 2648.6                           & 1786.4                    & 1251.6                     & 1282.5                           & 1606.1                     & 1245.8                     & 1228.8                              & 1889.6                    & 926.0                      & 852.4                            \\
    (M6) Yolo5-Middle           & 427.6                     & 1105.2                     & 1092.7                           & 450.0                     & 365.9                      & 362.9                            & 408.1                      & 406.2                      & 386.4                               & 457.6                     & 145.8                      & 145.2                            \\
    (M7) Bert-base              & 744.6                     & 915.2                      & 910.2                            & 702.4                     & 285.8                      & 279.6                            & 681.8                      & 246.2                      & 247.2                               & 574.8                     & 92.8                       & 81.6                             \\
    (M8) MobileBert             & 873.5                     & 855.1                      & 849.8                            & 935.8                     & 310.1                      & 291.1                            & 865.0                      & 239.3                      & 236.2                               & 945.0                     & 120.2                      & 113.6                            \\
    (M9) Bert-Small             & 155.2                     & 21.2                       & 21.8                             & 161.3                     & 9.2                        & 7.8                              & 158.1                      & 7.1                        & 6.7                                 & 166.8                     & 4.6                        & 4.7                              \\ \hline
    Average                     & 1507.6                    & 1813.6                     & 1719.3                           & 1313.0                    & 1242.1                     & 1222.8                           & 1138.9                     & 673.2                      & 627.0                               & 1052.8                    & 352.7                      & 321.2                            \\
    \bottomrule[1.5pt]
    \end{tabular}%
    }
\end{table*}
\begin{table*}[h]
    \caption{Latency (ms) of models in WebGL backend. ``-I/-D" denotes integrated/discrete GPU. \revise{
    ``/'' indicates the model is not supported by the framework and backend.
    We found that the two frameworks differ in their support for models on the WebGL backend. Specifically, the WebGL backend of TF.js provides better support for models, supporting six more models than ORT.js. The prediction latency using a discrete GPU is lower than that using an integrated GPU, being 38.9\% and 45.2\% lower for TF.js and ORT.js respectively. However, the warmup latency on GPU is more significant, being 43.5\% higher than the prediction latency for both frameworks and both types of GPU.
    }}
    \label{tab:e2e-webgl}
    \vspace{-4mm}
    \resizebox{0.95\linewidth}{!}{%
    \begin{tabular}{@{}l|rrrrrr|rrrrrr@{}}
\toprule[1.5pt]
                        & \multicolumn{3}{c|}{ORT.js-WebGL-I}                                                           & \multicolumn{3}{c|}{ORT.js-WebGL-D}                                                                      & \multicolumn{3}{c|}{TF.js-WebGL-I}                                                                     & \multicolumn{3}{c}{TF.js-WebGL-D}                                                              \\ \cmidrule(l){2-13} 
                        & \multicolumn{1}{c}{setup} & \multicolumn{1}{c}{warmup}     & \multicolumn{1}{c|}{prediction}  & \multicolumn{1}{c}{setup}          & \multicolumn{1}{c}{warmup}     & \multicolumn{1}{c|}{prediction} & \multicolumn{1}{c}{setup}          & \multicolumn{1}{c}{warmup}     & \multicolumn{1}{c|}{prediction} & \multicolumn{1}{c}{setup} & \multicolumn{1}{c}{warmup}     & \multicolumn{1}{c}{prediction}   \\ \midrule
(M1) MobilenetV2        & 210.7                     & 2094.4                         & 33.1                             & 306.6                              & 2866.8                         & 31.4                            & 209.5                              & 2900.7                         & 28.7                            & 530.9                     & 5751.0                         & 20.1                             \\
(M2) Resnet50           & 720.6                     & 1599.6                         & 183.0                            & 2011.8                             & 2810.6                         & 78.2                            & 1635.7                             & 2814.2                         & 248.3                           & 2482.7                    & 5443.6                         & 79.8                             \\
(M3) VGG16              & 733.5                     & 1793.5                         & 545.6                            & 2086.3                             & 2538.4                         & 146.4                           & 9109.0                             & 3414.5                         & 1748.7                          & 10090.0                   & 4719.0                         & 913.9                            \\
(M4) SSD-MobilenetV2    & /                         & /                              & /                                & /                                  & /                              & /                               & 503.1                              & 7047.6                         & 1039.0                          & 1125.2                    & 9097.9                         & 616.6                            \\
(M5) EfficientDet-D1    & /                         & /                              & /                                & /                                  & /                              & /                               & 734.8                              & 9235.6                         & 2736.1                          & 2542.3                    & 11127.3                        & 1342.7                           \\
(M6) Yolo5-Middle       & /                         & /                              & /                                & /                                  & /                              & /                               & 1739.7                             & 5796.1                         & 99.4                            & 2054.3                    & 10873.6                        & 82.9                             \\
(M7) Bert-base          & /                         & /                              & /                                & /                                  & /                              & /                               & 1902.7                             & 3126.1                         & 682.0                           & 3412.0                    & 3128.2                         & 133.4                            \\
(M8) MobileBert         & /                         & /                              & /                                & /                                  & /                              & /                               & 2837.3                             & 3169.5                         & 943.1                           & 3538.0                    & 2196.4                         & 400.5                            \\
(M9) Bert-Small         & /                         & /                              & /                                & /                                  & /                              & /                               & 664.0                              & 2089.0                         & 41.1                            & 839.3                     & 2273.4                         & 58.3                             \\ \hline 
Average                 & 554.9                     & 1829.2                         & 253.9                            & 1468.2                             & 2738.6                         & 85.3                            & 2148.4                             & 4399.3                         & 840.7                           & 2957.2                    & 6067.8                         & 405.4                            \\
\bottomrule[1.5pt]
\end{tabular}%
}
\end{table*}

\begin{figure*}[t]
    \centering
    \includegraphics[width=0.8\textwidth]{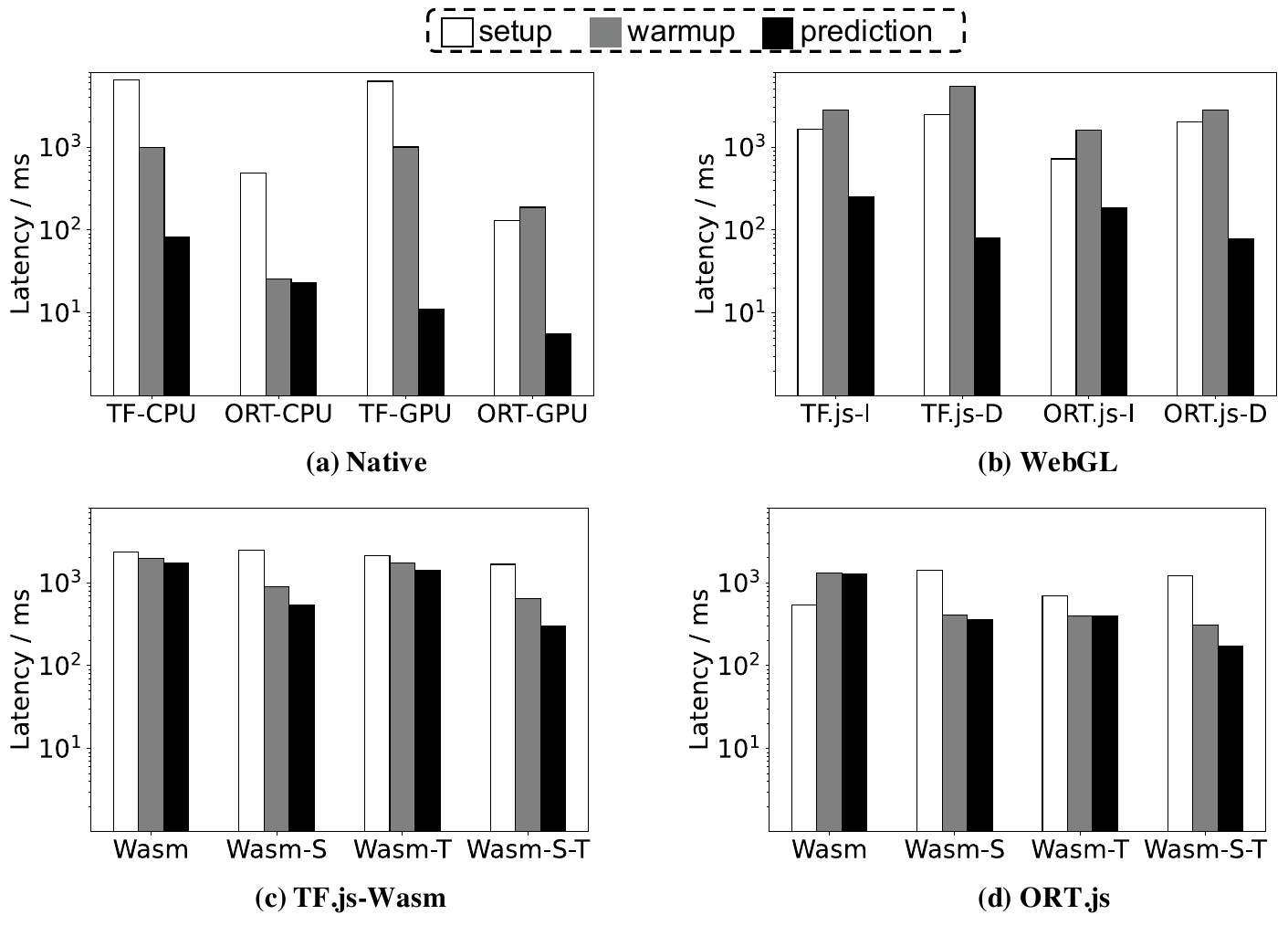}
    \vspace{-4mm}
    \caption{ResNet50 average latency. ``-I/-D'' denotes integrated/discrete GPU. ``-S/-T'' denotes SIMD/multithreading. \revise{
        We found that for ResNet50, both CPU and GPU consistently offer lower native inference prediction latency. Specifically, on CPU, the average native prediction latency is 5.5$\times$ lower on average, while on GPU it is 10.5$\times$. For both in-browser inference and native inference, using GPU achieves lower latency compared to CPU, being 3.0$\times$ and 5.8$\times$, respectively.
    }}
    \vspace{-6mm}
    \label{fig:e2e-resnet50}
\end{figure*}

\subsection{Inference Latency Analysis}\label{subsec:latency-analysis}

\subsubsection{Model-level Latency}\label{subsubsec:model-level-latency-analysis}
First, we compare the average inference latency in Web browsers and native environments. 
\revise{
    We deployed the Web service in a dedicated real-world server, which only serves for our experiments and is located in a LAN with experiment devices. We downloaded the model after the page loading phase because of the considerations that (1) the download time of Web pages may influence the results of the setup stage, and (2) the page load time, which is a critical indicator of the QoE, may be compromised if we download models and the framework during the page loading phase.
}
We illustrate the results of ResNet50 in Figure~\ref{fig:e2e-resnet50} and provide the complete results in Table~\ref{tab:e2e-native-all}, Table~\ref{tab:e2e-tfjs-wasm}, Table~\ref{tab:e2e-ort-wasm}, and Table~\ref{tab:e2e-webgl}.

\textbf{Average CPU latency.} We observe that there is a latency gap between inference in the Wasm backend and in native. We analyze the gap of prediction latency, warmup latency, and setup latency. In this part, both SIMD and multithreading are enabled by default and we set 4 threads.

As for \textit{prediction latency}, the in-browser prediction latency of TF.js is 3.7-18.4$\times$ higher than the latency of TF (Table~\ref{tab:e2e-tfjs-wasm} vs. Table~\ref{tab:e2e-native-all}); the gap is 2.1-36.4$\times$ between ORT.js and ORT (Table~\ref{tab:e2e-ort-wasm} vs. Table~\ref{tab:e2e-native-all}). The average gap of both frameworks is 16.9$\times$.
\revise{
    For both frameworks, the performance of in-browser inference is inferior consistently compared to native inference.
}
The gap mainly comes from three aspects. 
First, advanced SIMD instruction sets are unavailable in browsers, like AVX2 and AVX512 on Intel CPU~\cite{AVX2, AVX512}. The SIMD instruction length in wasm is only 128-bit~\cite{wasm-simd}, while the length is 256-bit in AVX2 and 512-bit in AVX512. This will introduce an up to 4$\times$ latency gap between in-browser and native inference. Besides instruction length, some advanced instructions, such as Fused-Multiply-Add (FMA)~\cite{fma}, are not available in Wasm~\cite{wasm-relaxed-simd-proposal}. The absence of FMA instructions in Wasm could introduce a performance gap of up to 2$\times$.
Second, the Wasm VM introduces additional overhead, because it needs to interpret and execute the Wasm binary code, while native inference frameworks can directly run native binaries. In general, Wasm VM itself can lead up to 2.5$\times$ additional overhead~\cite{jangda2019not}.
Third, there are inherent threads in the Web page that introduce inevitable resource competition, such as the I/O thread and compositor thread~\cite{v8}.

Regarding the two in-browser inference frameworks, there is no consistent superiority of one framework over the other. For instance, we find that ORT.js exhibits lower prediction latency for ResNet50 when employing both multithreading and SIMD, while TF.js exhibits lower latency in inferring VGG16 and MobileNetV2. 
\revise{
    The difference can be primarily attributed to how the two frameworks are implemented. We delved into the implementations of both frameworks and discovered that both frameworks use XNNPACK for the underlying kernel computational, where XNNPACK is a library that provides numerous kernel implementations. However, there are differences in how a computation graph is constructed at a higher level, as well as the kernels themselves. 
    For TF.js, graph parsing and construction, as well as kernel scheduling, are implemented on the JavaScript side, with subsequent calls to XNNPACK~\cite{xnnpack} kernels for execution. In contrast, ORT.js implements these within Wasm. Furthermore, TF.js utilizes models directly converted from TF SavedModels~\cite{saved-model} for loading and graph construction, whereas ORT.js requires a transformation of the SavedModel to adapt it to the specifications of ORT and ORT.js.
    These implementation differences lead to varied performance across different models between the two frameworks. Besides, TF.js shows wider model support compared with ORT.js.
}

As for \textit{warmup latency}, the in-browser warmup latency of TF.js is 0.4-1.0$\times$ of the latency of TF (Table~\ref{tab:e2e-tfjs-wasm} vs. Table~\ref{tab:e2e-native-all}); while the gap is 1.0-20.5$\times$ for ORT.js and ORT (Table~\ref{tab:e2e-ort-wasm} vs. Table~\ref{tab:e2e-native-all}).  
\revise{
    \sout{
        TF suffers from a more severe cold-start issue across all models compared to TF.js. This is because of the large library size of TF (more than 980MB), while the library size of TF.js is only 3MB.
    }
    TF suffers from higher warmup latency across all models compared to TF.js. This is because loading all necessary components, such as the compute graph engine, of TF for model inference requires more than 980MB of random access memory (RAM) size, while the library size of TF.js is only 3MB, indicating that the RAM size will not exceed 3MB. During the warmup stage, the primary tasks include allocating memory for inference and initializing execution.
}
Since memory allocation is not influenced by underlying system software and hardware instruction characteristics, the warmup latency gap between the browser and the native is not as significant as the gap in prediction latency. 
When comparing the warmup latency with the prediction latency, we find that warmup latency is 1.2-2.3$\times$ of the average prediction latency for TF.js (Table~\ref{tab:e2e-tfjs-wasm}); while the gap is 1.0-1.8$\times$ for ORT.js (Table~\ref{tab:e2e-ort-wasm}). Warmup additional overhead primarily arises from memory allocation in Wasm. After the warmup stage, the frameworks maintain the allocated memory for efficiency. The cold start issue is less severe in ORT.js compared to TF.js. This is primarily because ORT.js optimizes the \revise{\sout{model} computation} graph, reducing the frequency of memory allocations. Additionally, ORT.js leverages an Arena memory pool for further optimization~\cite{ort, ortjs}.

As for \textit{setup latency}, the in-browser inference setup latency of TF.js is 0.1-10.4$\times$ of the latency of TF (Table~\ref{tab:e2e-native-all} vs. Table~\ref{tab:e2e-tfjs-wasm}); while it is 0.9-2.4$\times$ for ORT.js and ORT (Table~\ref{tab:e2e-native-all} vs. Table~\ref{tab:e2e-ort-wasm}). It is worth noting that the setup stage of native inference only involves loading the model and the framework.
In-browser inference may outperform native inference w.r.t. the setup latency. One overhead of setup arises from loading the framework into memory. The native inference framework itself has a larger size. For example, simply importing TF in Python \revise{(i.e., executing Python command ``import tensorflow'', which is performed in the setup stage)} requires over 160MB of \revise{RAM size}, while the required files are only about 3MB for TF.js.
On the other hand, native inference may also outperform in-browser inference because in-browser inference needs to download model files to the browser. The download latency depends on the network and model size. For example, in TF.js, downloading VGG16 requires downloading 528MB of data, which introduces high network transfer latency. 
Among the selected models, the model file size varies up to 31.1$\times$. The significant disparity in data transfer latency results in a wide range of setup latency. 

\revise{
    To further explore the impact of network conditions during the setup stage, we deployed a monitor on the server to observe the network latency for each device as it downloaded the model and framework. To avoid competition for network bandwidth among devices, we ensured that only one device was downloading the model at any given time. We observed relatively stable changes in network bandwidth, with the bandwidth for transferring files between different devices ranging from 51.2 to 76.4MB/s. Regarding the latency in file transmission, the latency varied from 212ms to 9717ms.
    We compared this data transfer time with the setup stage duration and found that on average, data transmission latency constituted 88.5\% of the setup latency, indicating that data transmission dominates this stage and is the main bottleneck.
}

\revise{
    \sout{
        To summarize, the setup latency is primarily determined by the sizes of the framework and the model. In most cases, native inference exhibits lower latency compared to inference in browsers.
    }

    In summary, by analyzing data from Table~\ref{tab:e2e-native-all} regarding CPU performance and wasm-thread-SIMD data from Tables~\ref{tab:e2e-tfjs-wasm} and \ref{tab:e2e-ort-wasm}, several key findings emerge. 
    Firstly, for prediction latency, a significant gap persists between in-browser and native inference, primarily due to performance differences between native and browser environments. The lack of advanced SIMD instructions, such as FMA in the Wasm environment, results in lower acceleration compared to native inference, with additional impacts from resource competition among inherent browser threads. 
    Secondly, regarding the warmup latency, TF's substantial library size leads to exceptionally higher latency than TF.js. However, warmup latency is generally higher in all settings than prediction latency, reaching up to 2.3$\times$ higher, primarily due to the need to load and initialize programs and allocate memory during the warmup stage. 
    Lastly, setup latency is primarily influenced by network conditions. In-browser inference involves downloading models from servers, with model data transfer latency constituting a significant portion of setup latency, reaching 88.5\% on average. Despite this, current frameworks support download and load pipelining, which allows parallel operations, enhancing the efficiency of the setup stage.
}

\textbf{Breakdown analysis for Wasm backend.}
We explore the influence of multithreading and SIMD on latency. \revise{
    These two techniques are to accelerate computation, significantly impacting both the warmup and prediction stages.
}
We set 4 threads when enabling multithreading. The results of ResNet50 are illustrated in Figure~\ref{fig:e2e-resnet50} and the full results are presented in Table~\ref{tab:e2e-tfjs-wasm} and Table~\ref{tab:e2e-ort-wasm}. The results show that SIMD reduces 49.1\% and 50.7\% average prediction latency for TF.js and ORT.js, respectively; while multithreading reduces 10.7\% and 31.2\% average prediction latency for TF.js and ORT.js, respectively. When enabling both techniques, the average prediction latency reduces by 64.7\% and 63.4\% for TF.js and ORT.js, respectively. For both frameworks, SIMD provides a greater speedup compared to multithreading. This is mainly because multithreading synchronization is implemented through message passing in Wasm, introducing additional overhead for thread management and synchronization~\cite{webworker, wasm-threads}.
\revise{
    Besides, inherent resource competition from other threads in the browser further slows down the inference, such as the I/O thread and compositor thread in Chromium architecture. For example, for a CPU with four cores, if we have allocated four threads for inference, the remaining threads will compete with the inference threads for resources, thereby impacting the efficiency of the inference process.
}
Both multithreading and SIMD can provide up to 4$\times$ acceleration (since 128-bit instruction length is 4$\times$ the length of int32/float32 data; single thread vs. 4 threads). However, the two technologies only achieve a maximum acceleration of 2.8$\times$ for inference. This is because the model contains many memory-intensive kernels (37.7\% on average across all models), and both frameworks provide limited optimization for these kernels. 

\revise{
    Regarding the warmup latency, it was found that SIMD and multithreading respectively reduce latency by 38.5\% and 12.1\% on TF.js, and by 54.5\% and 41.5\% on ORT.js. These reductions are lower than those observed during the prediction stage, primarily because the warmup stage also involves memory allocation operations, which are less affected by these technologies. However, since computation still dominates during the warmup stage, both technologies are still able to facilitate acceleration.

    In summary, an acute analysis of data from Table~\ref{tab:e2e-tfjs-wasm} and Table~\ref{tab:e2e-ort-wasm} reveals that for in-browser inference, SIMD can provide up to 54\% lower latency on average compared to multithreading. This improvement is mainly due to the efficiency of synchronization between threads and the impact of resource competition. However, the presence of memory-intensive operations limits the maximum acceleration achievable by both technologies to only 2.8$\times$, as both SIMD and multithreading offer limited speed enhancements for memory-intensive operations.
}

\textbf{Average GPU latency.} We analyze the latency in WebGL and compare it with the latency in native and Wasm in the same three latency categories. Note that when comparing with native inference, we only consider discrete GPUs because the native DL framework does not support the integrated one.

As for \textit{prediction latency}, TF.js is 2.5-268.7$\times$ slower than TF, and the gap between ORT.js and ORT is 11.2-17.9$\times$ (Table~\ref{tab:e2e-native-all} vs. Table~\ref{tab:e2e-webgl}). The average gap for both frameworks is 30.6$\times$.
The gap mainly arises from inefficient libraries and GPU abstraction.
Native inference frameworks can utilize high-performance GPU libraries, such as CUDA~\cite{cuda}, to fully exploit the GPU's parallelism capacity. CUDA's low-level access to GPU allows developers to interact directly with the hardware, enabling fine-grained control and optimized program execution. 
However, in-browser inference frameworks rely on WebGL. WebGL is primarily designed for rendering tasks and requires using graphics-related mechanisms like explicit textures and frame buffers. Developers cannot access GPU directly through WebGL, making it inefficient for parallel computing tasks.

When comparing the latency in the WebGL backend with the Wasm backend, we also find that WebGL exhibits lower latency than Wasm. For TF.js, inference in WebGL on integrated and discrete GPUs results in 1.4$\times$ and 2.7$\times$ average prediction latency reduction, respectively, compared to Wasm. For ORT.js, the reduction is 2.2$\times$ and 4.9$\times$, respectively. The speedup contributes to the GPU capacity, despite the limitations of WebGL in fully utilizing the hardware features.

As for \textit{warmup latency}, \revise{
    \sout{
        TF.js is 5.8$\times$ slower than TF on average, and the average gap between ORT.js and ORT is 186.7$\times$ (Table~\ref{tab:e2e-native-all} vs. Table~\ref{tab:e2e-webgl}). The warmup latency of TF.js on discrete GPU is 64.6$\times$ longer than the prediction latency, while the gap is 26.8$\times$ on integrated GPU. The gap for ORT.js is 48.1$\times$ and 25.1$\times$, respectively (Table~\ref{tab:e2e-webgl}). 
    }
    we observed that the warmup latency for in-browser inference is significantly high. Specifically, when compared to native inference, TF.js is on average 5.8$\times$ slower than TF, and the average gap between ORT.js and ORT is 186.7$\times$ (Table~\ref{tab:e2e-native-all} vs. Table~\ref{tab:e2e-webgl}). Additionally, when comparing warmup latency to prediction latency, we found that the warmup latency of TF.js on discrete GPUs is 64.6$\times$ longer than its prediction latency, while on integrated GPUs, the gap is 26.8$\times$. For ORT.js, these gaps are 48.1$\times$ and 25.1$\times$, respectively (as detailed in Table~\ref{tab:e2e-webgl}).
}
In-browser inference exhibits severe warmup latency in WebGL because the shader is compiled in this stage~\cite{webgl-shader}. 
\revise{
    Taking Chrome on the Windows platform as an example, when executing WebGL, Chrome uses ANGLE to compile WebGL shader code into an intermediate representation, such as Direct3D, which is then compiled by the GPU hardware into executable binaries~\cite{ANGLE}. This compilation process is completed during the first execution of WebGL, and the compiled results can be reused in subsequent executions. The compilation process is time-consuming. Therefore, in-browser inference using the WebGL backend experiences high warmup latency.
}
In addition, memory allocation also introduces overhead in this stage.
Integrated GPU outperforms discrete GPU because the inference process needs to invoke the GPU and synchronize the process state~\cite{getSyncParameter, WebGL2RenderingContext} \revise{during the initializing execution of WebGL}, which is slow on discrete GPU.
We also find that ORT.js outperforms TF.js by 1.7$\times$ on integrated GPU and 1.9$\times$ on discrete GPU, because ORT.js optimizes the model via kernel fusion during model conversion, resulting in shorter compilation time.

As for \textit{setup latency}, TF.js is 0.1-3.3$\times$ of TF, and the gap is 0.2-15.4$\times$ for ORT.js and ORT (Table~\ref{tab:e2e-native-all} vs. Table~\ref{tab:e2e-webgl}). The reason is similar to that of Wasm. We also find that for in-browser inference, integrated GPU outperforms discrete GPU consistently. Specifically, the gap is 1.1-3.5$\times$ for TF.js and 1.5-2.8$\times$ for ORT.js (Table~\ref{tab:e2e-webgl}). This is mainly because of high invoking overhead on discrete GPU~\cite{getSyncParameter, WebGL2RenderingContext}. \revise{
    Unlike the invocation during the warmup stage, the invocation in the setup stage primarily involves loading the model parameters and the inference framework onto the GPU.
}

\revise{
    In summary, based on the analysis of data from the GPU part of Table ~\ref{tab:e2e-native-all} and Table ~\ref{tab:e2e-webgl}, we find that using WebGL for inference is a faster solution compared to Wasm. Compared with Wasm, WebGL achieves up to 2.7$\times$ faster prediction than Wasm. However, compared to native inference using CUDA, WebGL exhibits a significantly larger performance gap, up to 268.7$\times$, due to its inherent inefficiencies. Different from Wasm, using WebGL involves in-browser compilation of WebGL shaders during the warmup stage, which results in a severe warmup latency, up to 64.6$\times$ higher than the prediction latency. Regarding the setup stage, invoking a discrete GPU incurs higher latency.
}

\begin{figure*}[t]
    \centering
    \includegraphics[width=0.8\linewidth]{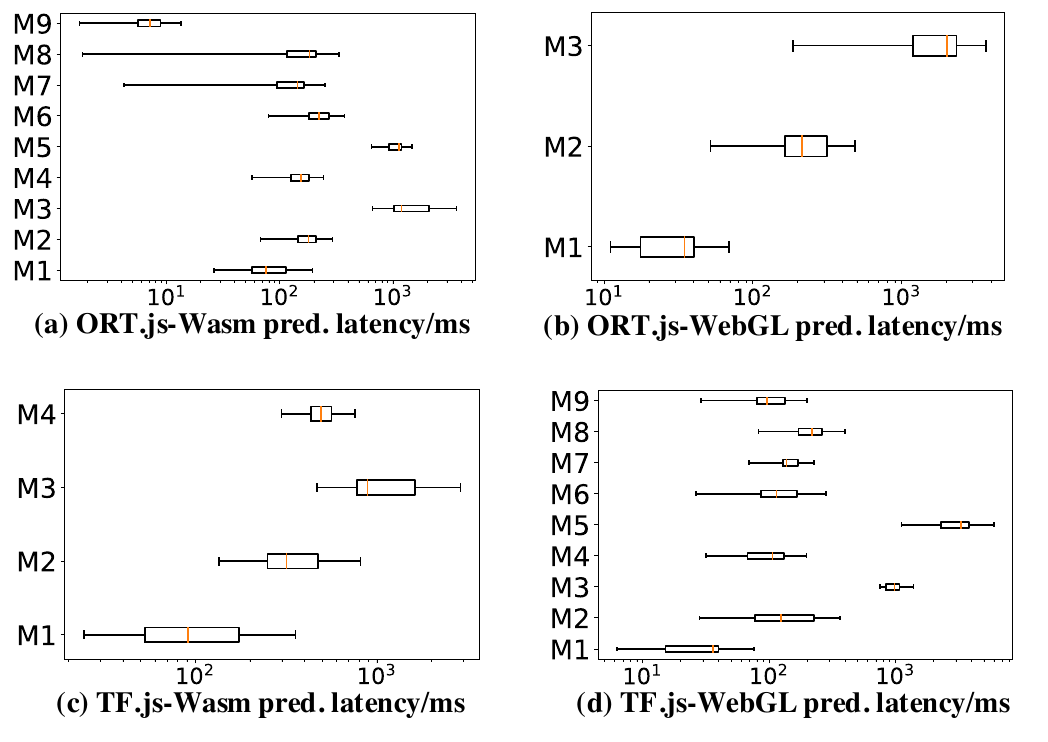}
    \vspace{-3mm}
    \caption{Prediction latency distribution. ``M*'' is model ID and is the same with \S\ref{subsec:experimental_setup}.
    \revise{
        We found that the variance in prediction latency reached 28.4$\times$ on the Wasm backend and 19.4$\times$ on the WebGL backend for all models and both frameworks, primarily due to differences in device hardware performance.
    }}
    \label{fig:inter-latency-distribution}
    \vspace{-5mm}
\end{figure*}

\begin{figure*}[h]
    \centering
    \includegraphics[width=0.8\linewidth]{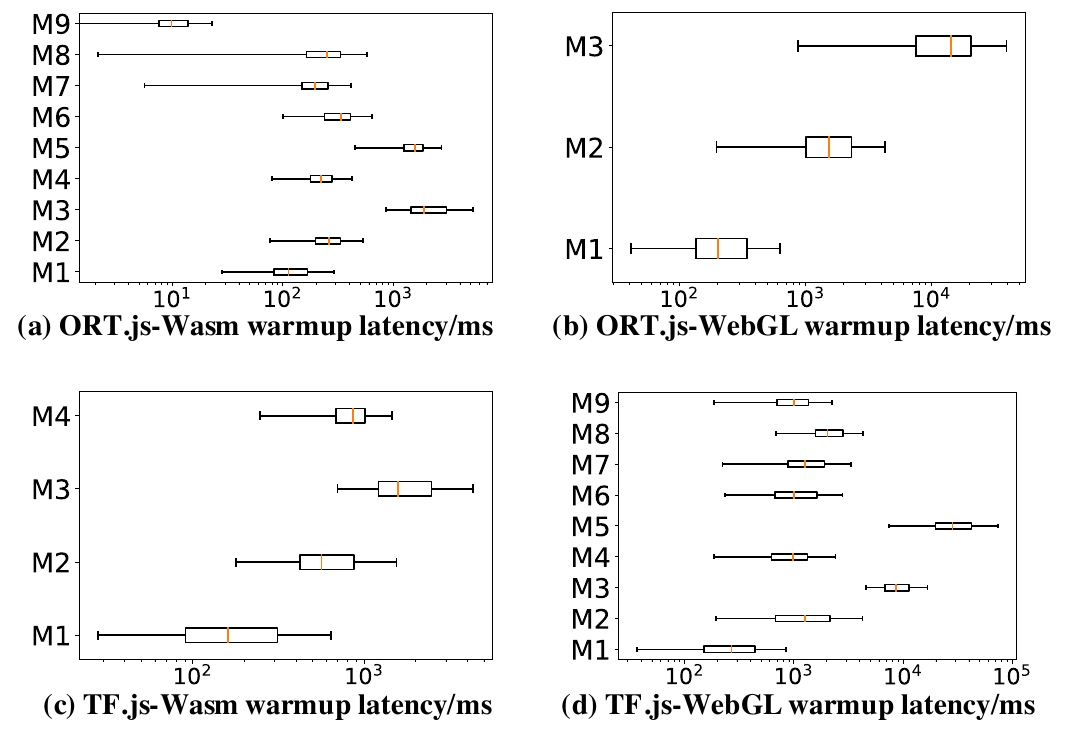}
    \vspace{-2mm}
    \caption{Warmup latency distribution. ``M*'' is model ID and is the same with \S\ref{subsec:experimental_setup}. \revise{
        We found that the variance in warmup latency reached 25.3$\times$ on the Wasm backend and 14.4$\times$ on the WebGL backend for all models and both frameworks.
    }}
    
    \label{fig:warmup-latency-distribution}
\end{figure*}
\textbf{Latency variance.} Figure~\ref{fig:inter-latency-distribution} and Figure~\ref{fig:warmup-latency-distribution} show the prediction latency and warmup latency variance, respectively. 
\revise{
    The variance is the latency gap between the highest and the lowest average latency among devices.
}
We exclude setup latency variance due to the impact of different network conditions.

The \textit{prediction latency} varies differently among devices. Take ResNet50 in the Wasm backend of TF.js as an example, the latency quartiles are 257.6ms, 358.4ms, and 680.9ms. The prediction latency variance of all models in the Wasm backend reaches up to 28.4$\times$. Such variance mainly derives from hardware capacities. We observe that the benchmark score of the lowest-end CPU involved, Core i5 Skylake, is 4.1$\times$ lower than the highest-end CPU, Core i7 Rocket Lake (6,030 vs. 24,673) for one core~\cite{cpu-benchmark}. Moreover, the same kernel can also exhibit different performances on different hardware~\cite{wang2021asymo, chen2018tvm}.
In the WebGL backend in TF.js, the prediction latency quartiles of ResNet50 are 77.5ms, 124.3ms, and 225.3ms. The variance of all models reaches up to 19.4$\times$. Such variance also arises from the large device capacity gap. We observe that the score of the highest-end GPU in our dataset, Nvidia GTX 2060, is 207.4$\times$ better than the lowest-end Intel HD 530 according to the GPU benchmark~\cite{gpu-benchmark}.
Since WebGL has limited capacity to harness discrete GPUs, the variance is less pronounced compared to the differences in GPU capacity.

\revise{
    Regarding the performance differences between different models on integrated GPUs, we compared the Apple silicon GPUs with GPUs integrated with Intel chips. We observed that Apple silicon GPUs generally outperform those in Intel chips. Specifically, for TF.js, prediction latency on Apple silicon GPUs is on average 7.1$\times$ lower than that on Intel chips; for ORT.js, the latency on Apple silicon GPUs is on average 6.2$\times$ lower than that on Intel chips. This enhanced performance is primarily due to the optimizations in the Apple GPU for parallel computing, which makes it more adept at handling model inference tasks.
    In summary, for in-browser inference, the standards of Wasm and WebGL are consistent across all devices, indicating that any device capable of running a Web browser can support in-browser inference. The differences in performance are indeed related to the capabilities of the different devices or hardware types, such as the capacity gap between integrated and discrete GPUs.

    Regarding the performance differences in in-browser inference across different CPU architectures, we compared the inference capabilities between ARM CPUs and x86 CPUs. For ARM architecture, we selected Apple Silicon, with our dataset including both the M1 and M2 chips, while for x86 architecture, we used Intel CPUs. We found that the latency on ARM CPUs is on average lower than x86 CPUs. Specifically, for TF.js, the latency on ARM CPUs was 2.2$\times$ slower than x86 CPUs, and for ORT.js, the gap is 2.9$\times$. This disparity is significantly smaller than the previously analyzed factor of 28.6$\times$.
    The primary reason for this smaller gap is the considerable variation in the performance of the Intel chips used in our dataset, which ranged from Core-i5 Skylake models to Core-i7 Rocket Lake models. Lower-end Intel chips performed worse than Apple chips, whereas higher-end Intel models surpassed the performance of Apple chips.
    In summary, there is a greater variance in inference performance across different Intel CPUs, and the average performance gap between Intel and Apple CPUs is relatively small. The highest-end Intel CPUs demonstrated the best inference performance.
}

As for \textit{warmup latency}, the variance of all models reaches up to 25.3$\times$ in the Wasm backend and 14.4$\times$ in the WebGL backend. The variance is not as significant as the prediction latency. 
\revise{
    \sout{
        The reason is that the warmup latency consists of memory allocation overhead, which is not affected by the hardware capacity. Despite this, the variance is still large, and such a large variance is also attributed to the large hardware capacity gap.
    }
    The reason is that the warmup latency consists of memory allocation overhead. Compared to computation, the impact of hardware capacity on memory allocation is negligible. Consequently, when memory allocation is involved, the variance of warmup latency decreases compared to the variance in prediction latency. However, warmup latency remains predominantly driven by computation, i.e., initializing the first model inference. Therefore, due to significant performance differences between hardware capacities, the variance in warmup latency can still reach up to 25.3$\times$.
}

\revise{
    In summary, after analyzing the results of Figure~\ref{fig:inter-latency-distribution} and Figure~\ref{fig:warmup-latency-distribution} the variance in prediction latency and warmup latency across devices is quite significant, primarily attributed to variations in the hardware capacity of user devices, as both types of latency are predominated by computation. Compared to prediction latency, the variance in warmup latency across devices is relatively smaller. This is mainly because warmup latency also encompasses memory allocation, where the performance variance across devices is minimal, resulting in this observation.
}

\revise{
    \textbf{Model-level latency difference across different models and tasks.}
    We explore the latency difference across different models within the same task. Taking prediction latency as an example, we found that MobileNetV2 exhibits the lowest latency in image classification tasks in all settings, while Bert-Small shows the lowest latency in grammar checking tasks in all settings. This is primarily because these two models are the lightest within their respective task model sets. Specifically, MobileNetV2 has a model file size of only 14MB, and Bert-Small is only 17MB. However, for object detection tasks, SSD-MobileNetV2 performs best in the Wasm backend of ORT.js, while Yolo5-Middle shows the best performance in the WebGL backend of TF.js. This is mainly due to the differences in the implementations of the two backends across the two frameworks.
    We also observed significant differences in prediction latency among different models. Taking image classification as an example, in our experiments, the file sizes for the MobileNetV2 and ResNet50 models were 14MB and 98MB, respectively. Additionally, the primary kernels included in these models differ: MobileNetV2 mainly incorporates depthwise separable convolutions and pointwise convolutions, whereas ResNet50 predominantly uses 3x3 convolutions. These structural differences between the two models lead to variations in setup latency, warmup latency, and prediction latency.
    As for models used in different tasks, take for example the image classification model MobileNetV2 and the object detection model SSD-MobileNetV2. Although SSD-MobileNetV2 uses MobileNetV2 as its feature extractor, it omits the last four bottleneck layers of MobileNetV2 and incorporates additional detection layers. Such structural differences between the two models affect their inference latency.
    In summary, the differences in inference latency between models are primarily due to variations in their model structure.
}

\revise{
    \textbf{Results on Firefox browser.}    
    We further investigated the impact of different browsers. For each application category, we selected a model and presented the results in Table~\ref{tab:e2e-tfjs-wasm-firefox}, Table~\ref{tab:e2e-ort-wasm-firefox}, and Table~\ref{tab:e2e-webgl-firefox}.

    \begin{table*}[h]
    \centering
    \caption{\revise{Latency (ms) of supported models in TF.js in Wasm backend in Firefox browser. ``-SIMD/thread" indicates enabled technique.
    The results show that the Wasm backend of TF.js can support only four models, indicating limited compatibility. Enabling SIMD and multithreading technologies resulted in decreases in in-browser inference prediction latency by 49.2\% and 9.9\% on average, respectively. When both techniques were applied, they achieved a further reduction in prediction latency by 65.2\%, confirming their effectiveness in optimizing in-browser inference performance.
    }}
    \label{tab:e2e-tfjs-wasm-firefox}
    \vspace{-4mm}
    \resizebox{0.95\linewidth}{!}{%
    \begin{tabular}{@{}l|rrrrrrrrrrrr@{}}
    \toprule[1.5pt]
                              & \multicolumn{3}{c|}{Wasm}                                                                   & \multicolumn{3}{c|}{Wasm-thread}                                                          & \multicolumn{3}{c|}{Wasm-SIMD}                                                            & \multicolumn{3}{c}{Wasm-thread-SIMD}                                                      \\ \cmidrule(l){2-13} 
                              & \multicolumn{1}{c}{setup} & \multicolumn{1}{c}{warmup} & \multicolumn{1}{c|}{prediction}    & \multicolumn{1}{c}{setup} & \multicolumn{1}{c}{warmup} & \multicolumn{1}{c|}{prediction}  & \multicolumn{1}{c}{setup} & \multicolumn{1}{c}{warmup} & \multicolumn{1}{c|}{prediction}  & \multicolumn{1}{c}{setup} & \multicolumn{1}{c}{warmup} & \multicolumn{1}{c}{prediction}   \\ \midrule
    (M1) MobilenetV2          & 408.6                     & 345.9                      & 250.8                              & 351.4                     & 313.1                      & 239.5                            & 402.1                     & 205.1                      & 123.5                            & 260.8                     & 172.1                      & 93.9                             \\
    (M2) Resnet50             & 2529.2                    & 2102.9                     & 1905.0                             & 2261.8                    & 1936.4                     & 1574.4                           & 2546.9                    & 993.0                      & 592.4                            & 1766.0                    & 743.0                      & 341.0                            \\
    (M3) VGG16                & 10110.7                   & 8803.6                     & 6701.2                             & 10417.8                   & 8606.1                     & 5803.8                           & 10856.9                   & 6090.2                     & 3294.7                           & 9182.0                    & 2641.1                     & 1132.5                           \\
    (M4) SSD-MobilenetV2      & 612.1                     & 1532.9                     & 1249.2                             & 571.3                     & 1252.0                     & 1195.4                           & 644.5                     & 1148.6                     & 922.7                            & 498.3                     & 1073.5                     & 835.6                            \\ \hline
    Average                   & 3415.2                    & 3196.3                     & 2526.6                             & 3400.6                    & 3026.9                     & 2203.3                           & 3612.6                    & 2109.2                     & 1233.3                           & 2926.8                    & 1157.4                     & 600.8                            \\
    \bottomrule [1.5pt]
    \end{tabular}%
    }
\end{table*}
    \begin{table*}[h]
    \centering
    \caption{\revise{Latency (ms) of models in ORT.js Wasm backend in Firefox browser. ``-SIMD/thread" indicates enabled technique. 
    We found that ORT.js's Wasm backend supports all the nine models tested. Upon enabling SIMD and multithreading, we recorded reductions in prediction latency of 59.2\% and 49.2\% on average, respectively. A combination of both acceleration techniques resulted in an additional latency decrease of 73.8\% compared to vanilla Wasm.}}
    \label{tab:e2e-ort-wasm-firefox}
    \vspace{-4mm}
    \resizebox{0.95\linewidth}{!}{%
    \begin{tabular}{@{}l|rrrrrrrrrrrr@{}}
    \toprule[1.5pt]
                                & \multicolumn{3}{c|}{Wasm}                                                                 & \multicolumn{3}{c|}{Wasm-thread}                                                          & \multicolumn{3}{c|}{Wasm-SIMD}                                                                & \multicolumn{3}{c}{Wasm-thread-SIMD}                                                      \\ \cmidrule(l){2-13} 
                                & \multicolumn{1}{c}{setup} & \multicolumn{1}{c}{warmup} & \multicolumn{1}{c|}{prediction}  & \multicolumn{1}{c}{setup} & \multicolumn{1}{c}{warmup} & \multicolumn{1}{c|}{prediction}  & \multicolumn{1}{c}{setup}  & \multicolumn{1}{c}{warmup} & \multicolumn{1}{c|}{prediction}     & \multicolumn{1}{c}{setup} & \multicolumn{1}{c}{warmup} & \multicolumn{1}{c}{prediction}   \\ \midrule
    (M1) MobilenetV2            & 589.6                     & 159.7                      & 132.8                            & 1389.0                    & 233.6                      & 134.9                            & 1276.4                     & 175.2                      & 115.5                               & 355.1                     & 151.8                      & 112.6                            \\
    (M2) Resnet50               & 558.6                     & 1509.5                     & 1363.6                           & 711.2                     & 434.8                      & 455.7                            & 1423.6                     & 458.4                      & 398.5                               & 1271.0                    & 360.0                      & 191.1                            \\
    (M3) VGG16                  & 7612.3                    & 10248.4                    & 9627.2                           & 4790.3                    & 8853.1                     & 8932.5                           & 2797.2                     & 3331.8                     & 3236.4                              & 2444.9                    & 1429.4                     & 1444.8                           \\
    (M4) SSD-MobilenetV2        & 1165.4                    & 304.9                      & 295.5                            & 1425.3                    & 229.8                      & 136.4                            & 1173.5                     & 186.8                      & 140.9                               & 1533.6                    & 93.0                       & 95.3                             \\
    (M5) EfficientDet           & 1640.8                    & 2893.8                     & 2821.9                           & 1898.6                    & 1384.2                     & 1434.3                           & 1710.9                     & 1427.4                     & 1330.1                              & 1961.7                    & 1029.6                     & 920.6                            \\
    (M6) Yolo5-Middle           & 428.9                     & 1186.3                     & 1280.8                           & 477.4                     & 406.5                      & 412.5                            & 421.4                      & 460.1                      & 450.0                               & 467.9                     & 160.9                      & 167.1                            \\
    (M7) Bert-base              & 804.1                     & 1017.8                     & 1020.3                           & 744.2                     & 310.5                      & 296.3                            & 716.4                      & 287.1                      & 290.3                               & 576.7                     & 100.6                      & 90.1                             \\
    (M8) MobileBert             & 875.0                     & 928.8                      & 930.0                            & 946.0                     & 351.2                      & 334.2                            & 868.7                      & 259.6                      & 257.4                               & 959.0                     & 135.6                      & 124.0                            \\
    (M9) Bert-Small             & 155.9                     & 22.6                       & 24.8                             & 172.1                     & 10.2                       & 8.7                              & 167.5                      & 7.7                        & 7.8                                 & 171.0                     & 4.9                        & 5.3                              \\ \hline
    Average                     & 1536.7                    & 2030.2                     & 1944.1                           & 1394.9                    & 1357.1                     & 1349.5                           & 1172.8                     & 732.7                      & 691.9                               & 1082.3                    & 385.1                      & 350.1                            \\
    \bottomrule[1.5pt]
    \end{tabular}%
    }
\end{table*}
    \begin{table*}[h]
    \caption{\revise{Latency (ms) of models in WebGL backend in Firefox browsers. ``/'' indicates the same as Table~\ref{tab:e2e-webgl}.
    The results demonstrate that the WebGL backend of TF.js can accommodate all nine models. Furthermore, the warmup latency for performing inference on GPU surpasses the prediction latency, with average increases of 45.8$\times$ for TF.js and 38.9$\times$ for ORT.js.}}
    \label{tab:e2e-webgl-firefox}
    \vspace{-4mm}
    \resizebox{0.95\linewidth}{!}{%
    \begin{tabular}{@{}l|rrrrrr|rrrrrr@{}}
\toprule[1.5pt]
                        & \multicolumn{3}{c|}{ORT.js-WebGL-I}                                                           & \multicolumn{3}{c|}{ORT.js-WebGL-D}                                                                   & \multicolumn{3}{c|}{TF.js-WebGL-I}                                                                    & \multicolumn{3}{c}{TF.js-WebGL-D}                                                             \\ \cmidrule(l){2-13} 
                        & \multicolumn{1}{c}{setup} & \multicolumn{1}{c}{warmup}     & \multicolumn{1}{c|}{prediction}  & \multicolumn{1}{c}{setup}          & \multicolumn{1}{c}{warmup}     & \multicolumn{1}{c|}{prediction} & \multicolumn{1}{c}{setup}          & \multicolumn{1}{c}{warmup}     & \multicolumn{1}{c|}{prediction} & \multicolumn{1}{c}{setup} & \multicolumn{1}{c}{warmup}     & \multicolumn{1}{c}{prediction}   \\ \midrule
(M1) MobilenetV2        & 225.4                     & 2225.2                         & 34.1                             & 313.6                              & 3299.9                         & 32.9                            & 222.6                              & 3312.2                         & 30.9                            & 579.5                     & 6072.4                         & 21.3                             \\
(M2) Resnet50           & 769.9                     & 1825.3                         & 207.3                            & 2145.9                             & 3122.1                         & 80.2                            & 1668.6                             & 2927.9                         & 265.9                           & 2679.9                    & 5794.3                         & 85.9                             \\
(M3) VGG16              & 737.5                     & 1884.5                         & 617.4                            & 2257.1                             & 2825.5                         & 166.4                           & 9340.9                             & 3834.8                         & 1999.3                          & 10618.5                   & 5431.4                         & 1028.8                           \\
(M4) SSD-MobilenetV2    & /                         & /                              & /                                & /                                  & /                              & /                               & 511.1                              & 7510.0                         & 1183.9                          & 1196.2                    & 10075.0                        & 664.2                            \\
(M5) EfficientDet-D1    & /                         & /                              & /                                & /                                  & /                              & /                               & 772.1                              & 9914.6                         & 2845.2                          & 2654.2                    & 12557.6                        & 1521.1                           \\
(M6) Yolo5-Middle       & /                         & /                              & /                                & /                                  & /                              & /                               & 1895.3                             & 6487.3                         & 101.8                           & 2173.0                    & 11244.7                        & 94.5                             \\
(M7) Bert-base          & /                         & /                              & /                                & /                                  & /                              & /                               & 2014.7                             & 3251.5                         & 697.4                           & 3574.9                    & 3598.9                         & 141.2                            \\
(M8) MobileBert         & /                         & /                              & /                                & /                                  & /                              & /                               & 3100.0                             & 3397.4                         & 1079.5                          & 3748.9                    & 2504.1                         & 426.2                            \\
(M9) Bert-Small         & /                         & /                              & /                                & /                                  & /                              & /                               & 709.7                              & 2250.9                         & 45.8                            & 848.4                     & 2578.4                         & 61.1                             \\ \hline
Average                 & 577.6                     & 1978.3                         & 286.3                            & 1572.2                             & 3082.5                         & 93.2                            & 2248.3                             & 4765.2                         & 916.6                           & 3119.3                    & 6650.8                         & 449.4                            \\
\bottomrule[1.5pt]  
\end{tabular}%
}
\end{table*}
    
    Starting with the CPU results. The first is prediction latency. By comparing Table~\ref{tab:e2e-tfjs-wasm-firefox} and Table~\ref{tab:e2e-ort-wasm-firefox} with Table~\ref{tab:e2e-native-all}, we still observed that a gap still exists between in-browser inference and native inference, averaging 8.8$\times$. This is primarily due to the differences between the browser and native environments, including the absence of advanced instructions and inevitable resource competition. Additionally, we compared in-browser inference between Firefox and Chrome (Table~\ref{tab:e2e-tfjs-wasm-firefox} and Table~\ref{tab:e2e-ort-wasm-firefox} vs Tables~\ref{tab:e2e-tfjs-wasm} and Table~\ref{tab:e2e-ort-wasm}). We found that for the same framework, the Chrome browser outperforms the Firefox browser consistently. Specifically, for TF.js, the average prediction latency on Firefox is 1.08$\times$ that of Chrome; for ORT.js, it is 1.11$\times$.
    The second is warmup latency. Compared with native inference latencies, the results are similar to those on Chrome, i.e., in-browser inference warmup latency is on average 0.37-1.14$\times$ of native inference for TF.js; the gap for ORT.js and ORT is 1.07-22.32$\times$. The difference between browsers remains low, with Firefox's results varying from Chrome's by no more than 10.3\% for both frameworks. The main difference between the two browsers with respect to warmup and prediction latency arises from the implementation of the browsers because the standard of Wasm remains the same between the two browsers.
    The third is setup latency. Compared to native inference, the setup latency for TF.js is 0.05-10.5$\times$ that of TF, and for ORT.js it is 0.84-2.6$\times$ that of ORT. Since in-browser inference setup latency is mainly influenced by library size and network conditions, there are significant differences between models as well as between native and browser environments. Among different browsers, the setup latency on Firefox is 1.03$\times$ that of Chrome, which is slightly lower than prediction and warmup latencies. This is because the setup stage is predominantly by network conditions as previously analyzed. In summary, the difference between the two browsers is minimal and can be considered negligible compared to the native vs. browser differences.

    Furthermore, we analyzed the impact of different acceleration technologies, specifically multithreading and SIMD. As with previous experiments, we set up 4 threads. We found that SIMD can reduce the prediction latency for TF.js and ORT.js by 49.2\% and 59.2\%, respectively, while multithreading can reduce it by 9.9\% and 49.2\% for TF.js and ORT.js, respectively. When both technologies were enabled, the prediction latency was reduced by 65.2\% and 73.8\% for TF.js and ORT.js, respectively. Similar to Chrome, the reason multithreading is less effective than SIMD is due to unavoidable resource competition and the overhead of thread synchronization, which is determined by the Wasm itself and is independent of the browser architecture.

    Next, we analyzed the average latency on the GPU. Firstly, the setup latency for in-browser inference is 0.1-12.2$\times$ that of native inference for TF.js, and the gap is 0.2-16.4$\times$ for ORT.js and ORT. Compared to Chrome, Firefox's setup latency on a discrete GPU is 1.06$\times$ on average higher, and on an integrated GPU, the average gap is 1.05$\times$, which is negligible compared with the gap between native inference. Secondly, as for warmup latency, compared to native inference, the average in-browser inference warmup latency is 6.1$\times$ that of native inference on average for TF.js and the gap is 13.8$\times$ for ORT.js, primarily due to time-consuming in-browser WebGL shader compilation; compared to Chrome, Firefox's data averages 1.11$\times$ on discrete GPU and 1.08$\times$ on integrated GPU. Finally, compared to native inference, the prediction latency for in-browser inference averages 40.2$\times$ that of native inference for TF.js, and the average gap is 15.5$\times$ for ORT.js; compared to Chrome, Firefox's data averages 1.09$\times$ higher for both frameworks. We found that similar to the results on the CPU, Firefox's performance is somewhat inferior to Chrome's. This is primarily due to the differences in framework implementation, as the two browsers use different engines, which accounts for the variations in performance.

    In summary, the performance differences in performing in-browser inference between browsers are minimal, compared with native inference. Although Chrome's in-browser inference performance is slightly better than Firefox's, the gap is still much smaller than the difference between browser-based and native inference. This discrepancy is mainly due to variations in the internal implementations of the browsers. The slight differences between the two browsers are also because both are seasoned engineering experts who optimize both to support general JavaScript execution and browsing tasks.
}

\revise{
    \textbf{Results on mobile devices.} 
    In addition, we explored the inference latency performance in-browser on mobile devices. We selected a specific model for each type of application and presented the results in Table~\ref{tab:e2e-native-all-mobile}, Table~\ref{tab:e2e-tfjs-wasm-mobile}, Table~\ref{tab:e2e-ort-wasm-mobile}, and Table~\ref{tab:e2e-webgl-mobile}. The experiments were conducted using TensorFlow Lite (TFLite) and ONNX Mobile Runtime (mORT).


\begin{table*}[h]
    \caption{\revise{Latency (ms) of different models running in native on mobile devices. ``/'' indicates that the model is not supported. 
    We observed that TFLite offers less comprehensive model support compared to mORT on both CPU and GPU. Both frameworks facilitate GPU acceleration, resulting in GPU inference achieving lower average prediction latencies compared to CPU by 2.5$\times$ for TFLite and 1.7$\times$ for mORT \minor{for the models that are supported by both frameworks in each backend. For TFLite, the prediction latency averages 319.5ms on the CPU and 62.8ms on the GPU; for mORT, the prediction latency averages 141.6ms on the CPU and 101.4ms on the GPU.}}
    }
    \label{tab:e2e-native-all-mobile}
    \vspace{-4mm}
    \resizebox{0.95\linewidth}{!}{%
        \begin{tabular}{@{}l|rrrrrr|rrrrrr@{}}
        \toprule[1.5pt]
                                  & \multicolumn{6}{c|}{TFLite}                                                                                                                                                                   & \multicolumn{6}{c}{mORT}                                                                                                                                                            \\ \cmidrule(l){2-13} 
                                  & \multicolumn{3}{c|}{CPU}                                                                      & \multicolumn{3}{c|}{GPU}                                                                      & \multicolumn{3}{c|}{CPU}                                                                       & \multicolumn{3}{c}{GPU}                                                                    \\
                                  & \multicolumn{1}{l}{setup} & \multicolumn{1}{l}{warmup}     & \multicolumn{1}{l|}{prediction}  & \multicolumn{1}{l}{setup} & \multicolumn{1}{l}{warmup}     & \multicolumn{1}{l|}{prediction}  & \multicolumn{1}{l}{setup} & \multicolumn{1}{l}{warmup}     & \multicolumn{1}{l|}{prediction}   & \multicolumn{1}{l}{setup} & \multicolumn{1}{l}{warmup}   & \multicolumn{1}{l}{prediction}  \\ \midrule
        (M1) MobilenetV2          & 22.0                      & 21.3                           & 7.8                              & 1123.8                    & 17.6                           & 9.9                              & 186.7                     & 37.7                           & 24.8                              & 54.2                      & 29.1                         & 13.7                            \\
        (M2) Resnet50             & 134.1                     & 130.9                          & 92.9                             & 1478.4                    & 68.4                           & 54.7                             & 265.4                     & 131.0                          & 100.3                             & 133.0                     & 84.0                         & 66.7                            \\
        (M3) VGG16                & 668.6                     & 425.8                          & 403.6                            & 3484.5                    & 140.9                          & 121.7                            & 1386.4                    & 366.1                          & 357.4                             & 935.4                     & 228.9                        & 216.0                           \\
        (M4) SSD-MobilenetV2      & 658.1                     & 138.0                          & 130.4                            & /                         & /                              & /                                & 1420.7                    & 125.2                          & 95.4                              & 691.1                     & 65.2                         & 47.7                            \\
        (M5) EfficientDet         & 1103.5                    & 9983.9                         & 1339.1                           & /                         & /                              & /                                & 1570.2                    & 446.1                          & 407.2                             & 1570.2                    & 446.1                        & 407.2                           \\
        (M6) Yolo5-Middle         & 571.5                     & 295.8                          & 259.4                            & 2424.0                    & 74.2                           & 64.7                             & 765.1                     & 153.6                          & 125.3                             & 371.8                     & 98.7                         & 72.8                            \\
        (M7) Bert-base            & /                         & /                              & /                                & /                         & /                              & /                                & 1167.2                    & 127.1                          & 72.1                              & 572.3                     & 57.1                         & 34.9                            \\
        (M8) MobileBert           & /                         & /                              & /                                & /                         & /                              & /                                & 1629.2                    & 104.0                          & 82.7                              & 976.0                     & 69.4                         & 47.5                            \\
        (M9) Bert-Small           & 17.8                      & 8.7                            & 3.6                              & /                         & /                              & /                                & 160.0                     & 15.1                           & 9.2                               & 160.0                     & 9.3                          & 5.9                             \\ \hline
        \minor{Average}           & \minor{453.7}             & \minor{1572.1}                 & \minor{319.5}                    & \minor{2127.7}            & \minor{75.3}                   & \minor{62.8}                     & \minor{950.1}             & \minor{167.3}                  & \minor{141.6}                     & \minor{607.1}             & \minor{120.9}                & \minor{101.4}                   \\
        \bottomrule[1.5pt]
        \end{tabular}%
    }
\end{table*}

\begin{table*}[h]
    \centering
    \caption{\revise{Latency (ms) of supported models in TF.js in Wasm backend on mobile devices. ``-SIMD/thread" indicates enabled technique.
    The results revealed that TF.js's Wasm backend offers limited model support, accommodating only four models. \minor{In the vanilla Wasm backend, the prediction latency of the four supported models averages 5105.2ms.} By activating SIMD and multithreading, \minor{\sout{we observed reductions in in-browser inference prediction latency by 36.2\% and 13.3\%, respectively} the average prediction latency is 4260.2ms and 2696.8ms, respectively. The average prediction latency reduction of each model averages 36.2\% and 13.3\%, respectively}. Furthermore, enabling both technologies concurrently led to \minor{an average} decrease in prediction latency by 54.1\%, \minor{where the latency of all models averages 1649.7ms}, highlighting their combined effectiveness in accelerating in-browser inference.}}
    \label{tab:e2e-tfjs-wasm-mobile}
    \vspace{-4mm}
    \resizebox{0.95\linewidth}{!}{%
    \begin{tabular}{@{}l|rrrrrrrrrrrr@{}}
    \toprule[1.5pt]
                              & \multicolumn{3}{c|}{Wasm}                                                                   & \multicolumn{3}{c|}{Wasm-thread}                                                          & \multicolumn{3}{c|}{Wasm-SIMD}                                                            & \multicolumn{3}{c}{Wasm-thread-SIMD}                                                      \\ \cmidrule(l){2-13} 
                              & \multicolumn{1}{c}{setup} & \multicolumn{1}{c}{warmup} & \multicolumn{1}{c|}{prediction}    & \multicolumn{1}{c}{setup} & \multicolumn{1}{c}{warmup} & \multicolumn{1}{c|}{prediction}  & \multicolumn{1}{c}{setup} & \multicolumn{1}{c}{warmup} & \multicolumn{1}{c|}{prediction}  & \multicolumn{1}{c}{setup} & \multicolumn{1}{c}{warmup} & \multicolumn{1}{c}{prediction}   \\ \midrule
    (M1) MobilenetV2          & 4163.9                    & 618.7                      & 532.4                              & 4238.9                    & 652.7                      & 484.9                            & 4431.0                     & 667.4                      & 428.5                           & 3691.0                    & 463.1                     & 324.7                            \\
    (M2) Resnet50             & 23855.5                   & 3867.0                     & 3370.6                             & 26953.4                   & 3205.1                     & 2838.8                           & 30543.8                    & 1745.2                     & 1187.1                          & 18561.5                   & 926.2                     & 846.3                            \\
    (M3) VGG16                & 125750.1                  & 15957.2                    & 13540.6                            & 131675.1                  & 13852.1                    & 11047.3                          & 133282.5                   & 10018.9                    & 6435.4                          & 101524.8                  & 4681.6                    & 3236.3                           \\
    (M4) SSD-MobilenetV2      & 7749.1                    & 3025.4                     & 2977.1                             & 7174.2                    & 3161.8                     & 2669.8                           & 7557.0                     & 2910.0                     & 2736.0                          & 7025.8                    & 2843.5                    & 2191.4                           \\ \hline
    \minor{Average}           & \minor{40379.7}           & \minor{5867.1}             & \minor{5105.2}                     & \minor{42510.4}           & \minor{5217.9}             & \minor{4260.2}                   & \minor{43953.6}            & \minor{3835.4}             & \minor{2696.8}                  & \minor{32700.8}           & \minor{2228.6}            & \minor{1649.7}                   \\
    \bottomrule [1.5pt]
    \end{tabular}%
    }
\end{table*}
    \begin{table*}[h]
    \centering
    \caption{\revise{Latency (ms) of models in ORT.js Wasm backend on mobile devices. ``-SIMD/thread" indicates enabled technique.
    The results revealed that ORT.js's Wasm backend is fully compatible, supporting all nine models. \minor{In the vanilla Wasm backend, the average prediction latency of the models is 2007.7ms.} Significant reductions in prediction latency were observed after enabling SIMD and multithreading, \minor{where the average prediction latency is 1239.9ms and 1415.5ms, respectively. For each model, the prediction latency decrease of each model averages 37.8\% and 17.0\%, respectively. \sout{with decreases of 37.8\% and 17.0\%, respectively.}} Enabling both technologies further reduced the prediction latency \minor{to 1165.4ms, and the prediction latency decrease of each model averages 42.4\%.}}}
    \label{tab:e2e-ort-wasm-mobile}
    \vspace{-4mm}
    \resizebox{0.95\linewidth}{!}{%
    \begin{tabular}{@{}l|rrrrrrrrrrrr@{}}
    \toprule[1.5pt]
                                & \multicolumn{3}{c|}{Wasm}                                                                 & \multicolumn{3}{c|}{Wasm-thread}                                                          & \multicolumn{3}{c|}{Wasm-SIMD}                                                                & \multicolumn{3}{c}{Wasm-thread-SIMD}                                                      \\ \cmidrule(l){2-13} 
                                & \multicolumn{1}{c}{setup} & \multicolumn{1}{c}{warmup} & \multicolumn{1}{c|}{prediction}  & \multicolumn{1}{c}{setup} & \multicolumn{1}{c}{warmup} & \multicolumn{1}{c|}{prediction}  & \multicolumn{1}{c}{setup}  & \multicolumn{1}{c}{warmup} & \multicolumn{1}{c|}{prediction}     & \multicolumn{1}{c}{setup} & \multicolumn{1}{c}{warmup} & \multicolumn{1}{c}{prediction}   \\ \midrule
    (M1) MobilenetV2            & 4622.5                    & 442.5                      & 408.1                            & 6605.8                    & 384.8                      & 276.6                            & 6373.1                     & 350.1                      & 254.9                               & 5992.8                    & 324.1                      & 226.2                            \\
    (M2) Resnet50               & 15954.1                   & 1571.0                     & 1545.3                           & 18448.7                   & 1248.7                     & 1186.3                           & 18609.3                    & 1213.1                     & 1015.9                              & 17918.3                   & 1103.1                     & 901.8                            \\
    (M3) VGG16                  & 195834.1                  & 4085.7                     & 4229.6                           & 204477.7                  & 3546.4                     & 2847.4                           & 183901.6                   & 3298.8                     & 2490.5                              & 191393.7                  & 3013.0                     & 2398.2                           \\
    (M4) SSD-MobilenetV2        & 23861.4                   & 1172.4                     & 1317.5                           & 23468.6                   & 956.9                      & 927.7                            & 23181.6                    & 913.2                      & 875.4                               & 21861.1                   & 854.7                      & 779.1                            \\
    (M5) EfficientDet           & 20333.2                   & 6328.3                     & 5689.6                           & 18789.1                   & 4361.7                     & 4080.2                           & 18469.7                    & 4073.2                     & 3589.3                              & 18689.6                   & 4926.0                     & 3430.4                           \\
    (M6) Yolo5-Middle           & 4928.9                    & 2018.6                     & 1669.9                           & 4700.4                    & 1577.6                     & 1170.3                           & 4568.2                     & 1354.1                     & 1026.5                              & 4547.6                    & 1145.8                     & 994.8                            \\
    (M7) Bert-base              & 24412.7                   & 1515.9                     & 1365.7                           & 22893.9                   & 920.9                      & 961.8                            & 22875.6                    & 960.8                      & 754.6                               & 21673.4                   & 825.4                      & 732.4                            \\
    (M8) MobileBert             & 19406.6                   & 1586.2                     & 1746.0                           & 20109.7                   & 1226.8                     & 1215.1                           & 19340.8                    & 1140.8                     & 1090.0                              & 18144.7                   & 1070.8                     & 966.6                            \\
    (M9) Bert-Small             & 6550.4                    & 89.9                       & 97.4                             & 5820.4                    & 78.6                       & 73.7                             & 5667.4                     & 70.5                       & 61.9                                & 5802.1                    & 63.4                       & 58.7                             \\ \hline
    \minor{Average}             & \minor{35100.4}           & \minor{2090.1}             & \minor{2007.7}                   & \minor{36146.0}           & \minor{1589.2}             & \minor{1415.5}                   & \minor{33665.3}            & \minor{1486.1}             & \minor{1239.9}                      & \minor{34002.6}           & \minor{1480.7}             & \minor{1165.4}                   \\
    \bottomrule[1.5pt]
    \end{tabular}%
    }
\end{table*}
    \begin{table*}[h]
    \caption{\revise{Latency (ms) of models in WebGL backend on mobile devices. ``/'' indicates the same as Table~\ref{tab:e2e-webgl}.
    \minor{
        The average prediction latency of TF.js in the WebGL backend is 541.8ms; the average prediction latency of ORT.js is 1132.2ms.
    }
    The results indicate that TF.js's WebGL backend supports more models, supporting six more models than ORT.js, the same as the results on PC devices. Additionally, we noted that the warmup latency for GPU inference exceeds the prediction latency, with an average increase of 34.6$\times$ and 15.0$\times$ on average, for TF.js and ORT.js respectively.}}
    \label{tab:e2e-webgl-mobile}
    \vspace{-4mm}
    \resizebox{0.65\linewidth}{!}{%
    \begin{tabular}{@{}l|rrr|rrr@{}}
\toprule[1.5pt]
                        & \multicolumn{3}{c|}{ORT.js-WebGL}                                                              & \multicolumn{3}{c}{TF.js-WebGL}                                                              \\ \cmidrule(l){2-7} 
                        & \multicolumn{1}{c}{setup} & \multicolumn{1}{c}{warmup}     & \multicolumn{1}{c|}{prediction}   & \multicolumn{1}{c}{setup} & \multicolumn{1}{c}{warmup}    & \multicolumn{1}{c}{prediction}   \\ \midrule
(M1) MobilenetV2        & 6934.2                    & 1424.6                         & 66.8                              & 7962.4                    & 8986.7                        & 47.8                             \\
(M2) Resnet50           & 80720.6                   & 3559.6                         & 456.3                             & 33607.8                   & 8284.6                        & 626.3                            \\
(M3) VGG16              & 198767.6                  & 17329.5                        & 1102.2                            & 154695.5                  & 7281.8                        & 1892.0                           \\
(M4) SSD-MobilenetV2    & /                         & /                              & /                                 & 17363.4                   & 17247.1                       & 1223.2                           \\
(M5) EfficientDet-D1    & /                         & /                              & /                                 & 34967.2                   & 18185.6                       & 3712.5                           \\
(M6) Yolo5-Middle       & /                         & /                              & /                                 & 30862.0                   & 21188.9                       & 374.2                            \\
(M7) Bert-base          & /                         & /                              & /                                 & 47464.7                   & 5313.2                        & 984.5                            \\
(M8) MobileBert         & /                         & /                              & /                                 & 47378.7                   & 3528.3                        & 1136.4                           \\
(M9) Bert-Small         & /                         & /                              & /                                 & 12449.0                   & 4233.5                        & 192.6                            \\ \hline
\minor{Average}         & \minor{95474.1}           & \minor{7437.9}                 & \minor{541.8}                     & \minor{42972.3}           & \minor{10472.2}               & \minor{1132.2}                   \\
\bottomrule[1.5pt]
\end{tabular}%
}
\end{table*}

    Starting with the analysis of the results of the CPU. 
    The first aspect is prediction latency, where we enabled SIMD and set up four threads. By comparing Table~\ref{tab:e2e-native-all-mobile} with Table~\ref{tab:e2e-tfjs-wasm-mobile} and Table~\ref{tab:e2e-ort-wasm-mobile}, we found that the difference between native and browser environments on mobile devices is still significant as on PCs. Specifically, the latency using TF.js for prediction is 18.9$\times$ that of using TFLite, and for ORT.js and mORT, this difference is 14.5$\times$. For both frameworks, the average gap reaches 15.8$\times$. The difference between native and browser mainly stems from resource contention. This is primarily because mobile devices inherently have limited resources, and in-browser inference also requires browser support. Running Chrome on mobile devices inevitably leads to resource competition. When performing native inference, we directly utilized the C++ implementation and executed the model inference via ``adb shell'', thus significantly avoiding resource contention. In contrast, in-browser inference relies on the Chrome browser. Additionally, the absence of FMA instructions in Wasm also impacts performance.
    The second aspect is warmup latency. In the mobile environment, the latency for TF.js is 25.7$\times$ of TFLite; \minor{
        \sout{for mORT, it is 8.0$\times$ of ORT.js}
        while the latency of ORT.js is 8.0$\times$ of mORT. For both frameworks, the average warmup latency gap between the native and browser environment is 10.2$\times$
    }.  Additionally, we observed that compared to TFLite, mORT also demonstrated lower warmup latency, a result that is similar to what we see on the PC side. Compared to prediction latency, the warmup latency for TF.js is 1.3$\times$ that of the prediction latency, while for ORT.js, this gap is 1.2$\times$. The difference in latency between the two stages is not as significant as that between the browser and the native environment. The variation between the two stages mainly arises from additional overheads such as memory allocation during the first model prediction.
    The final aspect is setup latency. We found that the latency for TF.js is 117.2$\times$ of TFLite; \minor{\sout{for mORT, it is 28.1$\times$ of ORT.js} for ORT.js, its latency is 28.1$\times$ of mORT. For both frameworks, the average setup latency gap is 62.0 $\times$.}
    We discovered that this gap is much larger than that on PC devices. Compared to warmup and prediction, setup latency is mainly affected by model size, library size, and network conditions. The setup latency of in-browser inference is high because mobile devices can only connect to wireless networks. According to our measurement, the download bandwidth ranges from 5.4MB/s to 7.8MB/s, which is much slower than that on PC devices using high-speed wired networks. Different from the PC side, on mobile devices both libraries are lightweight. Specifically, the TFLite library is only 5.3MB, and the mORT library is just 14MB. This indicates that the latency for loading libraries on mobile devices is low. However, as the model changes, there are significant differences in the latency for loading models. For example, the setup latency for MobileNetV2 is only 22.0ms, while for VGG16 it reaches 668.6ms. This difference is primarily because the setup stage also includes loading the model. During this process, the framework loads the model to memory, involving parsing the model structure, and thus, as the model structure becomes more complex, the setup latency increases accordingly.

    Furthermore, we analyzed the impact of different acceleration technologies, i.e., multithreading and SIMD. As with previous experiments, we set four threads when enabling multithreading. We found that SIMD could reduce the prediction latency for TF.js by 36.2\% and for ORT.js by 37.8\%, while multithreading could reduce the latency by 13.3\% and 17.0\% for TF.js and ORT.js, respectively. When both technologies were enabled, prediction latency decreased by 54.1\% and 42.4\% for TF.js and ORT.js, respectively. Similar to PC devices, SIMD also brings more acceleration than multithreading on mobile devices, and the effect of multithreading is not as pronounced as on PCs due to limited resources on mobile devices, leading to significant resource competition when four threads are enabled, thus lowering the multithreading performance.

    Next, we analyzed data on the GPU.
    The first aspect is prediction latency. In the mobile environment, the latency for TF.js is 9.4$\times$ of TFLite; \minor{\sout{for mORT, it is 5.6$\times$ of ORT.js.}for ORT.js, its latency is 5.6$\times$ of mORT. The average gap for both frameworks is 7.8$\times$.} This difference is much smaller than on PC devices, mainly because mobile GPUs do not possess the powerful parallel computing capabilities of Nvidia GPUs and do not have advanced libraries like CUDA for PCs.hence the smaller difference.
    In-browser inference is slower than native inference, primarily because kernel implementation of native inference is implemented directly via OpenCL kernels, which can run on mobile devices directly after compilation. In contrast, in-browser inference requires implementation through WebGL.  
    The second aspect is warmup latency. In the mobile environment, the latency for TF.js is 242.2$\times$ of TFLite on average; \minor{\sout{for mORT, it is 29.2$\times$ of ORT.js on average} the latency of ORT.js is 29.2$\times$ of mORT on average. For both frameworks, the average gap between the mobile native and browser environment is 162.3$\times$}. Such larger gaps mainly arise from the poorer performance of mobile GPUs, and in-browser WebGL shader code compilation depends on GPU performance, thus the significant difference. The gap on mobile devices compared to the gap on PC GPUs is quite different, primarily due to the use of completely different hardware. On PCs, due to limitations of the native inference framework, we use discrete GPUs and perform inference with CUDA, which involves considerable initialization latency for the discrete GPUs, whereas on mobile devices, the GPU is on the same chip as the CPU. Additionally, TFLite's lack of support for some models due to framework implementation also impacts the results. Nevertheless, there is still a large gap between the warmup latencies of native inference and in-browser inference on mobile devices.
    The last aspect is setup latency. In the mobile environment, the latency for TF.js is 21.7$\times$ of TFLite; \minor{\sout{for mORT, it is 315.8$\times$ of ORT.js} for ORT.js, it is 315.8$\times$ of mORT. For both frameworks, the average setup latency gap is 81.7$\times$}. This difference is also larger compared to PCs. The setup stage of in-browser inference is heavily influenced by network conditions, as analyzed in the CPU result part. Mobile devices suffer from higher setup stage latency due to lower network transmission bandwidth, a problem not present in native inference. Native inference merely requires loading from the local file system.

    In summary, on mobile devices, the performance differences between native and browser environments are still large, primarily due to the constraints of SIMD on CPUs, and inevitable resource competition. Compared to PC devices, in-browser inference latency on mobile devices is generally higher, largely due to limited resources. This scenario highlights both advantages and disadvantages. On the positive side, the uniformity between native and browser environments simplifies the development and deployment processes of in-browser inference. However, the lack of powerful computational libraries and resource constraints can lead to increased latency, presenting a significant challenge for real-time applications on mobile platforms.
}

\subsubsection{Kernel-level Latency}\label{subsubsec:kernel-level-latency-analysis}

We explore the kernel-level inference performance of in-browser inference. 
Due to the immature support for profiling in ORT.js, we use TF.js for analysis. We provide prediction results in Table~\ref{tab:tfjs-prediction-kernel-latency-proportion} and warmup results in Table~\ref{tab:tfjs-warmup-kernel-latency-proportion}. The setup stage is excluded because kernels are not executed in this stage.

First, we explore which kernels dominate the \textit{prediction latency}. Both multithreading and SIMD are enabled and we set four threads in the Wasm backend. 
\revise{
    \sout{
        Table~\ref{tab:tfjs-prediction-kernel-latency-proportion} shows the latency proportion of top-5 kernels.
    }
    Each percentage value in Table~\ref{tab:tfjs-prediction-kernel-latency-proportion} shows the latency proportion out of the total latency in a setting.
}
We find that computation-intensive kernels predominantly contribute to the overall latency in both backends, although the number of memory-intensive kernels makes up 37.7\% of the total \revise{kernels}. Specifically, Einsum and FusedConv2d contribute more than 70\% of the total latency.
We also find that the top-5 kernels are different in \revise{\sout{2} the two} backends due to different computation and memory-access patterns. In the Wasm backend, memory-intensive kernels introduce more latency. For example, GatherV2 accounts for the third most prediction latency in Wasm (12.5\%), but it is not in the top-5 kernels in WebGL. 
The reason is that inter-thread synchronization relies on message passing in Wasm~\cite{webworker}, which introduces additional overhead and becomes particularly pronounced for memory-intensive kernels.
\revise{
    These results indicate that computation-intensive kernels remain a focal point for optimization in both backend kernels.
}

We also explore the impact of SIMD and multithreading on the kernel during the prediction stage. 
\revise{
    \sout{
        When enabling multithreading, we find that the prediction latency of memory-intensive kernels increases by up to 3.7\%, and their latency proportion increases by up to 8.6\%.
    }
    We set four threads when enabling multithreading. We compared the latency of each kernel with various acceleration technologies enabled, compared to the latency in the vanilla Wasm setting, and represented these variations using different cell colors in Table~\ref{tab:tfjs-prediction-kernel-latency-proportion}. Firstly, we found that enabling multithreading led to an increase of up to 3.7\% in the prediction latency of memory-intensive kernels compared to the settings without multithreading enabled. Furthermore, the proportion of latency contributed by these kernels within the overall inference process increased by up to 8.6\% (as seen in the results for GatherV2 in the ``Wasm-S'' and ``Wasm-S-T'' settings).
}
This is because these memory-intensive kernels have limited support for multithreading. We dive into the kernel implementation in the source code and find that multithreading is only employed in some kernels, including Pad, among others.
When enabling SIMD, we find that the latency proportion of computation-intensive kernels decreases by up to 6.8\% and memory-intensive kernel latency changes negligibly 
\revise{
    (, as indicated by the cell color). Additionally, we observed that the proportion of computation-intensive kernel latency within all kernels decreased by up to 12.8\% (as seen in the results for Einsum in the ``Wasm-T'' and ``Wasm-S-T'' settings).
}
This is because Wasm has limited support for data-moving SIMD instruction, hindering the potential acceleration for memory-intensive kernels through SIMD.
\revise{
    When computation-intensive kernels' latency decreases and memory-intensive kernels' latency remains unchanged, the proportion of latency attributed to memory-intensive kernels also increases. Conversely, the reason for the decrease in the proportion of latency for computation-intensive kernels is the same.
}
Nevertheless, each technique effectively accelerates computation-intensive kernels. For instance, compared with vanilla Wasm, SIMD and multithreading reduce the average prediction latency of FusedConv2D by 86.9\% and 66.7\%, respectively. When applying both techniques, the prediction latency decreases by 89.9\%. 

During the warmup stage, our observations remain consistent with previous findings. The latency in both backends is predominantly influenced by computation-intensive kernels. However, the Wasm backend exhibits a higher latency contribution from memory-intensive kernels compared to the WebGL backend. Upon enabling SIMD, we note a reduction in the latency proportion attributable to computation-intensive kernels by up to 2.0\%. This reduction is smaller than that observed in the prediction stage, primarily because memory allocation incurs additional overhead that is unaffected by SIMD.
\revise{
    \sout{
        , leading to a less pronounced decrease in latency. 
    }
    Although the computation time for computation-intensive kernels decreased, extra overhead is added due to memory allocation, which remains constant. Moreover, the latency of memory-intensive kernels also does not significantly decrease, these result in a modest reduction in the proportion of computation-intensive kernels after SIMD is enabled.
}
Conversely, when multithreading is enabled, the latency proportion of memory-intensive kernels experiences an increase of up to 3.2\%. This trend mirrors the prediction stage, where multithreading offers negligible acceleration to memory-intensive kernels while introducing extra overhead.


\revise{
    \sout{
        These results suggest that memory-intensive kernels also require attention, as they can introduce high latency. The acceleration techniques have not yet been fully leveraged in memory-intensive kernels, emphasizing the need for framework vendors to focus on optimizing these kernels.
    }
    
    In summary, by analyzing the data from Tables~\ref{tab:tfjs-prediction-kernel-latency-proportion} and Table~\ref{tab:tfjs-warmup-kernel-latency-proportion}, we find that computationally intensive kernels, such as Einsum and FusedConv2D, account for at least 76.0\% of total latency. Within the Wasm backend, SIMD has almost no effect on memory-intensive kernels, while multithreading, in contrast, leads to an increase in memory-intensive kernel latency. In contrast, both technologies reduce the latency of computationally intensive kernels by more than 10\%. Upon diving into the kernel implementations, we find that this is primarily due to insufficient multithreading optimizations for memory-intensive kernels by the framework. In the WebGL backend, computation-intensive kernels similarly dominate the latency. Therefore, for both backends, improving the latency of computationally intensive kernels has a greater impact on overall latency, and framework vendors should also focus on optimizing multithreading for memory-intensive kernels.
}

\finding{
    \textbf{Finding 1}: A large inference performance disparity exists between browser and native environments. The prediction latency gap in CPU is 16.9$\times$ on average. The shorter SIMD instructions and Wasm VM-inherent inefficiency are key reasons for the low performance of the Wasm backend. 
    SIMD provides up to 50.7\% latency reduction, and multithreading provides up to 31.2\% latency reduction.
    WebGL significantly accelerates prediction by up to 4.9$\times$ compared to Wasm, but suffers from severe cold start issues, with a slowdown of up to 64.6$\times$, mainly due to in-browser shader compilation.
    The prediction latency spectrum across devices reaches up to 28.4$\times$ in Wasm and 19.4$\times$ in WebGL, which is mainly attributed to different hardware capacities.
    Memory-intensive kernels can contribute up to 12.5\% of total latency.
}

\begin{table*}[h]
    \caption{Kernel prediction latency proportion in TF.js. ``-S/T'' indicates SIMD/multithreading. ``Dw'' indicates ``Depthwise''. ``-I/D'' indicates integrated/discrete GPU. 
    \revise{
        \sout{
            Each entry is the kernel prediction latency proportion. Cell color gray means the latency is a little longer than that of Wasm (1\%< increment <5\%); yellow means the latency is close to that of Wasm (difference <1\%); blue means the latency is much less than that of Wasm (decrement >10\%). Each entry is the kernel prediction latency proportion out of the total warmup latency.
        }
        Each percentage value represents the proportion of total latency for the kernel under the setting, such as Wasm-S which indicates enabling SIMD in Wasm. For example, GatherV2 accounts for 3.9\% of the total latency in the Wasm-S setting. The cell color represents the change in latency for the same kernel when different technologies are enabled on top of the vanilla Wasm. Cell color gray means the latency is a little longer than that of Wasm (1\%< increment <5\%); cell yellow means the latency is close to that of Wasm (difference <1\%); cell blue means the latency is much less than that of Wasm (decrement >10\%). For example, the yellow cell color for GatherV2 indicates that enabling SIMD in Wasm leads to less than a 1\% difference in the prediction latency of GatherV2 compared to the vanilla Wasm backend. The percentages and cell colors provide independent information.
        The data indicates that across all settings in the Wasm backend and GPU, computation-intensive kernels consistently dominate the in-browser inference prediction latency. However, due to the synchronization overhead among threads, the Wasm backend exhibits a higher proportion of memory-intensive kernels compared to the WebGL backend.
    }}
    \label{tab:tfjs-prediction-kernel-latency-proportion}

    \resizebox{\linewidth}{!}{%
        \begin{tabular}{lrrrr|l|lrr}
            \toprule[1.5pt]
            Kernel                             & Wasm   & Wasm-S                         & Wasm-T                         & Wasm-S-T                         &  & Kernel                             & WebGL-I              & WebGL-D               \\ \cline{1-5} \cline{7-9} 
            \multicolumn{1}{l|}{Einsum}        & 48.2\% & \cellcolor[HTML]{34CDF9}46.5\% & \cellcolor[HTML]{34CDF9}48.0\% & \cellcolor[HTML]{34CDF9}41.4\%   &  & \multicolumn{1}{l|}{Einsum}        & 37.1\%                & 48.4\%               \\
            \multicolumn{1}{l|}{FusedConv2D}   & 47.6\% & \cellcolor[HTML]{34CDF9}44.3\% & \cellcolor[HTML]{34CDF9}47.4\% & \cellcolor[HTML]{34CDF9}34.6\%   &  & \multicolumn{1}{l|}{FusedConv2D}   & 43.6\%                & 23.7\%               \\
            \multicolumn{1}{l|}{GatherV2}      & 1.3\%  & \cellcolor[HTML]{F8FF00}3.9\%  & \cellcolor[HTML]{C0C0C0}1.5\%  & \cellcolor[HTML]{C0C0C0}12.5\%   &  & \multicolumn{1}{l|}{Add}           & 1.2\%                 & 1.6\%                \\
            \multicolumn{1}{l|}{Slice}         & 0.3\%  & \cellcolor[HTML]{F8FF00}1.1\%  & \cellcolor[HTML]{C0C0C0}0.3\%  & \cellcolor[HTML]{C0C0C0}2.8\%    &  & \multicolumn{1}{l|}{Multiply}      & 1.8\%                 & 1.4\%                \\
            \multicolumn{1}{l|}{FusedMatMul}   & 0.8\%  & \cellcolor[HTML]{34CDF9}1.8\%  & \cellcolor[HTML]{34CDF9}0.8\%  & \cellcolor[HTML]{34CDF9}1.6\%    &  & \multicolumn{1}{l|}{FusedMatMul}   & 2.3\%                 & 2.9\%                \\
            \multicolumn{1}{l|}{FusedDwConv2D} & 1.3\%  & \cellcolor[HTML]{34CDF9}1.2\%  & \cellcolor[HTML]{34CDF9}1.3\%  & \cellcolor[HTML]{34CDF9}1.9\%    &  & \multicolumn{1}{l|}{FusedDwConv2D} & 1.5\%                 & 2.1\%                \\ \bottomrule[1.5pt] 
        \end{tabular}%
    }
    
\end{table*}

\begin{table*}[h]
    \caption{Kernel warmup latency proportion in TF.js. The table format remains the same with Table~\ref{tab:tfjs-prediction-kernel-latency-proportion}.
    \revise{
        We found that similar to the prediction stage, computation-intensive kernels dominate the latency during the warmup stage. Due to the overhead of synchronization among threads in the Wasm backend, memory-intensive kernels introduce a higher latency proportion compared to those in the WebGL backend.
    }}
    \label{tab:tfjs-warmup-kernel-latency-proportion}

    \resizebox{\linewidth}{!}{%
        \begin{tabular}{lrrrr|l|lrr}
            \toprule[1.5pt]
            Kernel                             & Wasm               & Wasm-S                         & Wasm-T                         & Wasm-S-T                           &  & Kernel                             & WebGL-I              & WebGL-D              \\ \cline{1-5} \cline{7-9} 
            \multicolumn{1}{l|}{Einsum}        & 36.6\%             & \cellcolor[HTML]{34CDF9}33.5\% & \cellcolor[HTML]{34CDF9}35.0\% & \cellcolor[HTML]{34CDF9}33.4\%     &  & \multicolumn{1}{l|}{Einsum}        & 38.1\%               & 36.7\%               \\
            \multicolumn{1}{l|}{FusedConv2D}   & 54.1\%             & \cellcolor[HTML]{34CDF9}52.3\% & \cellcolor[HTML]{34CDF9}53.4\% & \cellcolor[HTML]{34CDF9}52.1\%     &  & \multicolumn{1}{l|}{FusedConv2D}   & 49.5\%               & 52.5\%               \\
            \multicolumn{1}{l|}{GatherV2}      & 2.6\%              & \cellcolor[HTML]{F8FF00}3.1\%  & \cellcolor[HTML]{C0C0C0}2.4\%  & \cellcolor[HTML]{C0C0C0}4.5\%      &  & \multicolumn{1}{l|}{Add}           & 1.8\%                & 2.6\%                \\
            \multicolumn{1}{l|}{Slice}         & 0.6\%              & \cellcolor[HTML]{F8FF00}1.2\%  & \cellcolor[HTML]{C0C0C0}2.9\%  & \cellcolor[HTML]{C0C0C0}3.8\%      &  & \multicolumn{1}{l|}{Multiply}      & 2.4\%                & 3.4\%                \\
            \multicolumn{1}{l|}{FusedMatMul}   & 2.4\%              & \cellcolor[HTML]{34CDF9}2.1\%  & \cellcolor[HTML]{34CDF9}1.4\%  & \cellcolor[HTML]{34CDF9}1.3\%      &  & \multicolumn{1}{l|}{FusedMatMul}   & 1.9\%                & 2.0\%                \\
            \multicolumn{1}{l|}{FusedDwConv2D} & 2.1\%              & \cellcolor[HTML]{34CDF9}1.3\%  & \cellcolor[HTML]{34CDF9}1.1\%  & \cellcolor[HTML]{34CDF9}1.1\%      &  & \multicolumn{1}{l|}{FusedDwConv2D} & 3.1\%                & 2.8\%                \\ \toprule[1.5pt] 
        \end{tabular}%
    }
    
\end{table*}

\revise{
    \textbf{Kernel-level latency difference across different models and tasks.}
    To explore the differences in latency across different kernels, we primarily examined the types of kernels included in various models and the relationship between the Floating Point Operations Per Second (FLOPs) of these kernels and their corresponding latencies. We focused on three specific operators: Einsum, FusedDwConv2D, and FusedConv2D. FusedDwConv2D is mainly used in the MobileNetV2 and SSD-MobileNetV2 models, while FusedConv2D is commonly found in other convolutional neural networks. Einsum, primarily used for matrix multiplication, is utilized across all models. 
    For instance, FusedConv2D accounts for 87.4\% prediction latency of ResNet50 in the Wasm backend with both multithreading and SIMD enabled, while Einsum accounts for less than 0\%. In Contrast, Einsum accounts for 92.1\% prediction latency of MobileBert in the same setting while FusedConv2d accounts for 0\%.
    We found that models utilizing FusedDwConv2D, typically used in classification and detection tasks, exhibit both lower FLOPs and latency. Specifically, FusedDwConv2D accounts for 0.6\% parameters and 1.9\% prediction latency. Although this may sacrifice some inference accuracy, it significantly reduces latency. Running more lightweight models in browsers holds potential, as these models are computationally efficient.
    In summary, all three kernels are computation-intensive kernels that dominate the total latency, as well as the number of parameters. Specifically, the three kernels account for 99\% of total parameters. These results indicate that optimization on these kernels can significantly improve the in-browser inference performance, i.e., reducing latency.
}

\revise{
    \textbf{Results on Firefox browser.}
    We also explored results on Firefox, presenting the results of prediction and warmup latencies in Tables~\ref{tab:tfjs-prediction-kernel-latency-proportion-firefox} and Table~\ref{tab:tfjs-warmup-kernel-latency-proportion-firefox}, respectively. Starting with the prediction stage, the results are similar to those in Chrome. Computationally intensive kernels, such as FusedConv2d and Einsum, still represent the largest latency proportion even after enabling multithreading and SIMD, accounting for 41.6\% and 34.1\% respectively. This is primarily because the prediction stage of model inference is still dominated by computation. Upon separately enabling multithreading and SIMD, we observed a significant reduction in latency for these computationally intensive kernels compared to vanilla Wasm. Specifically, SIMD reduced the latency of computation-intensive kernels by 84.5\%, and multithreading reduced it by 63.1\%. However, due to the presence of memory-intensive kernels and the framework's lack of multithreading support for these kernels, along with limited acceleration from SIMD, the absolute latency values of memory-intensive kernels remained essentially unchanged, while their latency proportion increased when both technologies were enabled. On the GPU side, we arrived at similar conclusions. Due to the absence of thread synchronization and message passing, the latency proportion of memory-intensive kernels is so as low that they do not appear in the table.

    \begin{table*}[h]
    \caption{\revise{Kernel prediction latency proportion in TF.js in Firefox browser. The table format remains the same with Table~\ref{tab:tfjs-prediction-kernel-latency-proportion}.
    The results show that in all settings and hardware environments, the prediction latency for in-browser inference is predominantly influenced by computation-intensive kernels, such as FusedConv2D. Additionally, the Wasm backend demonstrates a greater prevalence of memory-intensive kernels due to the overhead associated with thread synchronization, in contrast to the WebGL backend.
    }}
    \label{tab:tfjs-prediction-kernel-latency-proportion-firefox}

    \resizebox{\linewidth}{!}{%
        \begin{tabular}{lrrrr|l|lrr}
            \toprule[1.5pt]
            Kernel                             & Wasm   & Wasm-S                         & Wasm-T                         & Wasm-S-T                         &  & Kernel                             & WebGL-I              & WebGL-D               \\ \cline{1-5} \cline{7-9} 
            \multicolumn{1}{l|}{Einsum}        & 47.4\% & \cellcolor[HTML]{34CDF9}46.2\% & \cellcolor[HTML]{34CDF9}47.8\% & \cellcolor[HTML]{34CDF9}41.6\%   &  & \multicolumn{1}{l|}{Einsum}        & 38.2\%                & 48.1\%               \\
            \multicolumn{1}{l|}{FusedConv2D}   & 46.9\% & \cellcolor[HTML]{34CDF9}43.2\% & \cellcolor[HTML]{34CDF9}45.4\% & \cellcolor[HTML]{34CDF9}34.1\%   &  & \multicolumn{1}{l|}{FusedConv2D}   & 44.4\%                & 41.6\%               \\
            \multicolumn{1}{l|}{GatherV2}      & 1.4\%  & \cellcolor[HTML]{F8FF00}3.0\%  & \cellcolor[HTML]{C0C0C0}1.8\%  & \cellcolor[HTML]{C0C0C0}10.9\%   &  & \multicolumn{1}{l|}{Add}           & 1.4\%                 & 1.9\%                \\
            \multicolumn{1}{l|}{Slice}         & 0.4\%  & \cellcolor[HTML]{F8FF00}1.1\%  & \cellcolor[HTML]{C0C0C0}0.6\%  & \cellcolor[HTML]{C0C0C0}3.6\%    &  & \multicolumn{1}{l|}{Multiply}      & 1.7\%                 & 1.3\%                \\
            \multicolumn{1}{l|}{FusedMatMul}   & 0.9\%  & \cellcolor[HTML]{34CDF9}0.9\%  & \cellcolor[HTML]{34CDF9}0.8\%  & \cellcolor[HTML]{34CDF9}2.4\%    &  & \multicolumn{1}{l|}{FusedMatMul}   & 2.5\%                 & 2.6\%                \\
            \multicolumn{1}{l|}{FusedDwConv2D} & 1.2\%  & \cellcolor[HTML]{34CDF9}1.3\%  & \cellcolor[HTML]{34CDF9}1.4\%  & \cellcolor[HTML]{34CDF9}1.8\%    &  & \multicolumn{1}{l|}{FusedDwConv2D} & 3.1\%                 & 1.4\%                \\ \bottomrule[1.5pt] 
        \end{tabular}%
    }
    
\end{table*}
    
    \begin{table*}[h]
    \caption{\revise{Kernel warmup latency proportion in TF.js in Firefox browser. The table format remains the same with Table~\ref{tab:tfjs-prediction-kernel-latency-proportion}.
    We observed that during the warmup stage, the latency of computation-intensive kernels remains the predominant factor. Moreover, due to the additional overhead from thread synchronization in the Wasm backend, the latency proportion of memory-intensive kernels exceeds that of the WebGL backend.}}
    \label{tab:tfjs-warmup-kernel-latency-proportion-firefox}

    \resizebox{\linewidth}{!}{%
        \begin{tabular}{lrrrr|l|lrr}
            \toprule[1.5pt]
            Kernel                             & Wasm               & Wasm-S                         & Wasm-T                         & Wasm-S-T                           &  & Kernel                             & WebGL-I              & WebGL-D              \\ \cline{1-5} \cline{7-9} 
            \multicolumn{1}{l|}{Einsum}        & 35.9\%             & \cellcolor[HTML]{34CDF9}33.0\% & \cellcolor[HTML]{34CDF9}34.8\% & \cellcolor[HTML]{34CDF9}33.2\%     &  & \multicolumn{1}{l|}{Einsum}        & 39.2\%               & 36.9\%               \\
            \multicolumn{1}{l|}{FusedConv2D}   & 54.5\%             & \cellcolor[HTML]{34CDF9}51.6\% & \cellcolor[HTML]{34CDF9}52.8\% & \cellcolor[HTML]{34CDF9}52.3\%     &  & \multicolumn{1}{l|}{FusedConv2D}   & 49.1\%               & 51.4\%               \\
            \multicolumn{1}{l|}{GatherV2}      & 2.9\%              & \cellcolor[HTML]{F8FF00}3.3\%  & \cellcolor[HTML]{C0C0C0}2.6\%  & \cellcolor[HTML]{C0C0C0}4.1\%      &  & \multicolumn{1}{l|}{Add}           & 1.9\%                & 2.4\%                \\
            \multicolumn{1}{l|}{Slice}         & 0.8\%              & \cellcolor[HTML]{F8FF00}1.7\%  & \cellcolor[HTML]{C0C0C0}3.3\%  & \cellcolor[HTML]{C0C0C0}4.0\%      &  & \multicolumn{1}{l|}{Multiply}      & 2.1\%                & 3.6\%                \\
            \multicolumn{1}{l|}{FusedMatMul}   & 2.1\%              & \cellcolor[HTML]{34CDF9}2.0\%  & \cellcolor[HTML]{34CDF9}1.5\%  & \cellcolor[HTML]{34CDF9}1.5\%      &  & \multicolumn{1}{l|}{FusedMatMul}   & 1.8\%                & 2.1\%                \\
            \multicolumn{1}{l|}{FusedDwConv2D} & 2.0\%              & \cellcolor[HTML]{34CDF9}1.7\%  & \cellcolor[HTML]{34CDF9}1.1\%  & \cellcolor[HTML]{34CDF9}1.2\%      &  & \multicolumn{1}{l|}{FusedDwConv2D} & 2.9\%                & 2.4\%                \\ \toprule[1.5pt] 
        \end{tabular}%
    }
    
\end{table*}

    Regarding the warmup stage, shown in Table~\ref{tab:tfjs-warmup-kernel-latency-proportion-firefox}, we found that computation-intensive kernels continue to dominate the latency. However, memory-intensive kernels have a higher latency proportion in the Wasm backend. We also noted that compared to the vanilla Wasm, the proportion of computation-intensive kernels is lower during the warmup stage when acceleration techniques are enabled. For instance, under conditions enabling both SIMD and multithreading, the latency proportion of FusedConv2D and Einsum decreased by 2.2\% and 2.7\%, respectively. This stage includes memory allocation, which is not affected by hardware performance. Similar to the prediction stage, multithreading technology, due to the overhead of thread synchronization, actually increased the latency of memory-intensive kernels (as indicated by the gray cell color in the tables). In contrast, the WebGL backend does not face such issues, maintaining a consistent dominance of computation-intensive kernels in terms of warmup latency.

    In summary, the data on Firefox closely mirrors the results on Chrome, aligning with our previous analysis. The changes in kernel latency proportions and absolute values primarily depend on the model architecture and the implementation of the framework, independent of the browser. This is consistent with Wasm's browser-independent mechanism and corroborates our model-level conclusions in the Chrome browser.
}

\revise{
    \textbf{Results on mobile devices.}
    We also explored the results on mobile devices, displaying the prediction and warmup latencies in Tables~\ref{tab:tfjs-prediction-kernel-latency-proportion-mobile} and Tables~\ref{tab:tfjs-warmup-kernel-latency-proportion-mobile} respectively. Starting with the prediction stage, shown in Table~\ref{tab:tfjs-prediction-kernel-latency-proportion-mobile}, it was unsurprising that computationally intensive kernels continued to have the highest latency proportion, consistent with previous conclusions. After enabling SIMD, the proportions of FusedConv2d and Einsum decreased by 3.3\% and 1.5\%, respectively, mainly due to the significant acceleration SIMD brings to these computation-intensive kernels. However, SIMD provides limited acceleration for memory-intensive kernels. After enabling multithreading, we observed an increase in the proportions and the absolute latency values of these computation-intensive kernels by 2.0\% and 4.1\%, respectively. When both technologies were enabled, all computation-intensive kernels saw their largest proportion decrease, with FusedConv2d and Einsum reducing by 8.0\% and 5.2\%, respectively. These results align with the conclusions drawn from both browsers on the PC.

    \begin{table*}[h]
    \caption{\revise{Kernel prediction latency proportion in TF.js on mobile devices. The table format remains the same with Table~\ref{tab:tfjs-prediction-kernel-latency-proportion}.
    The results reveal that computation-intensive kernels primarily drive the prediction latency in in-browser inference during the prediction stage in all scenarios. For instance, the latency of FusedConv2D and Einsum accounts for more than 81\%. Memory-intensive kernels do not introduce much latency proportion in the WebGL backend compared to the Wasm backend, because of the need for thread synchronization in the Wasm backend, especially when multithreading is enabled.}}
    \label{tab:tfjs-prediction-kernel-latency-proportion-mobile}

    \resizebox{0.8\linewidth}{!}{%
        \begin{tabular}{lrrrr|l|lr}
            \toprule[1.5pt]
            Kernel                             & Wasm   & Wasm-S                         & Wasm-T                         & Wasm-S-T                         &  & Kernel                             & WebGL                \\ \cline{1-5} \cline{7-8} 
            \multicolumn{1}{l|}{Einsum}        & 43.4\% & \cellcolor[HTML]{34CDF9}41.9\% & \cellcolor[HTML]{34CDF9}42.9\% & \cellcolor[HTML]{34CDF9}38.2\%   &  & \multicolumn{1}{l|}{Einsum}        & 41.3\%               \\
            \multicolumn{1}{l|}{FusedConv2D}   & 51.1\% & \cellcolor[HTML]{34CDF9}47.8\% & \cellcolor[HTML]{34CDF9}49.9\% & \cellcolor[HTML]{34CDF9}43.1\%   &  & \multicolumn{1}{l|}{FusedConv2D}   & 45.1\%               \\
            \multicolumn{1}{l|}{GatherV2}      & 1.2\%  & \cellcolor[HTML]{F8FF00}3.9\%  & \cellcolor[HTML]{C0C0C0}2.5\%  & \cellcolor[HTML]{C0C0C0}9.8\%    &  & \multicolumn{1}{l|}{Add}           & 1.4\%                \\
            \multicolumn{1}{l|}{Slice}         & 0.6\%  & \cellcolor[HTML]{F8FF00}1.7\%  & \cellcolor[HTML]{C0C0C0}1.3\%  & \cellcolor[HTML]{C0C0C0}4.1\%    &  & \multicolumn{1}{l|}{Multiply}      & 1.0\%                \\
            \multicolumn{1}{l|}{FusedMatMul}   & 0.9\%  & \cellcolor[HTML]{34CDF9}1.8\%  & \cellcolor[HTML]{34CDF9}1.0\%  & \cellcolor[HTML]{34CDF9}2.0\%    &  & \multicolumn{1}{l|}{FusedMatMul}   & 3.7\%                \\
            \multicolumn{1}{l|}{FusedDwConv2D} & 1.8\%  & \cellcolor[HTML]{34CDF9}2.2\%  & \cellcolor[HTML]{34CDF9}1.7\%  & \cellcolor[HTML]{34CDF9}1.9\%    &  & \multicolumn{1}{l|}{FusedDwConv2D} & 2.8\%                \\ \bottomrule[1.5pt] 
        \end{tabular}%
    }
    
\end{table*}
    
    \begin{table*}[h]
    \caption{\revise{Kernel warmup latency proportion in TF.js in Firefox browser. The table format remains the same with Table~\ref{tab:tfjs-prediction-kernel-latency-proportion}.
    On mobile devices, computation-intensive kernels dominate the warmup total latency, with the latency of FusedConv2D and Einsum exceeding 80\% in all scenarios. Additionally, in the Wasm backend, the impact of memory-intensive kernel latency becomes significant upon enabling multithreading, resulting in a latency proportion of 12.5\% for GatherV2 and Slice.}}
    \label{tab:tfjs-warmup-kernel-latency-proportion-mobile}

    \resizebox{0.8\linewidth}{!}{%
        \begin{tabular}{lrrrr|l|lrr}
            \toprule[1.5pt]
            Kernel                             & Wasm               & Wasm-S                         & Wasm-T                         & Wasm-S-T                           &  & Kernel                             & WebGL           \\ \cline{1-5} \cline{7-8} 
            \multicolumn{1}{l|}{Einsum}        & 40.1\%             & \cellcolor[HTML]{34CDF9}36.2\% & \cellcolor[HTML]{34CDF9}37.8\% & \cellcolor[HTML]{34CDF9}35.1\%     &  & \multicolumn{1}{l|}{Einsum}        & 42.5\%          \\
            \multicolumn{1}{l|}{FusedConv2D}   & 50.5\%             & \cellcolor[HTML]{34CDF9}46.1\% & \cellcolor[HTML]{34CDF9}47.6\% & \cellcolor[HTML]{34CDF9}45.3\%     &  & \multicolumn{1}{l|}{FusedConv2D}   & 48.3\%          \\
            \multicolumn{1}{l|}{GatherV2}      & 3.1\%              & \cellcolor[HTML]{F8FF00}4.3\%  & \cellcolor[HTML]{C0C0C0}5.1\%  & \cellcolor[HTML]{C0C0C0}8.5\%      &  & \multicolumn{1}{l|}{Add}           & 1.4\%           \\
            \multicolumn{1}{l|}{Slice}         & 1.2\%              & \cellcolor[HTML]{F8FF00}1.9\%  & \cellcolor[HTML]{C0C0C0}3.9\%  & \cellcolor[HTML]{C0C0C0}4.0\%      &  & \multicolumn{1}{l|}{Multiply}      & 1.4\%           \\
            \multicolumn{1}{l|}{FusedMatMul}   & 2.4\%              & \cellcolor[HTML]{34CDF9}3.1\%  & \cellcolor[HTML]{34CDF9}3.0\%  & \cellcolor[HTML]{34CDF9}3.2\%      &  & \multicolumn{1}{l|}{FusedMatMul}   & 0.9\%           \\
            \multicolumn{1}{l|}{FusedDwConv2D} & 2.1\%              & \cellcolor[HTML]{34CDF9}1.9\%  & \cellcolor[HTML]{34CDF9}1.8\%  & \cellcolor[HTML]{34CDF9}2.0\%      &  & \multicolumn{1}{l|}{FusedDwConv2D} & 1.1\%           \\ \toprule[1.5pt] 
        \end{tabular}%
    }
    
\end{table*}

    Turning to the warmup stage results in Table~\ref{tab:tfjs-warmup-kernel-latency-proportion-mobile}, the proportion of latency attributed to memory-intensive kernels is higher in the Wasm backend compared to the WebGL backend, reaching 12.5\% after enabling SIMD and multithreading. This is mainly due to the limited optimization of these kernels by the framework when utilizing both acceleration technologies. The cost of memory allocation leads to a decrease in the proportion of computation-intensive kernels compared to the prediction stage, primarily because CPU capacity has minimal impact on memory allocation; this essentially adds a fixed value to the computational latency, thus reducing the proportion of computationally intensive kernels during the warmup stage. In the WebGL backend, there is no additional synchronization overhead like that in Wasm multithreading, hence the proportion of memory-intensive kernel latency remains low so that the memory-intensive kernels do not appear in the table.

    In summary, the results on mobile devices are consistent with those observed on PCs using Chrome and Firefox. The analysis involves only the framework and kernel implementation and optimization in two backends, thus conclusions and analyses applicable to PCs are equally relevant to mobile devices.
}

\subsection{Memory Footprint Analysis}\label{subsec:memory-footprint}

\begin{figure}[t]
    \centering
    \includegraphics[width=0.6\linewidth]{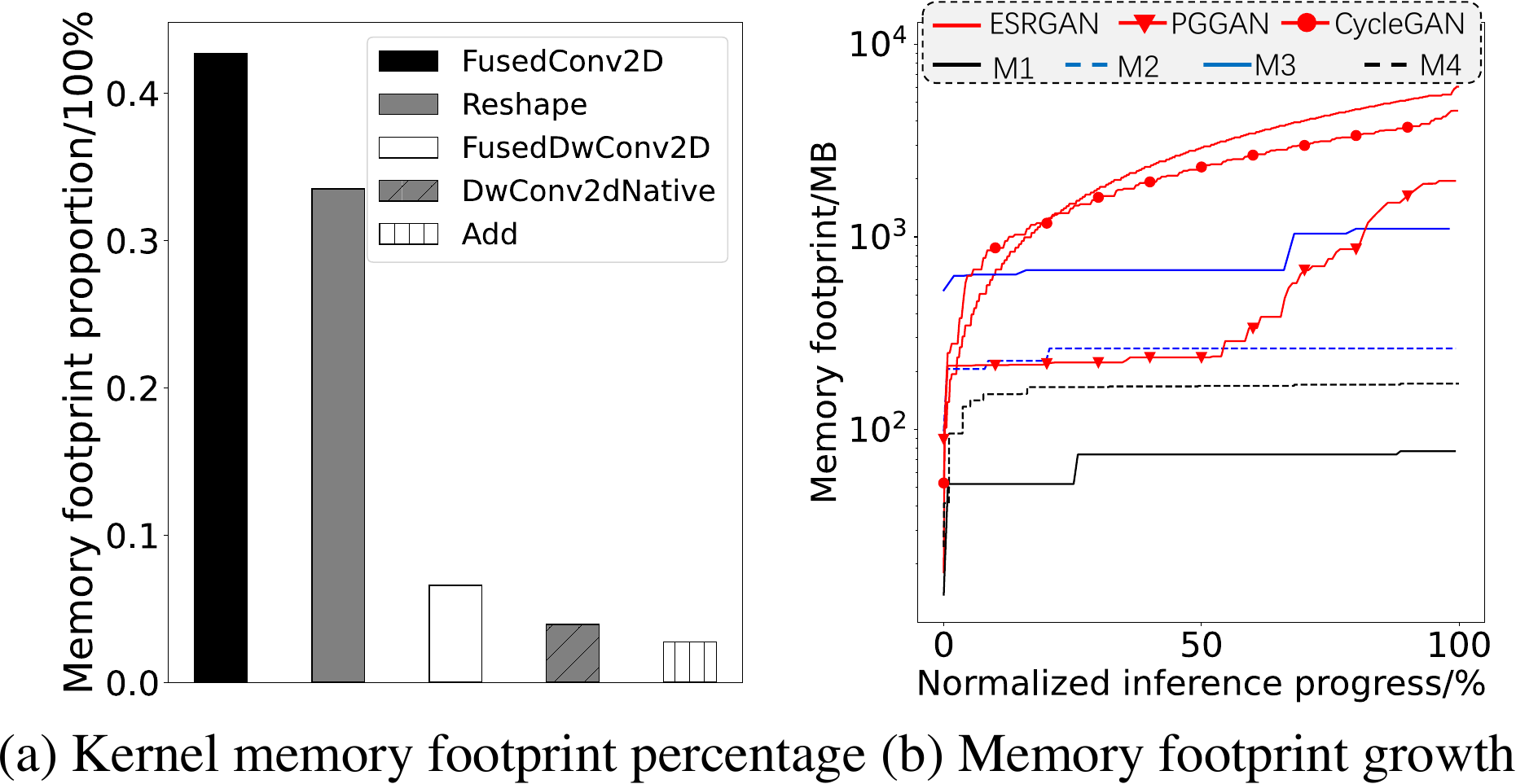}

    \caption{Memory footprint analysis.
    \revise{
        We found that: (a) Memory-intensive kernels can occupy a significant memory footprint. For example, the memory requirements for Reshape are second only to FusedConv2D, even though their latency proportions are quite low. (b) In most cases, inference memory tends to stabilize as the warmup progresses. However, for super-resolution models, the inference memory footprint continues to grow because the size of the intermediate data generated during inference gradually increases.
    }}
    \label{fig:memory-footprint}

\end{figure}

\revise{
    \sout{
        We find that the memory access pattern remains the same between the two backends and it is difficult to obtain WebGL memory usage in JavaScript. ORT.js implements and executes all logic in Wasm, hindering obtaining memory footprint in JavaScript. Owing to the above reasons, we only show the results in the Wasm backend of TF.js. SIMD and multithreading are enabled by default, but they do not affect the memory access.
    }
    
    We use the ``window.performance.memory'' JavaScript API to approximately measure the memory footprint. This API provides an approximation of a Web page's memory usage.
    We limit our analysis to the Wasm backend of TF.js for the following reasons:
    (1) After delving into the framework's implementation, we find that the memory access patterns are consistent across both backends within the same framework. 
    (2) The WebGL backend presents challenges in accurately profiling the memory footprint at the kernel level on the GPU side. This is because the API measures the JavaScript heap size, which is indicative of the memory footprint on the CPU, whereas GPU memory allocations are handled separately.
    (3) In the case of the Wasm backend, while ORT.js runs all logic, including kernel scheduling and execution, entirely in Wasm, this JavaScript API cannot execute within Wasm. We also lacked a reliable method to precisely capture the kernel memory footprint from the Wasm. In contrast, TF.js implements the scheduling logic in JavaScript, which subsequently invokes the Wasm kernels for computation, allowing us to use this API to effectively and accurately profile the kernel memory footprint.
    Due to these considerations, we focus on the Wasm backend of TF.js in our analysis. It's important to note that SIMD and multithreading, which are enabled by default, do not influence the memory footprint.
}
We measure the top-5 kernels that dominate the memory footprint during inference and illustrate the results in Figure~\ref{fig:memory-footprint}(a). We find that memory-intensive kernels, such as \textit{Reshape}, consume a significant amount of memory during inference, yet exhibit low latency. 

To further understand the memory access patterns of different models, we illustrate the memory growth during warmup in Figure~\ref{fig:memory-footprint}(b) because the framework only allocates memory in this stage.
We involve 3 super-resolution models in this experiment, i.e., ESRGAN~\cite{wang2018esrgan}, PGGAN~\cite{karras2017progressive} and CycleGAN~\cite{CycleGAN2017}. These models are excluded from the latency analysis because ORT.js does not support them in both backends. 
We find that the curves of the models used in latency analysis show a similar pattern, i.e., initially increasing and then stabilizing. This is because the framework retains the allocated memory without releasing it back to the operating system \revise{
    during warmup. Specifically, DL frameworks always manage allocated memory themselves to avoid the overhead of system calls. During the warmup stage, the framework requires memory from the operating system when self-managed free memory blocks are not enough, which is then consistently used for subsequent inferences. The blocks are released at the end of the inference process. This approach optimizes memory usage by reusing allocated memory and enhancing overall performance.
}
The memory footprint of the super-resolution model inference grows continuously.
Take ESRGAN \revise{inference} as an example, \revise{
    \sout{
        the peak memory footprint of ESRGAN inference bursts to 6,500MB, which is 334.6$\times$ of the model size
    }
    the peak memory footprint of the entire process spikes to 6,500MB, even though the ESRGAN model itself is only 19MB in size. This means that the memory required for the entire inference process is 334.6$\times$ greater than that of the model itself
}, despite \revise{the} batch size being one. \revise{\sout{and the model size being as small as 19MB.}}
This is attributed to the expansion of intermediate data size, surpassing the size of previously allocated memory blocks, leading to additional requests for larger memory blocks from the operating system and a continuous increase in memory footprint.
Such results emphasize the importance of efficient memory management design, as excessive memory footprint can hinder the ability for model inference.

\revise{
    \textbf{Results on Firefox browser and mobile devices.} 
    In the course of our analysis of DL inference memory footprint, we have considered extending our study to encompass data from Firefox and mobile devices. However, there are a couple of considerations that have led us to refrain from a detailed analysis in these areas. Firstly, on Firefox, the ``performance.memory'' API is unfortunately not available, and currently, we have not identified an alternative method that meets our needs for accurate memory usage assessment. Secondly, our preliminary observations of the memory usage data on mobile devices showed that the results are identical to those on PCs. This is expected since the memory footprint is only determined by the implementation of the framework, which is consistent across different devices. Given this uniformity and the lack of additional insights that new data might provide, we exclude these parts of the data in our analysis.
}

\revise{
    In summary, during model inference, there are numerous memory-intensive kernels. Although these kernels have low latency, they impose much memory pressure. DL frameworks manage their own memory allocation to reduce system call overhead. However, this approach may not be efficient. For instance, in super-resolution models, increasing intermediate data leads to continuous memory growth during inference. Therefore, more sophisticated memory management mechanisms, such as profile-based memory pool~\cite{wang2022melon}, are necessary. Additionally, we observed that memory footprint is solely related to model structure and framework implementation, independent of hardware, devices, or browser platforms.
}

\finding{
    \textbf{Finding 2}:
    Memory footprint can become a significant bottleneck during inference. For example, when performing super-resolution model inference like ESRGAN, the growing size of intermediate data can lead to a memory footprint surge of over 6,500MB, which is 334.6$\times$ of the model size. Inference of such models surpasses the capacity of most devices, necessitating the need for more effective memory management techniques in frameworks.
    The framework can utilize the model's profile information and allocate a large enough memory block to meet peak memory footprint requirements.
}

\section{Results of QoE Measurement}
\label{sec:qoe}


\subsection{Responsiveness Analysis}\label{subsec:qoe-responsiveness}
The responsiveness QoE results are listed in Table~\ref{tab:qoe-benchmark}.
Without inference, the benchmark yields a mean score of 142.0, with a standard deviation of 2.3.
When the DL is deployed, the mean score experiences a drop of up to 45.4\% for both frameworks, while the standard deviation increases up to 5.7$\times$. The degradation primarily stems from resource occupancy. 
\revise{
    Since the Speedometer benchmark performs tasks on both the CPU and GPU to simulate user interactions with Web pages, and model inference occupies at least one of the hardware resources, it inevitably leads to resource competition with the benchmark. This competition subsequently impacts the responsiveness scores.
}
In terms of CPU and integrated GPU, the mean score difference is not significant. 
\revise{
    \sout{
        For both frameworks, the scores on the integrated GPU are less than 2\% lower than the CPU.
    }
    The mean scores for the integrated GPU are 1.8\% lower than the CPU on average for TF.js, and 1.0\% lower for ORT.js. 
}
Since the benchmark contains tasks that run on both the CPU and GPU, the execution of the benchmark tasks is influenced by the DL inference in either backend. Moreover, the parallelism capacity of the integrated GPU is relatively weak, thereby impacting the final scores.
In contrast, the discrete GPU achieved the highest scores in all settings due to its outstanding performance.
\revise{
    According to the GPU benchmark~\cite{gpu-benchmark}, there exists a maximum benchmark score gap of 207.4$\times$ between the discrete GPU and the integrated GPU in our dataset.
}
We observed that the progress of benchmark tests slows down when inference and benchmark tasks run on the same hardware.
These results indicate a trade-off between responsiveness and inference performance. Achieving better responsiveness and low inference latency simultaneously becomes challenging. Timely responsiveness requires more resources to be allocated for processing requests, leaving fewer resources available for inference tasks.
\revise{
    For TF.js and ORT.js, the responsiveness QoE results are similar across both frameworks. Using TF.js as the baseline, ORT.js responsiveness scores are, on average, 99.5\% of TF.js on the Wasm backend. In the WebGL backend, on integrated and discrete GPUs, ORT.js responsiveness scores are 1.03$\times$ and 1.05$\times$ that of TF.js, respectively. This similarity is primarily because the responsiveness tests are only affected by resource competition, and both frameworks utilize similar resource configurations, such as multithreading and SIMD settings in Wasm. Therefore, the differences in responsiveness QoE resulting from using the two frameworks for inference are minimal.

    Additionally, we compared the responsiveness QoE of different models across various tasks. In the image classification task, the VGG16 model scored the highest, averaging 101.6. In the object detection task, the SSD-MobileNetV2 model performed best, with an average score of 115.1. In the grammar checking task, the MobileBert model achieved the best scores, averaging 113.9. However, we did not find a correlation between prediction latency and responsiveness QoE. For example, VGG16 introduces the highest prediction latency while its responsiveness is the highest; SSD-MobileNetV2 shows lower prediction latency than Yolo5-Middle but its responsiveness QoE is higher.
    We delved deeper into the distribution of test cases in the benchmark tests and found that CPU/GPU test cases alternated. In our experiments, inference and testing were also conducted alternately. We observed that if the WebGL backend was used during testing, the benchmark test would be blocked due to resource competition with WebGL. Testing would only resume when inference paused, and task switching within WebGL also incurred overhead. Therefore, the scores were affected by the distribution of inference and testing, but the fundamental cause was the resource competition introduced by inference.

    In summary, in-browser inference decreases the responsiveness QoE scores, primarily due to resource competition. The Speedometer benchmark includes responsiveness test cases for both CPU and GPU; thus, regardless of the hardware used as the inference backend, the inference process competes with the responsiveness test for resources. Discrete GPUs, due to their superior capacity, demonstrate the best results in responsiveness QoE benchmark tests.
}

\finding{
    \textbf{Finding 3}: In-browser inference can significantly degrade the responsiveness by up to 45.4\%. The reason is the computational resources required by inference compete with other critical components of the Web page.
    Allocating more CPU and GPU resources to the DL-powered Web applications to minimize inference latency comes at the expense of responsiveness QoE.
}

\begin{table*}[]
    \caption{Responsiveness of Web page. Each entry is the average responsiveness score and the value in the bracket is the standard deviation of the score. ``-D/-I'', \revise{``/''} and ``M*'' keep the same as Table~\ref{tab:e2e-webgl}. 
    \revise{We found that in-browser inference can negatively impact responsiveness, reducing the responsiveness score by up to 45.4\%. We found that the highest scores occur on discrete GPUs, due to their superior capacity.}}
    \label{tab:qoe-benchmark}

    \resizebox{\linewidth}{!}{%
    \begin{tabular}{lrrrrrrrrr}
        \toprule[1.5pt]
                        & \multicolumn{1}{c}{M1}       & \multicolumn{1}{c}{M2}   & \multicolumn{1}{c}{M3} & \multicolumn{1}{c}{M4}   & \multicolumn{1}{c}{M5}   & \multicolumn{1}{c}{M6}   & \multicolumn{1}{c}{M7}     & \multicolumn{1}{c}{M8}   & \multicolumn{1}{c}{M9} \\ \hline
        Baseline        & 142.0 \ \ (2.3)              & 142.0 \ \ (2.3)          & 142.0 (2.3)            & 142.0 \ \ (2.3)          & 142.0 \ \ (2.3)          & 142.0 \ \ (2.3)          & 142.0 (2.3)                & 142.0 \ \ (2.3)          & 142.0 \ \ (2.3)        \\
        TF.js-Wasm      & 92.7 \ \ (7.5)               & 89.5 (11.0)              & 91.2 (7.0)             & 112.0 \ \ (7.0)          & /                        & /                        & /                          & /                        & /                      \\
        ORT.js-Wasm     & 92.0 \ \ (5.0)               & 94.5 \ \ (9.2)           & 88.4 (9.8)             & 108.0 (11.0)             & 92.6 (12.0)              & 91.0 \ \ (7.0)           & 92.1 (6.6)                 & 115.0 (12.2)             & 83.7 \ \ (7.0)         \\
        TF.js-WebGL-I   & 87.6 (12.0)                  & 84.2 \ \ (7.5)           & 94.2 (8.3)             & 113.0 (13.0)             & 100.4 \ (9.3)            & 97.4 (11.2)              & 89.3 (9.5)                 & 106.0 \ \ (7.7)          & 83.1 (12.5)            \\
        TF.js-WebGL-D   & 118.1 \ \ (2.9)              & 119.2\ \ \ (3.8)         & 114.5 (3.6)            & 127.4 \ \ (7.4)          & 121.3 \ (6.3)            & 112.0 \ \ (2.4)          & 113.5 (3.1)                & 120.8 \ \ (3.6)          & 95.5 \ \ (8.6)         \\
        ORT.js-WebGL-I  & 89.7 \ \ (7.7)               & 88.2 \ \ (8.4)           & 96.2 (4.4)             & /                        & /                        & /                        & /                          & /                        & /                      \\
        ORT.js-WebGL-D  & 123.1 \ \ (5.1)              & 119.7\ \ \ (6.2)         & 125 (4.0)              & /                        & /                        & /                        & /                          & /                        & /                      \\ \bottomrule[1.5pt]
                           
    \end{tabular}%
    }

    \end{table*}
    
\begin{table*}[]
    \caption{Smoothness of Web page. Each entry is the average fps of the Web page/video and the value in the bracket is the standard deviation of fps across devices. ``M*''\revise{, ``/''} and ``-D/-I'' indicate the same as Table~\ref{tab:e2e-webgl}. \revise{
        We found that with in-browser inference, fps decreased by up to 62.7\%. Additionally, we found that fps performance on CPUs is better than on integrated GPUs, with an average fps reduction of 36.2\% on CPUs compared to 44.3\% on integrated GPUs. Discrete GPUs showed the best performance, with an average decrease of only 8.1\%.
    }}
    \label{tab:qoe-fps}

    \resizebox{\linewidth}{!}{%
        \begin{tabular}{lrrrrrrrrr}
        \toprule[1.5pt]
                       & \multicolumn{1}{c}{M1} & \multicolumn{1}{c}{M2}         & \multicolumn{1}{c}{M3}         & \multicolumn{1}{c}{M4} & \multicolumn{1}{c}{M5}      & \multicolumn{1}{c}{M6}    & \multicolumn{1}{c}{M7} & \multicolumn{1}{c}{M8} & \multicolumn{1}{c}{M9}    \\ \hline
        Baseline        & 30.0 (0.00)     & 30.0 (0.00)             & 30.0 (0.00)             & 30.0 (0.00)     & 30.0 (0.00)          & 30.0 (0.00)        & 60.0 (0.00)     & 60.0 (0.00)     & 60.0 (0.00)        \\  
        TF.js-wasm      & 23.8 (1.83)            & 22.9 (1.79)                    & 22.4 (1.61)                    & 23.9 (1.65)            & /                           & /                         & /                      & /                      & /                         \\
        ORT.js-wasm    & 24.1 (1.79)            & 23.1 (1.49)                    & 24.5 (1.71)                    & 23.8 (1.79)            & 21.7 (1.66)                 & 22.4 (1.66)               & 54.6 (4.07)            & 53.7 (3.71)            & 52.7 (3.51)               \\
        TF.js-WebGL-I   & 11.2 (0.77)            & 11.7 (0.85)                    & 12.9 (1.01)                    & 12.7 (0.82)            & 11.1 (0.84)                 & 11.5 (0.79)               & 26.8 (2.10)            & 27.2 (1.82)            & 23.5 (1.73)               \\
        TF.js-WebGL-D   & 27.2 (1.29)            & 26.4 (1.18)                    & 28.1 (1.51)                    & 28.3 (1.71)            & 27.9 (1.26)                 & 28.1 (1.49)               & 52.4 (2.44)            & 51.5 (3.19)            & 52.0 (2.85)               \\
        ORT.js-WebGL-I & 17.0 (1.11)            & 16.8 (1.11)                    & 17.1 (1.31)                    & /                      & /                           & /                         & /                      & /                      & /                         \\
        ORT.js-WebDL-D & 26.4 (1.39)            & 23.0 (1.08)                    & 24.5 (1.14)                    & /                      & /                           & /                         & /                      & /                      & /                         \\ \bottomrule[1.5pt]
        \end{tabular}%
    }

\end{table*}
\begin{table*}[]
    \caption{Inference accuracy on certain applications. Each entry is the average classification accuracy or detection mAP across devices. The value in the bracket is the standard deviation of the metric. ``M*'', \revise{``/''} and ``-D/-I" indicate the same as Table~\ref{tab:e2e-webgl}. \revise{We found that in-browser inference can reduce classification accuracy by up to 9.1\% and lower the detection mAP by up to 24\%.}}
    \label{tab:qoe-dl}

    \resizebox{0.85\linewidth}{!}{%
    \begin{tabular}{lrrrrrr}
    \toprule[1.5pt]
                   & \multicolumn{1}{c}{M1}               & \multicolumn{1}{c}{M2}               & \multicolumn{1}{c}{M3}              & \multicolumn{1}{c}{M4}                & \multicolumn{1}{c}{M5}                & \multicolumn{1}{c}{M6} \\ \hline
    Baseline        & 17.2\% (0.0)\quad\quad\quad\quad     & 17.2\% (0.0)\quad\quad\quad\quad     & 17.2\% (0.0)\quad\quad\quad\quad      & 1.00 (0.0)\quad\quad\quad\quad        & 1.00  (0.0)\quad\quad\quad\quad       & 1.00  (0.0)\quad\quad\quad\quad       \\
    TF.js-Wasm      & 13.7\% (0.53$\times10^{-2}$)         & 13.5\% (0.52$\times10^{-2}$)         & 8.8\% (0.33$\times10^{-2}$)         & 0.78 (2.91$\times10^{-2}$)            & /                                     & /                              \\
    ORT.js-Wasm    & 13.7\% (0.50$\times10^{-2}$)         & 13.6\% (0.52$\times10^{-2}$)         & 8.0\% (0.30$\times10^{-2}$)         & 0.78 (3.10$\times10^{-2}$)            & 0.77 (2.84$\times10^{-2}$)            & 0.79 (3.00$\times10^{-2}$)     \\
    TF.js-WebGL-I   & 8.7\%  (0.33$\times10^{-2}$)         & 8.8\%  (0.32$\times10^{-2}$)         & 3.9\% (0.14$\times10^{-2}$)         & 0.76 (2.86$\times10^{-2}$)            & 0.76 (2.93$\times10^{-2}$)            & 0.78 (2.83$\times10^{-2}$)     \\
    TF.js-WebGL-D   & 14.9\% (0.54$\times10^{-2}$)         & 14.2\% (0.52$\times10^{-2}$)         & 11.1\% (0.44$\times10^{-2}$)        & 0.95 (3.26$\times10^{-2}$)            & 0.94 (3.41$\times10^{-2}$)            & 0.96 (3.59$\times10^{-2}$)     \\
    ORT.js-WebGL-I & 10.5\% (0.39$\times10^{-2}$)         & 11.1\% (0.45$\times10^{-2}$)         & 7.9\% (0.30$\times10^{-2}$)         & /                                     & /                                     & /                              \\
    ORT.js-WebGL-D & 15.1\% (0.57$\times10^{-2}$)         & 15.1\% (0.56$\times10^{-2}$)         & 13.5\% (0.54$\times10^{-2}$)        & /                                     & /                                     & /                              \\ \bottomrule[1.5pt]
    \end{tabular}%
    }

\end{table*}

\subsection{Smoothness Analysis}\label{subsec:qoe-fluency}

\begin{figure}[t]
    \centering
    \includegraphics[width=0.7\textwidth]{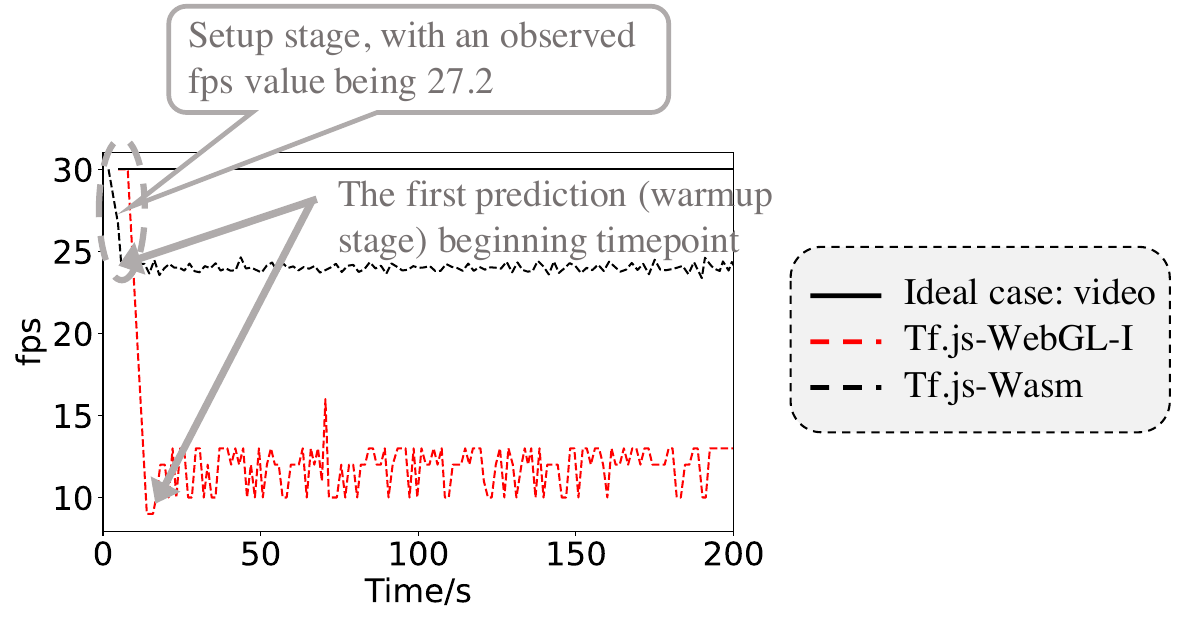}
    \caption{Fps trace during SSD-MobileNetV2 inference. \revise{We observed that during in-browser inference, fps drops sharply and then continues to fluctuate, with the amplitude reaching 31.5\% of the average value when inference is performed on the integrated GPU. The fps during the setup stage is higher than that during the prediction stage.}}
    \label{fig:fps-example}
\end{figure}

The average fps QoE results are listed in Table~\ref{tab:qoe-fps}. Additionally, Figure~\ref{fig:fps-example} illustrates the fps trace during SSD-MobileNetV2 inference in two backends in TF.js on a device (MacBook Pro 2020). Without inference, the average Web page rendering fps on the device is 60, and the average video playing fps is 30. When DL is deployed, the fps value inevitably drops (up to 62.7\%). 
The integrated GPU exhibits the poorest performance in terms of smoothness, while the performance of the CPU is remarkably good, approaching that of the discrete GPU. Specifically, when performing DL model inference on the CPU, the fps experiences a relatively smaller drop, averaging 36.2\% for both frameworks, compared to performance inference on the integrated GPU. Discrete GPU experiences 8.1\% less fps degradation than the CPU and 44.3\% less fps degradation than the integrated GPU. Furthermore, we observe that fps fluctuations are more severe when inferring on the integrated GPU, as depicted by the red dashed line in Figure~\ref{fig:fps-example}. In Figure~\ref{fig:fps-example}, the fps amplitude reaches 4, which is 31.5\% of the average value (12.7). The fps amplitude can reach up to 37.3\% of the average fps for all models.
The reason is that rendering and video playing rely on GPU resources, leading to inevitable resource competition when performing inference in WebGL. The best performance on discrete GPU is attributed to its high capacity. Moreover, in WebGL, the discrete GPU is not fully harnessed, allowing rendering and video-playing tasks to utilize the remaining capabilities.
\revise{
    Additionally, we observed that during the setup stage, the fps is higher than the other stages. This is evidenced by the grey dashed line in Figure~\ref{fig:fps-example}, where there is a sharp initial decline marking the setup stage. At this stage, there is an inflection point corresponding to an fps value of 27.2. Since the setup stage requires less CPU and GPU resources, the fps during this stage is comparatively higher than the other stages. However, due to the very low setup latency of the SSD-MobileNetV2, this segment occupies only a brief segment on the x-axis of Figure~\ref{fig:fps-example}.
}

While performing inference on integrated GPU offers lower prediction latency, it may lead to fps degradation, because of its relatively lower capacity. On the contrary, performance inference on the CPU is slower but leads to less fps compromise since it alleviates resource competition. These results indicate the existence of a trade-off between smoothness and inference performance. In terms of smoothness, the priority of using integrated GPU for inference is lower than other hardware.

\revise{
    Furthermore, we also compared the smoothness QoE of different models across tasks. In the image classification task, the MobileNetV2 model had the best fps, averaging 21.6. In the object detection task, the SSD-MobileNetV2 model achieved the highest fps, averaging 22.2. In the grammar checking task, the Bert-Base model recorded the best fps, averaging 44.6. We observed that the fps differences among models within the same task are small. For example, during image classification tasks using the wasm backend of TF.js, the fps difference between the three models was only 1.4. This was mainly because the smoothness experiments only involved rendering tasks, which were not blocked due to the resource competition from inference. Although the page rendering continued normally, it became less smooth instead of completely stalling, thus resulting in minimal fps differences among different models. However, regardless of the model, the inevitable resource competition always resulted in a decrease in smoothness QoE.

    In summary, in-browser inference leads to significant resource contention, negatively impacting Web page rendering or video playing. Regardless of the framework or backend used, there is a noticeable drop in fps. Although WebGL offers lower inference latency, when inference is performed on integrated GPU, it competes with rendering tasks for GPU resources. In contrast, using Wasm as the inference backend results in a higher smoothness QoE. In our experiments, we found that discrete GPUs provide the best smoothness QoE, primarily because they are more powerful than other hardware options.
}

\finding{
    \textbf{Finding 4}: 
    In-browser inference impacts smoothness, with fps dropping by up to 62.7\%. This is due to resource competition. The resource competition can lead up to 37.3\% fps fluctuations. Achieving both smoothness and inference performance is a trade-off. Using the Wasm backend for inference can lead to better smoothness compared to the WebGL backend on devices without a powerful discrete GPU.
}

\subsection{Inference Accuracy Analysis}\label{subsec:qoe-dl}
The average inference accuracy QoE results are listed in Table~\ref{tab:qoe-dl}.
In the ideal scenario where all images can be processed by the DL model, the classification accuracy is 17.2\% and the detection mAP is 1.0. This lower classification accuracy is attributed to the dataset containing low-resolution images like 32$\times$32 pixel size, and some frames contain multiple objects, which can also affect classification accuracy. The high detection mAP is mainly because the object region typically covers a significant portion of the image, and the dataset exclusively consists of images with a single class (face). This characteristic is well-suited for in-browser detection tasks, like background blurring, where humans are the predominant subject. 
We chose these two benchmark datasets specifically for their inclusion of both video frames sourced from real-world video streams and corresponding timestamps, and they are used in numerous research papers~\cite{fan2019lasot, liu2023objects, cai2019exploring, wang2018cosface, pfister2015flowing, schroff2015facenet}.

When performing inference in the browser, we observe that the classification accuracy drops by up to 9.1\%, and the detection mAP drops by up to 24\%. 
\revise{
    \sout{
        Similar to \S\ref{subsec:qoe-fluency}, the performance of the integrated GPU remains inferior, while the CPU shows better performance. The reduced video fps and increased inference latency result in the DL model being unable to process all frames. As a result, some frames have to reuse the inference results from the previous frame, leading to a decrease in inference accuracy~\cite{canel2019scaling, kang2017noscope, chen2015glimpse}.
    }
    Specifically, on CPUs, the reduction in classification accuracy and mAP for object detection reached up to 9.2\% and 0.23, respectively. On integrated GPUs, the two metrics decreased by up to 13.3\% and 0.24, respectively. On discrete GPUs, the reductions were up to 6.1\% and 0.06, respectively. We found that these results are very similar to previous findings. Firstly, discrete GPUs consistently show the best results across all settings due to their performance advantages. In contrast, integrated GPUs perform worse than CPUs in terms of inference accuracy QoE. The main reason is that we considered the continuous inference scenario in the experiment, such as processing a video segment. If the inference is performed on integrated GPUs, it competes for resources with rendering tasks, leading to reduced rendering fps. This can force some video frames to reuse results from previous frames, deteriorating the quality of the output~\cite{canel2019scaling, kang2017noscope, chen2015glimpse}. Therefore, for tasks that require frequent inference, such as background blur in online video meetings, using integrated GPUs might not be a better choice than CPU in terms of inference accuracy QoE. Although model inference on CPU is slower than that on integrated GPUs, the lesser competition for resources can achieve higher QoE through parallel processing.

    Additionally, we compared the inference accuracy QoE of different models within the same task. In the image classification task, the MobileNetV2 model had the best inference accuracy, averaging 12.8\%, while the VGG16 model performed the worst, averaging 8.9\%. We examined how each model processed every video frame and found that the VGG16 model, due to its high prediction latency, led to more video frames having to reuse previous results, thereby degrading its performance. Conversely, the MobileNetV2 model reused the fewest results. Although the ResNet50 model has higher accuracy than MobileNetV2 on the ImageNet dataset, its performance deteriorated due to excessive frame reuse.
    In the object detection task, we found that the Yolo5-Middle model achieved the best inference accuracy, averaging 0.84, while the SSD-MobileNetV2 model showed the poorest performance, averaging 0.82. This is because SSD-MobileNetV2 is the only supported object detection model in the Wasm backend of TF.js, and the results on the CPU are poorer than the average. However, with respect to the other backends, EfficientDet has the worst performance.
    Since the detection tasks in our dataset were relatively straightforward, the performance of the EfficientDet model was only marginally lower than other models by at most 0.02. The Yolo5-Middle model, benefiting from its lower latency and superior pretrained accuracy, performed best in the inference accuracy QoE experiments. 
    Therefore, to achieve higher inference accuracy QoE, it should be avoided to use heavier models. Lightweight models can achieve comparable outcomes and are preferable due to their efficiency and lesser resource demands.

    In summary, in-browser inference leads to inference accuracy degradation. The constrained computational resources and reduced fps pose challenges for the DL model to process all inputs effectively. Performing inference in a slower backend may bring better inference accuracy.
}

\finding{
    \textbf{Finding 5}: In-browser inference can degrade the inference accuracy. We observe a decrease of up to 9.1\% in classification accuracy and 24\% in detection mAP. When performing inference in Wasm, applications could achieve higher inference accuracy at the cost of a minor latency increase, compared with inference in WebGL on integrated GPU.
    The results show that reducing inference latency does not necessarily equate to an improvement in inference accuracy QoE, \revise{especially for video applications that require continuous DL inference}.
}

\revise{
    \subsection{Analysis in an Integrated Environment}\label{sec:qoe-integrated}
    
    We conducted an analysis of three QoE metrics within a unified environment. Since we could not find any open-source websites deployed with DL inference, we developed an online video conferencing Web application. This app captures video from the device's camera and displays it on the Web page. At the same time, we embedded the Speedometer benchmark test within the page. The Web page uses SSD-MobileNetV2 for face detection, monitoring the video's fps data in real time while detecting faces. The logic for online face detection remains the same as before, reusing previous results when the current frame cannot be processed. Additionally, we captured the entire monitor's display and used YOLOv8 for detection to serve as ground truth. The results of this part of the experiment are displayed in Table~\ref{tab:qoe-integrated}.

    \begin{table*}[]
    \caption{\revise{QoE measurement in an integrated results. The format of each entry is the same as the previous tables. 
    We found that with in-browser inference, all three QoE metrics decreased, with a greater decline observed when evaluated individually. Specifically, the responsiveness score dropped by 18.8 more, and the fps decreased by 4.5 more.}}
    \label{tab:qoe-integrated}

    \resizebox{\linewidth}{!}{%
        \begin{tabular}{lrrrrrrr}
        \toprule[1.5pt]
                        & \multicolumn{1}{c}{Baseline} & \multicolumn{1}{c}{TF.js-wasm}     & \multicolumn{1}{c}{ORT.js-wasm}       & \multicolumn{1}{c}{TF.js-WebGL-I} & \multicolumn{1}{c}{TF.js-WebGL-D}      & \multicolumn{1}{c}{ORT.js-WebGL-I}    & \multicolumn{1}{c}{ORT.js-WebGL-I}     \\ \hline
        Responsiveness  & 142.0 \ \ (2.3)                  & 104.2 \ \ (7.2)                        & 101.7 (10.1)                          & 94.2 (14.1)                       & 115.8 (11.5)                           & /                                     & /                                      \\  
        Smoothness      & 30.0 (0.00)                  & 20.9 (1.82)                        & 21.4  (1.73)                          & 10.3 (2.84)                       & 23.8  (1.94)                           & /                                     & /                                      \\
        Infer. Acc.     & 1.0 (0.00)                   & 0.92 (2.80$\times10^{-2}$)         & 0.91 (3.10$\times10^{-2}$)            & 0.84 (4.60$\times10^{-2}$)        & 0.98 (4.60$\times10^{-2}$)             & /                                     & /                                      \\ \bottomrule[1.5pt]
        \end{tabular}%
    }

\end{table*}

    We found that compared to previous standalone tests, the QoE results obtained from this part of the test were worse. 
    First, regarding responsiveness, compared to the baseline (without in-browser inference), the average QoE score for responsiveness in this scenario decreased by up to 47.8, with the standard deviation increasing by up to 11.8; and compared to just deploying in-browser inference with the Speedometer benchmark test, the average QoE score dropped by up to 18.8. This is mainly because the benchmark test also included many rendering tasks, which also compete for resources with the benchmark test, hence worsening the QoE results. This demonstrates that resource competition can impact responsiveness. The difference between the two frameworks is very low; with TF.js as the baseline, the gap between ORT.js and TF.js is less than 2.4\%. We also found that responsiveness scores were lowest on integrated GPUs, primarily due to limited GPU performance and resource competition.
    Next, the results for smoothness. Compared to results without in-browser inference, fps in this scenario dropped by up to 19.7, and the standard deviation increased by up to 2.84. Compared to the scenario in \S\ref{subsec:qoe-fluency}, the fps in this scenario was lower, with the average fps being lower by up to 4.5. The reason remains the same as before, due to the impact of resource competition, as the Speedometer benchmark test also occupies CPU and GPU resources, thus worsening fps. As concluded in \S\ref{subsec:qoe-fluency}, integrated GPUs, being less powerful, showed the worst results.
    Lastly, the results for inference accuracy. Compared to the ideal case, where the DL model could process all video frames, the inference accuracy QoE, i.e., detection mAP, in this scenario decreased by up to 0.16. Since the dataset used differs from \S\ref{subsec:qoe-dl}, a direct comparison is not possible. Nevertheless, our previous conclusion still holds that resource competition causes inference to reuse previous results, leading to impaired inference accuracy QoE

    In summary, after testing in a unified environment, we found that our previous conclusions hold true: the impact of resource competition can cause a decline in QoE. We can also see that each QoE metric declines when in-browser inference leads to resource competition. Therefore, ensuring QoE is a trade-off and should be chosen based on specific application requirements. For video applications, smoothness usually has a higher priority; while for applications with intensive interactions, responsiveness is more important. Web application developers should adjust in-browser inference settings according to different application needs, such as choosing the backend and model to use.
}
\finding{
    \revise{
        \textbf{Finding 6}: In-browser inference introduces inevitable resource competition, resulting in a decline across three QoE metrics. Compared to the baseline, in a unified environment, we observed that responsiveness QoE dropped by up to 47.8, smoothness QoE by up to 19.7, and inference accuracy by up to 0.16. Balancing these three QoE metrics involves trade-offs and should be tailored based on the specific needs of the application, such as prioritizing page rendering smoothness. For applications where smoothness is a priority, using Wasm as the inference backend typically yields better results than integrated GPUs.
    }
}

\section{Implications and Suggestions}
This section discusses a series of actionable implications based on our findings.

\noindent  \textit{\textbf{For Web browser vendors:}} 
\textbf{(1) Extend Wasm SIMD instructions (Finding 1)}. 
Limited SIMD instructions in Wasm hinder the potential optimization. Browser vendors can consider extending the SIMD instructions with more features, such as FMA and data-moving instructions. This enhancement can empower the in-browser inference to leverage advanced hardware capabilities better.
\revise{
    There are two possible approaches to extending SIMD capabilities for enhanced in-browser inference performance. The first is updating the Wasm SIMD specification to include longer SIMD instructions, aiming for transparency in higher-level applications. This approach would require effort support from browser vendors to add support in browser Wasm VM, and may also require effort within the Web community to refine the Wasm specification. The second is adding browser extensions for DL inference. As in-browser inference grows and more Web services integrate DL models, this approach requires to development of browser extensions specifically for DL models. These extensions could enable the browser itself to execute kernels like matrix multiplication directly, providing interfaces to the JavaScript side. Since browsers can directly access native resources, they are well-positioned to implement extended SIMD capabilities. This method would only require support from browser vendors.
}
\textbf{(2) Support loading pre-compiled WebGL binaries (Finding 1)}. 
\revise{
    \sout{
        Browser vendors can consider extending the WebGL API to accommodate the loading of pre-compiled binaries, to avoid time-consuming in-browser WebGL shader compilation. 
    }
    Currently, the compilation of WebGL in browsers involves translating WebGL shader code into an intermediate representation, such as how Chrome on Windows platforms compiles WebGL into Direct3D~\cite{ANGLE}, which is then compiled by the GPU hardware into executable binaries. This entire process occurs within the browser. However, the form of the intermediate representation is limited, and converting WebGL to intermediate form is time-consuming. Therefore, browser vendors could consider implementing a feature that allows the loading of pre-compiled WebGL intermediate representations directly into the browser. This would help avoid the time-consuming in-browser compilation of WebGL shaders, potentially reducing the warmup latency of in-browser inference.
}
This improvement can not only benefit the in-browser inference but also enhance the performance of other Web applications.

\noindent  \textit{\textbf{For DL framework vendors:}}
\textbf{(1) Compile WebGL binaries ahead-of-time (Finding 1)}.
In-browser compilation introduces significant warmup overhead to in-browser inference.
\revise{
    \sout{
        Given the limited and hardware-agnostic WebGL compilation targets (e.g., Direct3D on Windows~\cite{direct3d} and Metal on OS X~\cite{metal}), pre-compiling WebGL binaries to these targets and loading them on demand can mitigate the issue.
    }
    In browsers, WebGL is first compiled into an intermediate representation. For example, ANGLE can compile WebGL to Direct3D on Windows and Metal on OS X~\cite{ANGLE, direct3d, metal}. These intermediate representations are then converted into executable machine code by the hardware. Given that the number of these intermediate representations is relatively limited, framework developers can pre-compile WebGL programs into these representations. During inference, the framework can download and load the necessary representations into the browser. This approach helps save on compilation time and can alleviate the issue of high warmup latency.
}
This suggestion complements and synergizes with the second suggestion for Web browser vendors.
\textbf{(2) Focus more on memory-intensive kernels (Finding 1)}.
Memory-intensive kernels can introduce high latency, underscoring the importance of paying attention to these kernels. For example, framework vendors can make efforts to fully utilize multithreading in the implementation of these kernels.
\textbf{(3) Manage memory through an efficient memory pool (Finding 2)}. 
When the size of intermediate data surpasses the allocated memory block size, frameworks need to require more memory. To mitigate the problem, memory pool techniques, optimized by profiling inference process or DL model, can be employed for efficient memory management~\cite{wang2022melon, zhou2023mpress}. 
\revise{
    \sout{
        For example, the memory access pattern can be saved during model conversion and the inference framework can allocate all memory directly during the setup stage. The size of the memory block should be able to meet the peak memory usage requirements. In this way, data can be accommodated within the block, avoiding redundant memory allocation and memory fragments as much as possible.
    }
    Currently, the models used for in-browser inference require format conversion from native models. During the conversion, it is possible to gather information about the model’s memory access patterns, including the kernels themselves and the amount of memory each kernel needs to utilize. This information can be saved and downloaded alongside the model conversion. During the setup stage, the framework can pre-allocate memory while loading the model, which can help reduce warmup latency. Since pre-allocation of memory and data transfer can occur concurrently, this process does not generate additional overhead.
    When allocating memory, the method pool proposed by Melon~\cite{wang2022melon} can be utilized. This approach involves pre-allocating a large enough memory block and then placing data during the inference process at different locations within this block, as long as data reads and writes do not conflict. In this way, data can be accommodated within the block, avoiding redundant memory allocation and memory fragments as much as possible. Implementing this strategy could significantly optimize memory management, thereby enhancing the efficiency and performance of in-browser inference.
}

\noindent  \textit{\textbf{For Web application developers:}}
\textbf{(1) Incorporate DL models with varying capacities (Finding 1)}. Web application developers can encapsulate multiple models within the application, each with different sizes and capacities, such as MobileNet~\cite{mobilenetv22018, mobilenetv32019} and ResNet~\cite{he2016deep} catering to image classification, and download the suitable model according to hardware information.
\revise{
    \sout{
        Web application developers can dynamically select the optimal model without introducing additional packaging and setup overhead.
    }
    Specifically, on the GPU side, developers can use the ``WEBGL\_debug\_renderer\_info'' extension~\cite{WEBGL-debug-renderer-info} to obtain information about the GPU, and then select the suitable model based on this information. For instance, if a discrete GPU is detected, a model with powerful capacity can be chosen, whereas a lighter model may be preferable for integrated GPUs.
    On the CPU side, since it is difficult for developers to directly obtain specifications such as the CPU model, a performance benchmark can be constructed in advance. This benchmark involves a fixed-size matrix multiplication kernel (or other DL kernels) and measures its latency on CPUs of varying capacities. By executing this kernel during the Web page loading process and estimating the performance of the user's CPU based on the kernel's latency, it becomes possible to choose either a lightweight model or a model with a more powerful capacity according to the assessed capability of the CPU.
}
\textbf{(2) Prioritize responsiveness (Finding 3)}. Due to the resource contention generated by handling user requests and DL model inference, adopting intermittent inference can alleviate resource competition, ensuring the timely responsiveness of Web pages to user requests.
\textbf{(3) Enhance smooth rendering (Finding 4)}. Since rendering is typically GPU-dependent, using the Wasm backend for inference could guarantee smooth rendering. While Wasm has slightly higher inference latency than WebGL, it circumvents rendering resource competition, especially for devices with only integrated GPUs. This can introduce a significant improvement in Web page smoothness at the cost of a little inference latency.
\textbf{(4) Guarantee inference accuracy (Finding 5)}. For the video applications we used, frame filtering techniques~\cite{li2020reducto} can effectively enhance inference accuracy. Additionally, utilizing Wasm as the inference backend is a practical choice, aligning with the previous suggestions and the findings.
\revise{
    \textbf{(5) Balancing resource allocation for optimal QoE (Finding 6)}. It is important to note that, deploying DL inference in browsers inevitably leads to resource competition, with inference tasks occupying either the CPU or GPU. This necessitates the consideration of inference configurations based on the specific applications involved. For example, while rendering tasks rely heavily on the GPU, user interactions depend more on the CPU. Therefore, ensuring optimal QoE across all three metrics becomes a trade-off. Prioritizing GPU for rendering will consume more CPU resources, and vice versa. 
    \minor{
        \sout{
            Developers need to make decisions based on the specific applications. For instance, users typically prefer smooth rendering for tasks such as online video streaming, whereas for interactive applications such as online documents, user engagement and responsiveness are more critical. Developers should choose the appropriate inference backend for different scenarios to better ensure QoE.
        }
        Developers can choose the appropriate QoE metric based on the application type, such as video playback or online document. Then developers can select the optimal backend to optimize the QoE metric based on the specific QoE metric. Responsiveness QoE can be quantified through the response time to user request, which is equal to the “runs per minute” defined in the paper; smoothness QoE can be quantified through fps, which can be obtained by the JavaScript API “requestAnimationFrame”. As for inference accuracy QoE, developers can quantify it through user feedback because there may not be ground truth for user inputs. For instance, if a developer is developing Web pages for video playback, the smoothness QoE should be prioritized, meaning that ensuring smooth video rendering should come first. When incorporating in-browser inference, it is essential to maximize inference accuracy QoE while maintaining smooth rendering. In such scenarios, the importance of responsiveness QoE may not be as significant as the other two. At this point, choosing CPU as the inference backend may be given higher priority. Developers can also optimize the QoE by monitoring the fps of video playback to select either a lightweight or more capable model. When developing Web pages for online documents, responsiveness QoE should be the primary focus, ensuring that timely response to user interactions. When introducing in-browser inference, it is necessary to maximize inference accuracy QoE without compromising responsiveness QoE, where smoothness QoE may become less critical compared to responsiveness. In this scenario, GPU could usually be a better choice. Depending on the type of Web page, developers can prioritize different QoE metrics based on user requirements and feedback.
    }
}
\section{Discussion}

\subsection{Threats to Validity}

\noindent \textbf{Selection of models.} Similar to previous work~\cite{ma2019moving}, we select models that cover several classical tasks, and the models are widely deployed to user devices. We excluded LSTM, Speech-to-text, and GPT models because TF.js and ORT.js do not support them. In addition, Bert outperforms LSTM in many language tasks~\cite{devlin2018bert}. Because models can be represented as a sequence of kernels~\cite{abadi2016tensorflow, onnx} and the basic kernel set is the same, we believe such selection introduces negligible bias in our results with respect to inference performance.

\minor{
    \noindent \textbf{Model format conversion.} We used the models from the TF model hub for different tasks. These models were pre-trained on servers using TF and then converted to the file format recognizable by the in-browser inference framework using model conversion tools. Although TF.js supports training, its efficiency of model training on servers is significantly lower than TF, and the computational costs of model training in browsers make it impractical. The conversion process does not modify the model architecture or its kernels, thus it does not impact inference performance. The models in the model hub are not specifically designed for in-browser inference, which may lead to conversion failures primarily due to model design. In practical deployments, developers can design models based on the specific needs of the applications. Although TF.js and ORT.js might not support certain models, we believe this will not hinder the progress of in-browser inference. Furthermore, our experimental results are independent of the model format conversion, affirming the validity of our results and findings.
}

\noindent \textbf{Selection of frameworks.} We select the two frameworks based on multiple considerations, including documentation, popularity, backend support, and maintenance. For example, Brain.js~\cite{brainjs} lacks support for CPU; Keras.js is no longer maintained; WebDNN~\cite{Webdnn} lacks implementation of most kernels in TF and ORT.
\minor{    
    TF.js is compatible with original TF models, whereas ORT.js handles models in the ONNX format. Popular DL frameworks, including PyTorch and TensorFlow, allow for their models to be converted to ONNX, a necessary step for mobile deployment in the case of PyTorch. As such, TF.js and ORT.js are capable of supporting most in-browser inference applications. These frameworks are robustly maintained over the long term due to their development under the auspices of Google and Microsoft, respectively. In contrast, the slower update pace of other frameworks can be attributed to the lack of dedicated maintenance teams and the high costs associated with keeping up with rapid model iterations. Because of the extensive adoption of in-browser inference and strong community support for TF.js and ORT.js, we believe that in-browser inference is promising. 
}
Although we exclude ORT.js in the kernel-level latency and memory footprint analysis, the underlying logic of the DL framework is similar, i.e., computing each kernel of the model. The kernel computation logic is the same and the latency distribution is framework-agnostic. \minor{Based on these points}, we believe our results are valid.

\noindent \textbf{Selection of tasks.} We select vision and text tasks for the experiment. These tasks are popular on the Web and there have been similar deployed services. Despite the ``inference accuracy'' in the QoE experiment being task-specific, the inference accuracy will degrade inevitably due to resource competition. We involve super-resolution tasks only in the memory footprint analysis because of the immature support. However, excluding this task in other parts does not introduce too much bias to the final conclusions, i.e., large latency gaps compared with native inference, large latency variations across devices, and degradation of the user QoE. So we believe our results and findings are still reliable.
\minor{
    Regarding the experimental setup, our focus was on the performance of model inference, which is independent of application logic. Deployed real-world applications are not open-sourced because of the confidentiality requirements of commercial applications. Nevertheless, considering that model inference performance is influenced solely by hardware and browser capabilities, we designed Web pages specifically for measuring inference performance, containing only the logic related to model inference.
}

\noindent \textbf{Selection of devices.} We used 50 PC devices \minor{and 20 mobile devices} for the experiment. The CPU and GPU of the devices cover 82\% devices in AI Benchmark~\cite{aibenchmark-ranking} w.r.t. the AI score. 
Although the device selection may introduce bias to results, we believe the bias is acceptable and the results and findings are still valid. 

\revise{
\subsection{Future Work}
Optimizing in-browser inference performance is a crucial issue. Our analysis has already indicated that on the CPU side, the main challenges during the prediction stage of in-browser inference stem from environmental differences between native and browser settings. Future work could focus on addressing these issues. For instance, expanding the SIMD instruction set in Wasm to support more efficient operations, such as FMA, could enhance performance significantly. Another potential approach could be to bypass Wasm altogether and implement model kernels, such as matrix multiplication, directly within the browser's JavaScript engine. Since browsers can directly access advanced hardware features and bypass the Wasm interpreter, this could substantially improve inference performance. We are currently exploring optimizations in this direction.
For the warmup stage, considering the high in-browser compilation overhead associated with WebGL, introducing support for importing precompiled intermediate representations into the browser could significantly reduce latency during this stage. This approach would streamline the process, allowing for quicker start-up times and enhancing the overall responsiveness of web applications utilizing in-browser inference. We leave this as a future direction to further optimize the in-browser inference performance.
As for the memory footprint of in-browser inference, we aim to optimize the memory management mechanism of the inference framework. Specifically, this optimization mainly includes the following steps: (1) First, during the model conversion phase, it is necessary to retain the model structure and information about each kernel. (2) Based on this information, obtain the memory access patterns and use optimized memory pool techniques to pre-calculate the strategy of memory management and allocation~\cite{wang2022melon}. (3) Save this information along with the model files so that the inference framework can download and reduce the inference memory footprint.

In summary, our future work primarily focuses on optimizing the performance of in-browser inference, as enhancing inference performance will inevitably lead to improvements in QoE. For instance, lower latency means that devices have more time to process user requests and the DL model can handle more video frames. Additionally, conserving memory footprint will allow browsers to operate more smoothly, avoid stuttering, and thus enhance the smoothness QoE.
}
\section{Related Work}
\label{sec:related}

\noindent \textbf{DL performance.} 
DL performance is critical to DL-powered software development and deployment, as well as to end users' QoE. For example, high accuracy and low inference latency are desired~\cite{xu2018deeptype, bonawitz2019towards}. 
To characterize the DL performance, many research efforts in the SE community have been presented~\cite{ha2019deepperf, cao2022understanding, chen2022deepperform, pham2020problems, du2019deepstellar, gao2020estimating}. For example, 
DeepPerf~\cite{ha2019deepperf} is an end-to-end DL-based solution that can predict the performance of a new DL task through a pre-trained model;
DeepPerform~\cite{chen2022deepperform} enables efficient performance testing for AdNN~\cite{bateni2018apnet};
DNNMem~\cite{gao2020estimating} estimates the memory consumption of both the DL model and the DL framework systematically and accurately.
Additionally, Ma et al.~\cite{ma2019moving} measured the in-browser inference latency at model level for manually constructed models.
\revise{
   \sout{
        In comparison, our study aims to anatomize in-browser inference of real-world DL models at both model level and kernel level, as well as the memory footprint and user QoE of in-browser inference.
   }
   Currently, most research focuses on inference performance in native environments, with a lack of detailed analysis of inference performance within browser environments. Unlike native environments, browsers isolate the inference process from the operating system, preventing access to advanced system features, such as efficient SIMD instructions. Our research aims to fill this knowledge gap by exploring inference performance from the model level to the kernel level, covering aspects such as latency and memory footprint. While Ma et al.~\cite{ma2019moving} explored manually constructed models, they neither considered models and applications deployed in the real world, nor analyzed and compared the differences between browser and native environments. In contrast, our study is designed to thoroughly dissect in-browser inference of real-world DL models at both the model and kernel levels, as well as explore the memory footprint and user QoE. This comprehensive approach allows us to provide deeper insights and recommendations for optimizing in-browser inference performance.
}

\noindent \textbf{DL deployment.} DL deployment focuses on fitting models into the different platforms after the DL model is well tested~\cite{mittal2016spotgarbage, xu2018deeptype, guo2019empirical, chen2018tvm}. Many frameworks facilitate the deployment in the cloud, \revise{\sout{like} such as} TensorFlow~\cite{abadi2016tensorflow} for servers, TFLite~\cite{tflite} for Android, and Core ML~\cite{coreml} for iOS. 
Many SE efforts are proposed to promote DL deployment~\cite{chen2021empirical, chen2020comprehensive, cummaudo2020interpreting, guo2019empirical}.
For example,  
Cummaudo et al.~\cite{cummaudo2020interpreting} analyzed pain points that developers face when deploying DL to servers.
Chen et al.~\cite{chen2021empirical, chen2020comprehensive} studied the DL deployment issues and challenges by exploring forums such as Stack Overflow.
Guo et al.~\cite{guo2019empirical} investigated the performance disparity when the pre-trained DL models are migrated to mobile and Web platforms. They analyze the model-level performance in the pure JavaScript backend. 
\revise{
    \sout{
        In contrast, our study seeks to delve into the performance of more advanced backends, namely Wasm and WebGL. We aim to dissect the underlying reasons for the performance gap between browser and PC environments, provide detailed analysis at the fine-grained kernel level, and assess the impact of in-browser inference on user QoE.
    }
    Most existing research still focuses on native environments. Although Guo et al.~\cite{guo2019empirical} considered DL model deployment using pure JavaScript backend, they did not explore advanced backends such as Wasm. At present, advanced backends are widely adopted for in-browser inference due to their superior performance capabilities. However, to our best knowledge, these backends have not yet been thoroughly investigated. In contrast, our study aims to delve into the performance of more advanced backends, i.e., Wasm and WebGL. We seek to understand the reasons behind the performance disparities between browser and native environments, provide detailed analysis at the fine-grained kernel level, and assess the impact of in-browser inference on user QoE.
}

\noindent \textbf{Web browsing performance.} Many metrics have been proposed to quantify the Web browsing performance, such as Page Load Time (PLT)~\cite{PLT}, Time-to-First-Byte (TTFB)~\cite{TTFB}, Speed Index~\cite{speedindex}, etc. 
Several efforts have been presented to explore and optimize the performance. 
Lighthouse~\cite{lighthouse} is a tool to analyze the quality of Web pages and is integrated into Chrome. WProf~\cite{wang2013demystifying} is proposed to profile the PLT, which is a key performance metric that many techniques aim to reduce~\cite{kelton2017improving, mardani2021horcrux}. SipLoader~\cite{liu2022fusing} is a Speed Index oriented page scheduler. Enrico et. al proposed a comprehensive study to measure the QoE of Web users~\cite{bocchi2016measuring}.
Vetter~\cite{yang2023visual} is an automated tool for visual distortions testing and debugging.
\revise{
    \sout{
        However, they mainly focus on the loading phase, and the performance after the loading stage is not well-studied.
    }
    Current research focusing on browser performance and QoE primarily targets the Web page loading phase or concerns visual distortions. None of these studies have considered scenarios involving the deployment of DL inference. The loading phase occurs only at the beginning of Web page navigation, whereas, by comparison, DL inference continues throughout the user's interaction with the page. Existing studies have not considered this scenario, and the existing QoE metrics are not applicable. In contrast, our work is the first to propose QoE metrics tailored for in-browser scenarios and carries out a detailed quantitative analysis of these metrics.
}

\section{Conclusion}
\label{sec:conclusion}

We anatomized the in-browser inference performance and its impact on QoE. Results show that: 
(1) the latency gap between in-browser inference and native inference reaches up to 16.9$\times$ on CPU and 30.6$\times$ on GPU. The discrepancy in CPU is mainly attributed to different SIMD instruction sets and Wasm runtime-inherent inefficiency, as well as resource competition within the browser; the discrepancy in GPU is mainly attributed to inefficient software libraries and GPU abstractions.
(2)~Memory can be the bottleneck. When inferring super-resolution models, the memory requirement can reach up to 334.6$\times$ the size of the models themselves.
(3) QoE will degrade due to inevitable resource competition. Inferring models on CPU may bring better QoE due to the tread-off between inference latency and QoE. 

\section{Acknowledgement}
This work was supported by the National Natural Science Foundation of China under grant numbers 62325201 and  62102009.

\bibliographystyle{ACM-Reference-Format}
\bibliography{ref}


\end{document}